\documentclass{article}

    \PassOptionsToPackage{numbers, compress}{natbib}

\usepackage{neurips_2021}

\usepackage[utf8]{inputenc} 
\usepackage[T1]{fontenc}    
\usepackage{hyperref}       
\usepackage{url}            
\usepackage{booktabs}       
\usepackage{amsfonts}       
\usepackage{nicefrac}       
\usepackage{microtype}      
\usepackage{xcolor}         

\usepackage[pdftex]{graphicx}
\usepackage{comment}
\usepackage{colortbl}
\usepackage[flushleft]{threeparttable}
\usepackage{multirow}
\usepackage{makecell}
\usepackage{tabu}
\usepackage{pifont}
\usepackage{hyperref}
\usepackage{amsmath, amsfonts, amssymb}
\usepackage{bm}
\usepackage{tabularx}

\usepackage{bbm} 
\usepackage{ulem} 
\usepackage{xspace}
\usepackage{float}
\usepackage{cellspace}

\usepackage{longtable}

\definecolor{maroon}{cmyk}{0,0.87,0.68,0.32}
\definecolor{Gray}{gray}{0.9}
\newcommand{\abs}[1]{\lvert #1 \rvert}
\newcommand{\norm}[1]{\left\lVert#1\right\rVert}
\newcommand\Tstrut{\rule{0pt}{2.2ex}}         
\newcommand\Bstrut{\rule[-1ex]{0pt}{0pt}}   

\newcommand*\encircle[1]{\raisebox{.5pt}{\textcircled{\raisebox{-.9pt} {#1}}}}

\newcommand{\pignpi}{PIG'N'PI\xspace}
\newcommand{\lemos}{GN+\xspace}

\newcommand{\potEnergy}{P}
\newcommand{\force}{\bm{F}}
\newcommand{\position}{\bm{r}}
\newcommand{\APPENDIX}{SI} 

\newcommand{\eg}{\textit{e.g.}}

\definecolor{mybrown}{RGB}{255, 213, 132}
\definecolor{mypink}{RGB}{214, 135, 249}

\title{Learning Physics-Consistent Particle Interactions}

\author{%
Anonymous Author(s)
}

\begin{document}
\normalem

\maketitle

\begin{abstract}
   Interacting particle systems play a key role in science and engineering. Access to the governing particle interaction law is fundamental for a complete understanding of such systems. However, the inherent system complexity keeps the particle interaction hidden in many cases. Machine learning methods have the potential to learn the behavior of interacting particle systems by combining experiments with data analysis methods. However, most existing algorithms focus on learning the kinetics at the particle level. Learning pairwise interaction, \textit{e.g.}, pairwise force or pairwise potential energy, remains an open challenge. Here, we propose an algorithm that adapts the Graph Networks framework, which contains an edge part to learn the pairwise interaction and a node part to model the dynamics at particle level. Different from existing approaches that use neural networks in both parts, we design a deterministic operator in the node part that allows to precisely infer the pairwise interactions that are consistent with underlying physical laws by only being trained to predict the particle acceleration. We test the proposed methodology on multiple datasets and demonstrate that it achieves superior performance in inferring correctly the pairwise interactions while also being consistent with the underlying physics on all the datasets. The proposed framework is scalable to larger systems and transferable to any type of particle interactions, contrary to the previously proposed purely data-driven solutions. The developed methodology can support a better understanding and discovery of the underlying particle interaction laws, and hence guide the design of materials with targeted properties.
\end{abstract}

\section{Introduction}
\label{sec:introduction}

Interacting particle systems play a key role in nature and engineering as they govern planetary motion~\cite{murray1999solar}, mass movement processes~\cite{dramis2016mass} such as landslides and debris flow, bulk material packaging~\cite{sawyer2014mechanistic}, magnetic particle transport for biomaterials~\cite{furlani2010magnetic}, and many more~\cite{wang2019coarse, angioletti2017theory}. Since the macroscopic behavior of such particle systems is the result of interactions between individual particles, knowing the governing interaction law is required to better understand, model and predict the kinetic behaviour of these systems. Particle interactions are  determined by a combination of various factors including contact, friction, electrostatic charge, gravity, and chemical interaction, which affect the particles at various scales.
The inherent complexity of particle systems inhibits the study of the underlying interaction law. Hence, they remain largely unknown and particle systems are mostly studied in a stochastic framework or with simulations based on simplistic laws. 

Recent efforts on developing machine learning (ML) methods for the discovery of particle interaction laws have shown great potential in overcoming these challenges~\cite{radovic2018machine, carleo2019machine, zhou2020graph, shlomi2020graph, atz2021geometric, mendez2021geometric, park2021accurate}. These ML methods, such as the \emph{Graph Network-based Simulators} (GNS)~\cite{sanchez2020learning} for simulating physical processes, \emph{Dynamics Extraction From cryo-em Map} (DEFMap)~\cite{matsumoto2021extraction} for learning atomic fluctuations in proteins, the \emph{SchNet}~\cite{schutt2017schnet, schutt2018schnet} which can learn the molecular energy and the \emph{neural relational inference model} (NRI)~\cite{kipf2018neural} developed for inferring heterogeneous interactions, can be applied on various types of interacting particle systems such as water particles, sand and plastically deformable particles. They allow implicit and explicit learning of the mechanical behavior of particle systems without prior assumptions and simplifications of the underlying mechanisms. A commonly applied approach is to predict directly the kinetics of the particles without explicitly modeling the interactions~\cite{schutt2017schnet, unke2019physnet, klicpera2020directional, klicpera2020fast, bapst2020unveiling, sanchez2020learning, hu2021forcenet}. The neural networks, then, map directly the input states to the particle acceleration, occasionally by virtue of macroscopic potential energy~\cite{sanchez2020learning, schutt2017schnet,unke2019physnet, klicpera2020directional, klicpera2020fast}. While these approaches give an accurate  prediction of the particle system as it evolves, they do neither provide any knowledge about the fundamental laws governing the particle interactions nor are they able to extract the particle interactions precisely. 

Recent work~\cite{cranmer2020discovering} proposed an explicit model for the topology of particle systems, which imposes a strong inductive bias and, hence, provides access to individual pairwise particle interactions. \cite{cranmer2020discovering} demonstrated that their Graph Network (GN) framework predicts well the kinetics of the particles system. However, as we will show, the inferred particle interactions violate Newtonian laws of motion, such as the action-reaction property, which states that two particles exert the same but opposed force onto each other. Therefore, the extracted pairwise particle interactions do not correspond to the real underlying particle interaction \emph{force} or \emph{potential}, which are the fundamental properties of a physical system. The origin of these discrepancies lies in the design of the GN approach, which does not sufficiently constrain the output space, and clearly demonstrates the need for a physics-consistent Graph Neural Network framework for particle interactions.

\begin{figure*}[t!]
    \centering
    \includegraphics[width=1.0\linewidth]{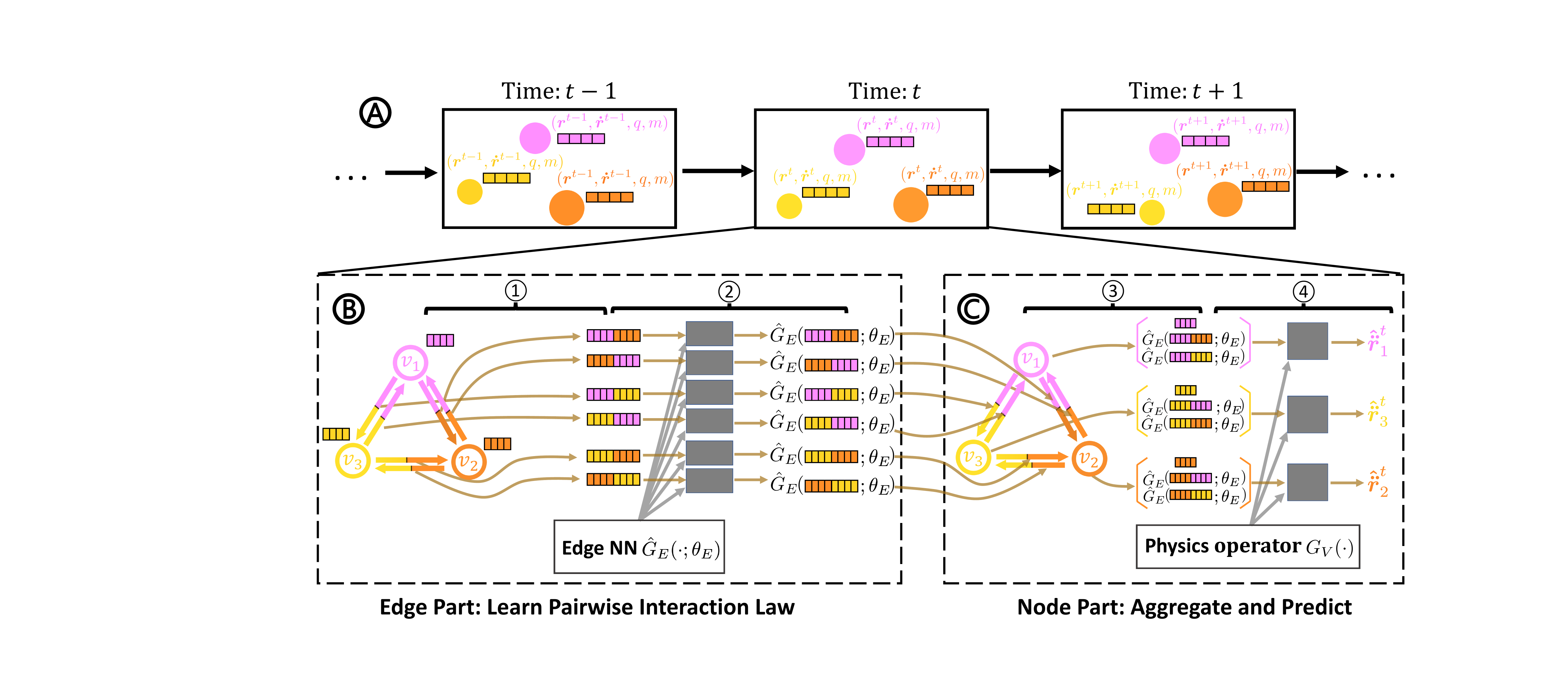}
    \caption{\textbf{Framework of the proposed model to learn pairwise force or pairwise potential energy. } (\encircle{A}) The  interacting particle system contains three particles that evolve over time. At every time step, each particle is described by multiple features, which include position, velocity, charge and mass (represented by the bar). Position and velocity evolve with time whereas other properties remain constant. (\encircle{B}-\encircle{C}) The proposed method learns physics-consistent pairwise force or pairwise potential at every time step $t$. The model has two components: the edge part \encircle{B} and the node part \encircle{C}. In the edge part (\encircle{B}), two nodes' vectors are concatenated as edge feature (process \ding{172}). An edge neural network $\hat{G}_E(\cdot; \theta_{E})$ ($\theta_{E}$ represents the trainable parameters) takes the edge feature as input (process \ding{173}) and outputs a learnt vector on that edge representing the pairwise force or potential energy. In the node part (\encircle{C}), the output vectors by the edge neural network and the raw node feature are aggregated on each node (process \ding{174}). We design the deterministic node operator $G_V(\cdot)$ by incorporating physics knowledge to derive the net acceleration on nodes (process \ding{175}). By minimizing the loss on node-level accelerations, the edge neural network $\hat{G}_E(\cdot; \theta_{E})$ will output pairwise force or potential energy exactly.}
    \label{fig:framework}
\end{figure*}

Here, we propose a Graph Neural Network (GNN) framework that incorporates universal physical laws, specifically Newton's second law of motion, to learn the interaction potential and force of any physical particle system. The proposed algorithm, termed \underline{p}hysics-\underline{i}nduced \underline{g}raph \underline{n}etwork for \underline{p}article \underline{i}nteraction (\pignpi), combines the graph neural network methodology with deterministic physics operators to guarantee physics consistency when indirectly inferring the particle interaction forces and potential energy (Fig.~\ref{fig:framework}). We will show that \pignpi learns the pairwise particle potential and force by only being trained to predict the particle acceleration  (without providing any supervision on the particle interactions). Moreover, we will  show that the inferred interactions by \pignpi are physically consistent (contrary to those inferred by purely data-driven approaches). We will further demonstrate that predictions provided by \pignpi are more accurate, generalize better to larger systems and are more robust to noise than those provided by purely data-driven graph network approaches. Moreover, we will demonstrate on a case study that is close to real applications that the proposed algorithm is scalable to large systems and is applicable to any type of particle interaction laws.

\section{PIG'N'PI: Physics-induced Graph Network for Particle Interaction}
\label{sec:architecture}

We propose a framework that is able to infer pairwise  particle forces or potential energy by simply observing particle motion in time and space. In order to provide physics-consistent results, a key requirement is that  the learnt particle interactions need to satisfy Newtonian dynamics. One of the main challenges in developing such a learning algorithm is that only information on the particle position in time and space along with particle properties (\textit{e.g.}, charge and mass) can be used for training the algorithm and no ground truth information on the interactions is available since it is very difficult to measure it in real systems. 

The proposed framework comprises the following distinctive elements (Fig. \ref{fig:framework}): 1) a graph network with a strong inductive bias representing the particles, their properties and their pairwise interactions; and 2) physics-consistent kinetics imposed by a combination of a neural network for learning the edge function and a deterministic physics operator for computing the node function within the graph network architecture. In addition, the proposed framework consists of two steps: 1) training the network to predict the particle motion in time and space; and 2) extraction of the pairwise forces or the pairwise potential energy from the edge functions of the trained network. 

\textit{Particle systems} We consider particle systems that are moving in space and time and are subject to Newtonian dynamics without any external forces.  A particle system in this research is represented by the directed graph $G=(V, E)$, where nodes $V = \{v_1, v_2, \ldots, v_{|V|}\}$ correspond to the particles and the directed edges $E = \{e_{ij}: v_i, v_j \in V, i\neq j\} $ correspond to their interactions. {The graph is fully-connected if all particles interact with each other.} Each particle $i$, represented by a node $v_i$, is characterized by its time-invariant properties, such as charge $q_i$ and mass $m_i$ and time-dependent properties such as its position $\bm{r}_i^t$ and its velocity $\bm{\dot{r}}_i^t$. We use $\bm{\eta}_i^t$ to denote the features of particle $i$ at time $t$, $\bm{\eta}_i^t = [\bm{r}_i^t, \bm{\dot{r}}_i^t, q_i, m_i]$. We limit our evaluations to particle systems comprising homogeneous particle types. This results in particles exhibiting only one type of interaction with all its neighboring particles, leading to $|E| = |V| (|V| - 1)$. We further assume that the position $\bm{r}_i^t$ of each particle $i$  is observed at each time step $t$ and that this information is available during training. Based on the position information $\bm{r}_i^t$,  velocity $\bm{\dot{r}}_i^t$ and acceleration $\bm{\ddot{r}}_i^t$ are computed. 

\begin{figure*}[t!]
\centering
\includegraphics[width=0.9\linewidth]{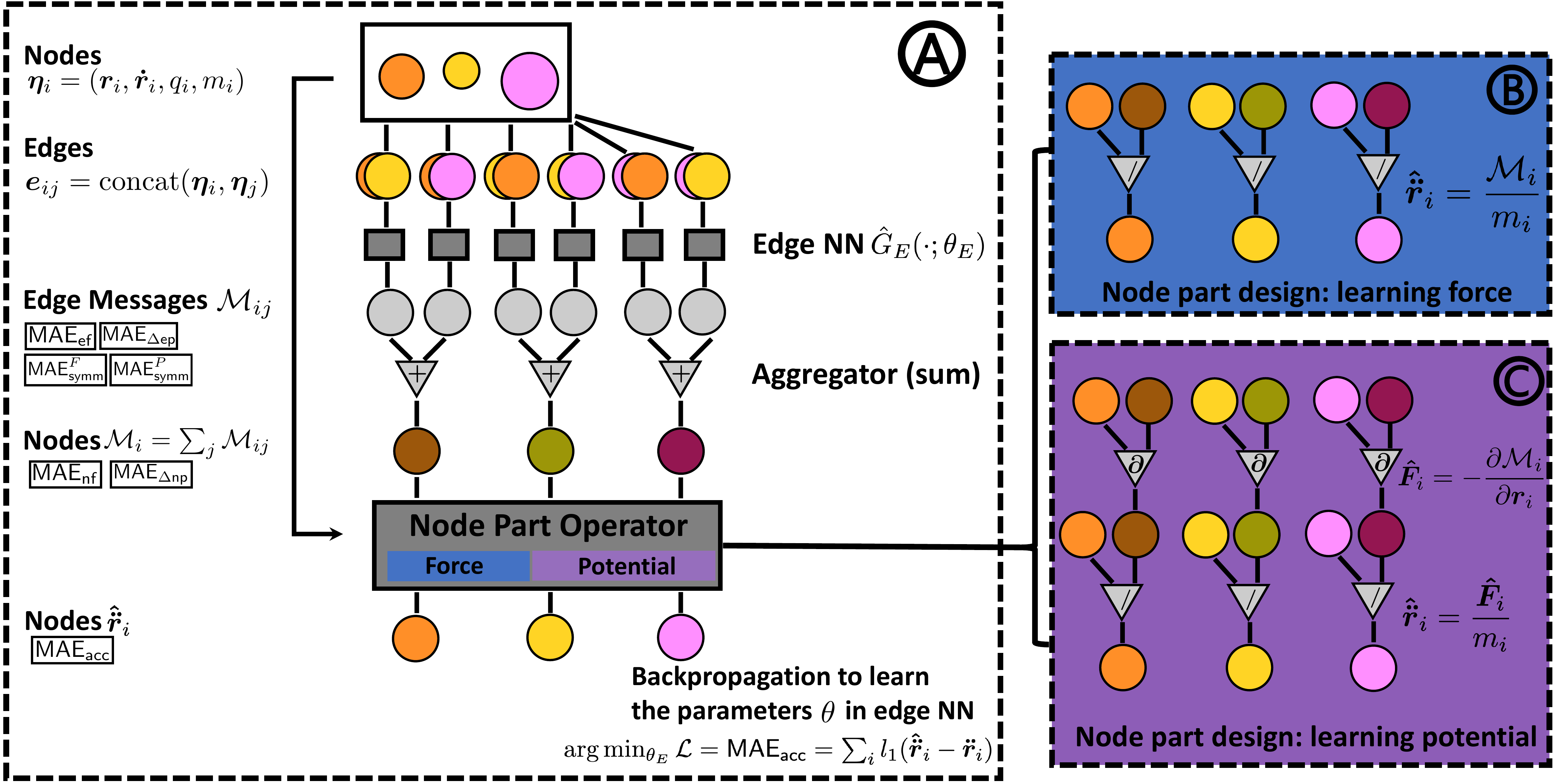}
\caption{\textbf{\underline{P}hysics-\underline{i}nduced \underline{g}raph \underline{n}etwork for \underline{p}article \underline{i}nteraction (\pignpi)}. (A) The workflow where the edge neural network $\hat{G}_E(\cdot; \theta_{E})$ takes edge features as input. The corresponding output message $\mathcal{M}_{ij}$ is the predicted pairwise force or potential energy, depending on the physics operator (B) or (C) in the node part. Parameters $\theta_{E}$ in $\hat{G}_E(\cdot; \theta_{E})$ are trained by minimizing the loss on particle acceleration.}.
\label{fig:method_illustration}
\end{figure*}

\textit{Proposed framework} The proposed  \underline{p}hysics-\underline{i}nduced \underline{g}raph \underline{n}etwork for \underline{p}article \underline{i}nteraction (\pignpi) framework extends the general Graph Network framework proposed by \cite{battaglia2018relational}, which is a generalized form of message-passing graph neural networks. The architecture of the proposed framework is illustrated in Fig.~\ref{fig:method_illustration}.  We use a directed graph to represent the interacting particle system where nodes correspond to the particles and edges correspond to their interactions. The framework imposes a strong inductive bias and enables to learn the position-invariant interactions across the entire particle system network. Given the particle graph structure, the input is then defined by the node features $\bm{\eta}_i$ (representing particle's characteristics). The target output is the acceleration $\bm{\ddot{r}}_i^{t}$ of each node at time step $t$. {The} standard GNs {block} \cite{battaglia2018relational}, typically, comprises two neural networks: an edge neural network $\hat{G}_E(\cdot; \theta_{E})$ and a node neural network $\hat{G}_V(\cdot; \theta_{V})$, where $\theta_{E}$ and $\theta_{V}$ are the trainable parameters. Here, we propose to substitute the node neural network $\hat{G}_V(\cdot; \theta_{V})$ by a deterministic node operator $G_V(\cdot)$ to ensure that the learned particle interactions are consistent with the underlying physical laws.  The main novelty compared to the standard GN framework is, that we impose known basic physical laws to ensure that the inferred pairwise force or potential energy corresponds to the real force or potential energy while only being trained on predicting the acceleration of the particles.

It is important to emphasize that only information on particle positions is used for training the algorithm and the ground-truth information on the forces and the potential energy is not available during training.  For each edge, the property vectors $\bm{\eta}_i$ of two nodes connected by an edge are concatenated as the edge feature vector. The edge neural network $\hat{G}_E(\cdot; \theta_{E})$ outputs a message on every edge that corresponds to the pairwise force or potential energy. The output dimension is set to be the same as the spatial dimension $d$ (two or three) if $\hat{G}_E(\cdot; \theta_{E})$ is targeted to learn the pairwise force or one to learn the pairwise potential energy. Edge messages are aggregated on nodes and the node part operator computes the output corresponding to the acceleration of nodes, imposing physics-consistency on edge messages. Trained to predict the node-level acceleration, once applied to a new particle system, the GN predicts the particle motion at consecutive time steps. The pairwise forces or the pairwise potential energy can then be extracted from the edge function for each time step.

\emph{Contributions of the present work compared to precious research} Here, we propose a methodology to learn the \emph{pairwise} force or \emph{pairwise} potential energy from the observed particle trajectories. This focus distinguishes our work from many previous works such as~\cite{greydanus2019hamiltonian, sanchez2019hamiltonian, lutter2019deep, schutt2017schnet, wang2019machine, finzi2020simplifying} that learn the energy of the system and then derive the \emph{per-particle} dynamics from the global energy. Moreover, as outlined above, our proposed approach does not have access to any ground truth information during training but rather learns to infer the force and potential energy indirectly. This is contrary to the previously proposed approaches that rely on such information~\cite{unke2019physnet, hu2021forcenet}. 

While our proposed framework has several similarities with two previously proposed frameworks that are also aiming to infer pairwise force and pairwise potential energy using also only the particle accelerations for training~\cite{cranmer2020discovering, cranmer2020discovering}, none of the previously proposed methods is able to infer the underlying particle interactions that are consistent with the underlying physical laws. We demonstrate in our experiments that the learnt particle interactions of the previously proposed approaches are not consistent with the underlying physical laws and do not correspond to the real forces or potential energy. 

In fact, the proposed algorithm has also  similarities to the Physics-informed neural networks (PINNs)~\cite{karniadakis2021physics} which aim to solve partial differential equations. Both PINNs and \pignpi integrate known physical laws. While \pignpi integrates Newton's second law, PINNs enforce the structure imposed by partial differential equation  at a finite set of collocation points.

\section{Results and discussion}
\label{sec:experiment}

\subsection{Performance evaluation metrics}
\label{sec:performance_evaluation}

We evaluate the performance of the proposed \pignpi framework on synthetic data generated from two- ($d=2$) and three- ($d=3$) dimensional numerical simulations. The key distinctive property of the generated datasets is the definition of the inter-particle potential energy $\potEnergy$, which defines the inter-particle pairwise force by $\force = -\partial \potEnergy / \partial \position$. The selected cases, which have also been used in prior work~\cite{cranmer2020discovering} and can be considered as a benchmark case study, cover a wide range of particle interaction features, including dependence on particle properties, \textit{e.g.}, mass and charge, dependence on interaction properties, \textit{e.g.}, stiffness, and varying degrees of smoothness (see Table~\ref{table:dataset-property} and \APPENDIX~\ref{sec:force-potential-functions-visualization} for visualization). 



\begin{table}[t!]
  \centering
  \caption{The force and potential energy equations for different datasets,  where $\bm{F}_{ij}$ is the force from particle $j$ to particle $i$ , $P_{ij}$ is the potential incurred by particle $j$ on particle $i$, $r_{ij}$ is the Euclidean distance between particle $i$ and particle $j$, $\bm{n}_{ij}$ is the unit vector pointing from particle $i$ to particle $j$, $q_i$ and $m_i$  are the electric charge and mass of particle $i$. $k$, $L$, $c$ and $\Theta$ are constants. 
  }
  \label{table:dataset-property}
  \begin{tabular}{Sc Sl Sl }
    \toprule
\textbf{Dataset} & \textbf{Pairwise force ($\bm{F}_{ij}$)}& \textbf{Pairwise Potential ($P_{ij}$)} \\
    \midrule
       {Spring} & $k(r_{ij}- L)\bm{n}_{ij}$ &$\frac{1}{2}k(r_{ij}- L )^2$\\
       \hline
       {Charge} & $-cq_i q_j\bm{n}_{ij}/r_{ij}^2$ & $cq_i q_j/r_{ij}$ \\
       \hline
       {Orbital} & $m_i m_j\bm{n}_{ij}/r_{ij}$ & $m_i m_j \text{ln}(r_{ij})$\\
       \hline
       {Discnt} & \makecell[l]{
       $\bm{0}$, {if $r_{ij} < \Theta$}\\
       $(r_{ij} -1)\bm{n}_{ij}$,  {otherwise}
       }

       & \makecell[l]{$0$, {if $r_{ij} < \Theta$}
       \\
       $0.5 (r_{ij} -1)^2$, {otherwise}
       }

    \\
       
  \bottomrule
\end{tabular}
\end{table}


The method developed by \cite{cranmer2020discovering}, which applies multilayer perceptrons (MLPs \cite{goodfellow2016deep}) in the edge and node part, serves as the baseline for comparison. We do not change the architecture of the baseline except for changing the output dimension of its edge part MLPs when learning the pairwise force or potential energy. 
The output dimension is a $d$-dimensional vector for learning the pairwise force and a one-dimensional scalar for the potential energy.
{Besides the baseline, we compare the performance of \pignpi to an alternative method proposed by \cite{lemos2021rediscovering} that is also based on GN and was specifically designed to infer pairwise forces. We denote this method as \lemos and the details are introduced in Sec.~\ref{sec:method_lemos2021rediscovering}}.

We split each dataset into training, validation and testing datasets and use  $\mathcal{T}_{\text{train}}$, $\mathcal{T}_{\text{valid}}$ and $\mathcal{T}_{\text{test}}$ to indicate the corresponding simulation time steps for these different splits. Details regarding the dataset generation are provided in Sec.~\ref{sec:experiment_setting}. The baseline algorithm and \pignpi are trained and evaluated on the same training and testing datasets from simulations with an 8-particle system. Further, the evaluation of the generalization ability uses a 12-particle system. 

{It should be noted that \cite{cranmer2020discovering} measure the quality of the learnt forces by quantifying the linear correlation between each dimension of the learnt edge message and all dimensions of the ground-truth pairwise force. This is a necessary but not sufficient condition to claim the correspondence of the learnt edge message with the pairwise interactions and to evaluate the performance of the indirect inference of the pairwise interactions. }
Instead, we evaluate the proposed methodology with a focus on two key aspects: 1) supervised learning performance, and 2) consistency with underlying physics. For all the evaluations, the mean absolute error on the testing dataset of various particle and interaction properties is used and is defined as follows
\begin{equation}
    \textsf{MAE}^{\mathrm{inter}} (\hat{\phi}, \phi) = \frac{1}{|\mathcal{T}_{\text{test}}|} \frac{1}{|E|}  \sum_{t\in \mathcal{T}_{\text{test}}}\sum_{i,j}^{i\neq j} l_1(\hat{\phi}_{ij}^t,\phi_{ij}^t) ~,
\end{equation}
and
\begin{equation}
    \textsf{MAE}^{\mathrm{part}} (\hat{\phi}, \phi) = \frac{1}{|\mathcal{T}_{\text{test}}|} \frac{1}{|V|} \sum_{t \in \mathcal{T}_{\text{test}}} \sum_{i=1}^{|V|} l_1(\hat{\phi}_i^t,\phi_i^t)
    \label{eq:mae_particle}
\end{equation}
respectively, where the superscript hat indicates the predicted values. Here, $\hat{\phi}_{ij}$ and ${\phi}_{ij}$ are the predicted and corresponding ground-truth, respectively, of a physical quantity between particle $j$ and particle $i$ (\textit{e.g.}, pairwise force), and $\hat{\phi}_{i} = \sum_j \hat{\phi}_{ij}$ and ${\phi}_{i} = \sum_j {\phi}_{ij}$ are the aggregated prediction and the corresponding ground-truth, respectively, on particle $i$ (\textit{e.g.}, net force). $l_1(x, y)$ computes the sum of absolute differences between each element in $x$ and $y$, $l_1(x, y) = \sum_i\abs{x_i-y_i}$, if $x$ and $y$ are vectors or the absolute difference, $l_1(x, y)=\abs{x-y}$, if $x$ and $y$ are scalars. Hence, $\textsf{MAE}^{\mathrm{part}}$ measures the averaged error of the physical quantity on particles over $\mathcal{T}_{\text{test}}$, and  $\textsf{MAE}^{\mathrm{inter}}$ is the averaged error of the inter-particle physical quantity over $\mathcal{T}_{\text{test}}$.

The supervised learning performance is evaluated on the prediction of the acceleration \textsf{MAE\textsubscript{acc}}$= \textsf{MAE}^{\mathrm{part}}(\bm{\hat{\ddot{r}}},\bm{\ddot{r}})$. The true acceleration values serve as target values during training. The physical consistency is evaluated on two criteria. First, we evaluate the ability of the proposed framework to infer the underlying physical quantities that were not used as the target during training (\textit{e.g.},  pairwise force), and second, we evaluate physical consistency by verifying whether Newton's action-reaction property is satisfied.

The following metrics are used to evaluate the consistency with the true pairwise interaction.
For pairwise force, we use \textsf{MAE\textsubscript{ef}}$= \textsf{MAE}^{\mathrm{inter}}(\bm{\hat{F}},\bm{F})$;
and for potential energy case, we evaluate the increment in potential energy \textsf{MAE\textsubscript{$\Delta$ep} }  $= \textsf{MAE}^\mathrm{inter} (\hat{P}-\hat{P}^0, P-P^0)$, 
where superscript $0$ refers to the initial configuration.

For the second part of the evaluation of the physical consistency, we verify whether Newton's action-reaction property is satisfied. For that, we evaluate the symmetry in either inter-particle forces with
$\textsf{MAE}_\textsf{symm}^{F}= \frac{1}{|\mathcal{T}_{\text{test}}|} \frac{1}{|E|}  \sum_{t\in \mathcal{T}_{\text{test}}}\sum_{i,j}^{i\neq j} l_1(\bm{\hat{F}}_{ij}^t, - \bm{\hat{F}}_{ji}^t)$
or inter-particle potential with 
$\textsf{MAE}_\textsf{symm}^{P}= \frac{1}{|\mathcal{T}_{\text{test}}|} \frac{1}{|E|}  \sum_{t\in \mathcal{T}_{\text{test}}}\sum_{i,j}^{i\neq j} l_1(\hat{P}_{ij}^t, \hat{P}_{ji}^t)$.


\begin{figure}[t!]
    \centering
    \includegraphics[width=1.0\linewidth]{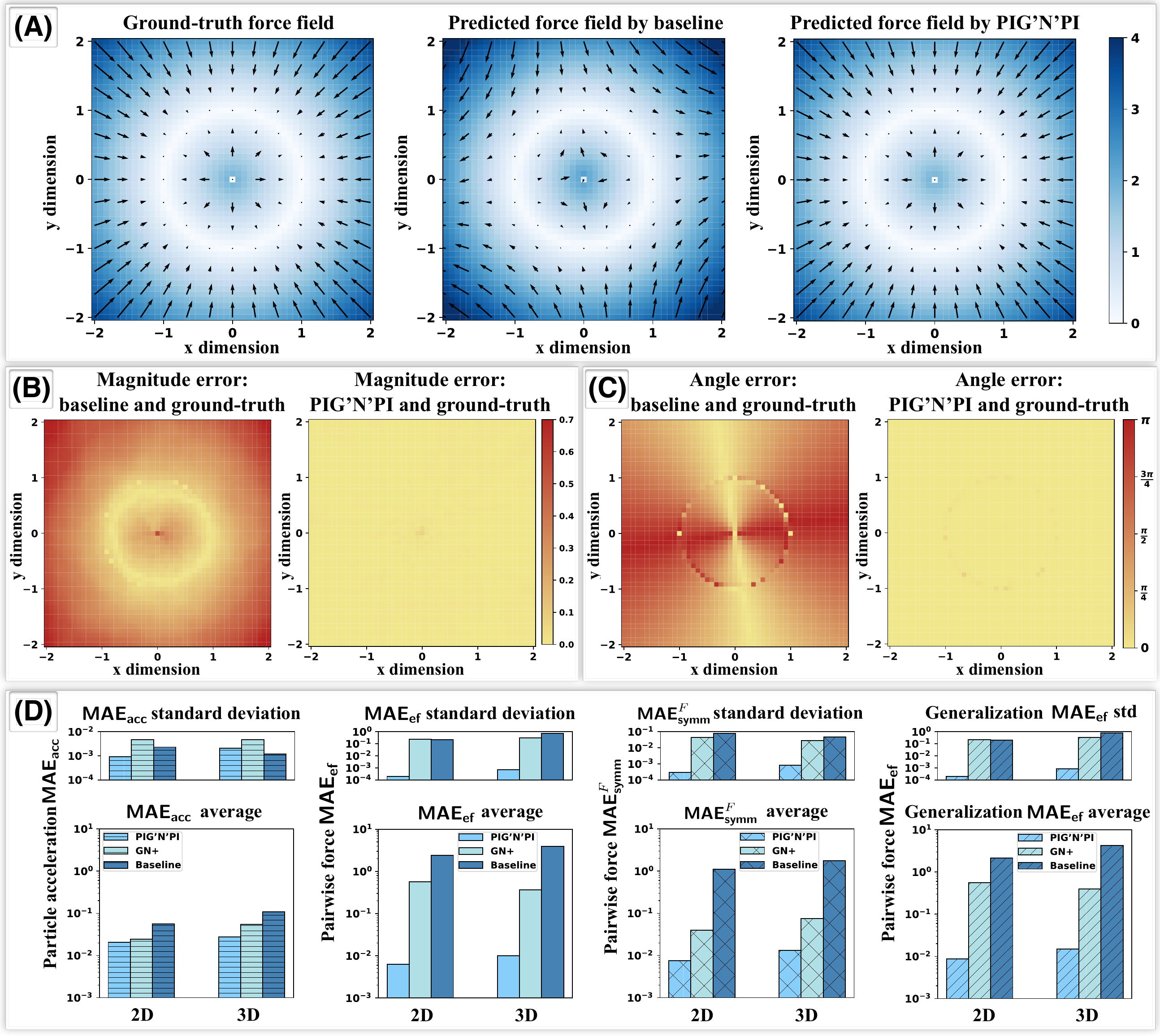}
    \caption{\textbf{Case study: Quality of pairwise force prediction of \pignpi and the baseline model on two-dimensional spring dataset. } (A) The spring force field around a given particle. Color indicates the force amplitude. From left to right: ground-truth spring force field, predicted force field by the baseline model, predicted force field by \pignpi. (B) The magnitude error between predicted force and the ground-truth force ($\abs{\text{norm}(\bm{\hat{F}})-\text{norm}(\bm{F})}$). Left is the result of baseline model and right is the result of \pignpi. (C) The angle difference between  predicted force and the ground-truth force ($\text{Angle}(\bm{\hat{F}}, \bm{F})$, in radian). Left is the result of baseline model and right is the result of \pignpi.  (D) Comparison of the quality of \pignpi, \lemos and baseline model on learning the pairwise force, where bottom is average result of five experiments and top is the corresponding standard deviation. From left to right (in logarithmic scale): acceleration error \textsf{MAE\textsubscript{acc}}, pairwise force error \textsf{MAE\textsubscript{ef}}, force symmetry error $\textsf{MAE}_\textsf{symm}^{F}$ and pairwise force error \textsf{MAE\textsubscript{ef}} on generalization dataset.}
    \label{fig:spring_force_results}
\end{figure}

\subsection{Performance evaluation between \pignpi{, \lemos} and baseline for pairwise force}
\label{sec:results_pignpi_vs_baseline_force}

First, we analyse \pignpi for application on particle systems with interactions given by pairwise forces. We start with evaluating the supervised learning performance by evaluating the prediction of the acceleration using \textsf{MAE\textsubscript{acc}}. The results show that \pignpi provides slightly better predictions than the {\lemos and the} baseline model for both the spring dataset (see Fig.~\ref{fig:spring_force_results}D) and all other datasets (see \APPENDIX~\ref{sec:force-potential-test-results}).
 
\begin{figure}[h!]
    \centering
    \includegraphics[width=1.0\linewidth]{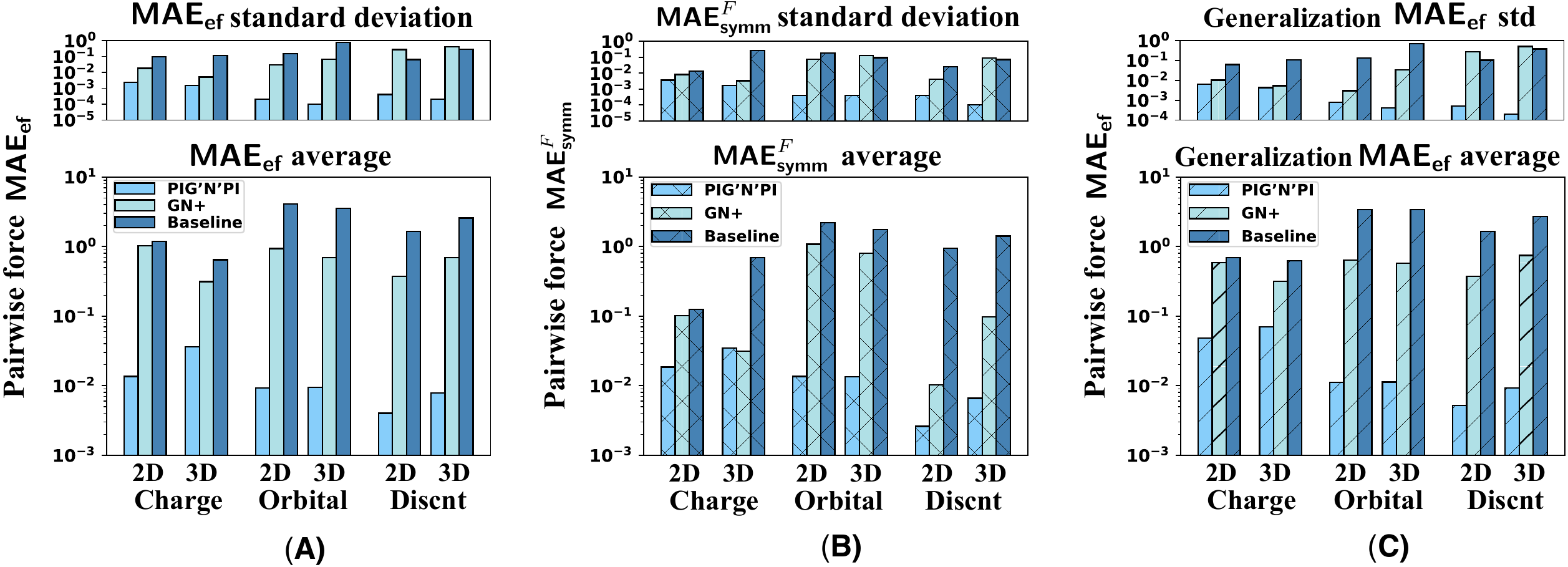}
    \caption{\textbf{Quality of pairwise force prediction of \pignpi, the \lemos and the baseline model.} We report the average and standard deviation of different errors with five experiments, in logarithmic scale. (A) Pairwise force prediction error \textsf{MAE\textsubscript{ef}}. (B) Pairwise force symmetry error $\textsf{MAE}_\textsf{symm}^{F}$. (C) Pairwise force error \textsf{MAE\textsubscript{ef}} on generalization dataset.}
    \label{fig:force_summary}
\end{figure}

To verify the physical consistency, we first evaluate if the implicitly inferred pairwise forces are consistent with the true physical quantity. \pignpi is able to infer the  force field around a particle correctly, while the baseline model fails to predict the force field (see Fig.~\ref{fig:spring_force_results}A for the spring dataset). A force field needs to be precise in both amplitude and direction. The error of the magnitude (see Fig.~\ref{fig:spring_force_results}B) and angle (see Fig.~\ref{fig:spring_force_results}C) demonstrate unambiguously the superior performance of \pignpi compared to the baseline model. We quantitatively summarize the  performance of the pairwise force inference with \textsf{MAE\textsubscript{ef}}, which shows \pignpi is able to infer the pairwise force correctly, while both {\lemos and the} baseline models fail to infer the pairwise force ($2-3$ orders of magnitude worse inference performance for the spring dataset (see Fig.~\ref{fig:spring_force_results}D) and all other datasets (see Fig.~\ref{fig:force_summary}A and \APPENDIX~\ref{sec:force-potential-test-results}). 

Secondly, we verify the consistency of the implicitly inferred pairwise forces with  Newton's action-reaction law by evaluating the symmetry of the inter-particle forces with
$\textsf{MAE}_\textsf{symm}^{F}$. Our results demonstrate that \pignpi satisfies the symmetry property. However, {\lemos and the} baseline model are not able to satisfy the underlying Newton's laws for the spring dataset (see Fig.~\ref{fig:spring_force_results}D) and the other datasets (see Fig.~\ref{fig:force_summary}B and \APPENDIX~\ref{sec:force-potential-test-results}).

Furthermore, we test the robustness of \pignpi and the baseline model to learn from noisy data. We impose noise to the measured positions and then compute the noisy velocities (first-order derivative of position) and noisy accelerations (second-order derivative of position). The noisy accelerations serve then as the target values for the learning tasks of all the models. The performance of \pignpi decreases with increasing noise level (see \APPENDIX~\ref{sec:noisy_input_experiments}). This is to be expected given that adding noise makes the training target (particle accelerations) less similar to the uncorrupted target that is associated with particle interactions. However, \pignpi can still learn reasonably well the particle interactions despite the corrupted data. The performance of the baseline model fluctuates, however, with different noise levels significantly. This is due to the fact that the baseline model does not learn the particle interactions but rather the particle kinematics and is, therefore, more sensitive to noise.

Finally, we note that the proposed algorithm is also able to generalize well when trained on an eight-particle system and applied to a 12-particle system for all datasets (see Fig.~\ref{fig:spring_force_results}D, Fig.~\ref{fig:force_summary} and Table~\ref{table:transfer_generalization_force}). 

Overall, the results demonstrate that the proposed algorithm learns correctly the pairwise force (that is consistent with the underlying physics) without any direct supervision, \textit{i.e.}, without access to the pairwise force in the first place, and that the inferred forces are consistent with the imposed underlying physical laws.

\subsection{Performance evaluation between \pignpi and baseline for pairwise potential energy}
\label{sec:results_pignpi_vs_baseline_potential}

\begin{figure}[h!]
    \centering
    \includegraphics[width=1.0\linewidth]{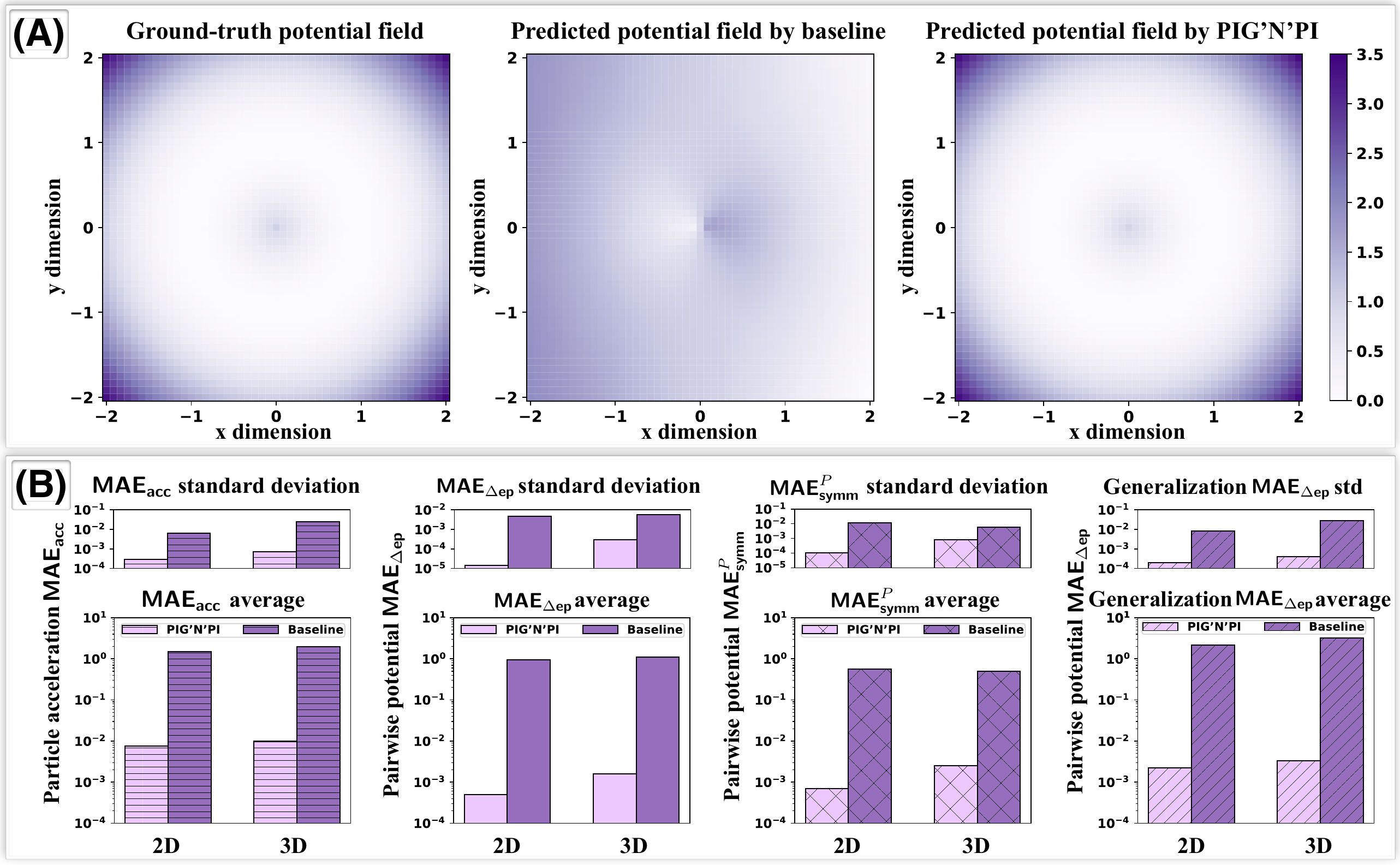}
    \caption{\textbf{Case study: Quality of pairwise potential prediction of \pignpi and the baseline model on Spring dataset. } (A) The spring potential field around a given particle. Color indicates the potential amplitude. From left to right: ground-truth spring potential field, predicted potential field by the baseline model, predicted potential field by \pignpi. (B) Comparison of the quality of \pignpi and baseline model on learning the pairwise force. From left to right (in logarithmic scale): acceleration error \textsf{MAE\textsubscript{acc}}, pairwise potential error \textsf{MAE\textsubscript{$\Delta$ep}}, potential symmetry error $\textsf{MAE}_\textsf{symm}^{P}$ and pairwise potential error \textsf{MAE\textsubscript{$\Delta$ep}} on generalization dataset.}
    \label{fig:spring_potential_result}
\end{figure}

\begin{figure}[h!]
    \centering
    \includegraphics[width=1.0\linewidth]{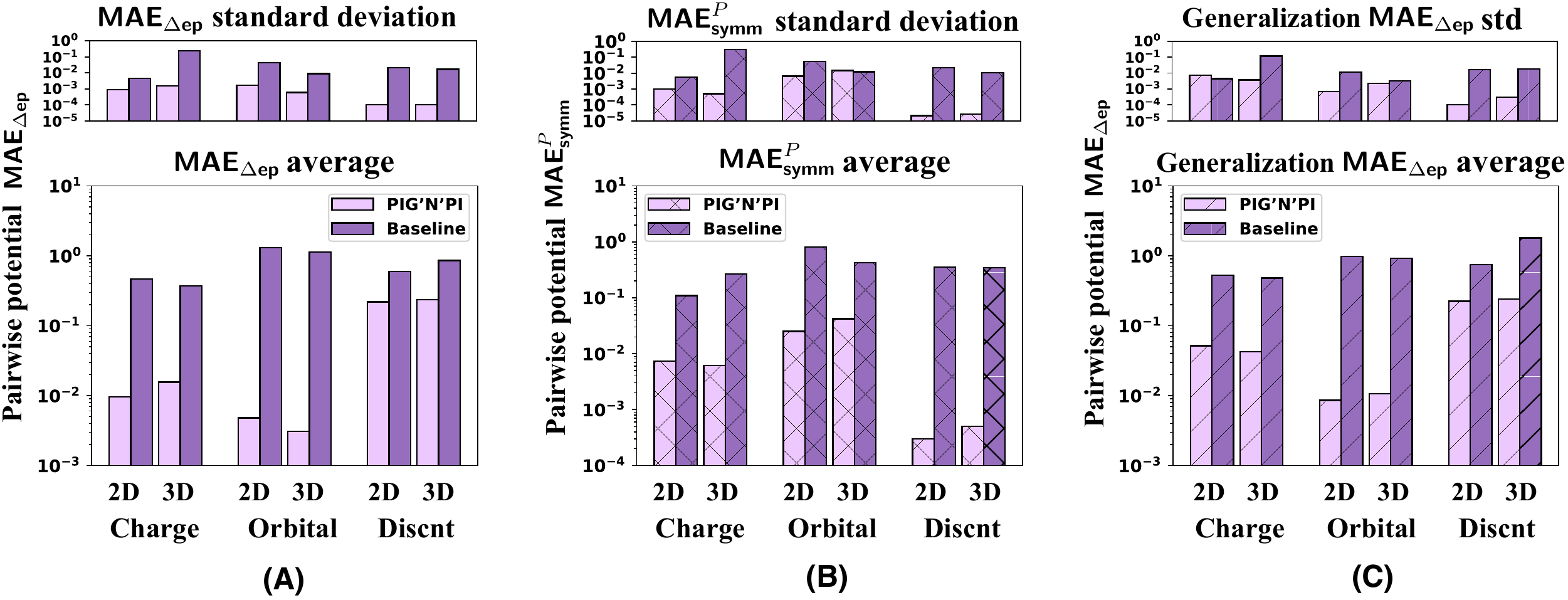}
    \caption{\textbf{Quality of pairwise potential prediction of \pignpi and the baseline model.} We report different errors in terms of consistency with the underlying physical laws, in logarithmic scale. (A) Pairwise potential incremental error \textsf{MAE\textsubscript{$\Delta$ep}}. (B) Pairwise potential symmetry error $\textsf{MAE}_\textsf{symm}^{P}$. (C) Pairwise potential error incremental error \textsf{MAE\textsubscript{$\Delta$ep}} on generalization dataset.}
    \label{fig:potential_summary}
\end{figure}

Besides learning the pairwise force, the proposed methodology is extended to learn the pairwise potential energy (see node part design for learning potential in Sec.~\ref{sec:architecture}). In this case, the physics operator computes the pairwise force via partial derivative.
{Since \lemos was solely designed for learning the pairwise force, it is not possible to apply it to infer the pairwise potential. Therefore, for the task of pairwise potential energy inference, we only compare \pignpi with the baseline model.}
Our results show that \pignpi performs well in the supervised learning of the acceleration (\textsf{MAE\textsubscript{acc}}). Here again, its performance is considerably better compared to the baseline model (see Fig.~\ref{fig:spring_potential_result}B). Moreover, the performance is similar to that in the force-based version of the algorithm. 

However, when comparing the performance of the baseline model on the supervised learning task  between the potential-based version and the force-based version of the model, the performance reduces significantly in the potential-based implementation (compare to Fig.~\ref{fig:spring_force_results}D). 
This drop of performance is potentially explained by the adjustment of the output dimension of the edge neural network in the baseline model to enable the extraction of the potential energy.  

Further, our results demonstrate a superior performance of the \pignpi algorithm on consistency with underlying physics. Firstly, it infers well the increment of the potential energy (see Fig.~\ref{fig:spring_potential_result}A). It clearly infers the increment of the potential field correctly. On the contrary, the baseline model fails to infer the potential energy. This can be quantitatively assessed with \textsf{MAE\textsubscript{$\Delta$ep}}. The results (see Fig.~\ref{fig:spring_potential_result}B and \ref{fig:potential_summary}) show that \pignpi is able to infer the potential energy, while the performance of the baseline model indicates that it is not able to learn the potential energy only from observing the particles movement. 

It is important to note that our algorithm cannot predict the absolute value of the potential energy; only the increment (see \APPENDIX~\ref{sec:discontinuous_potential_prediction}). The reason is that the model is trained on the acceleration, which is computed from the derivative of the potential energy (\textit{i.e.}, the force). Hence, the model only constrains the derivative, and the constant of integration remains unknown. This limitation can be overcome by one of the two following options:  either the potential energy is constrained by a spatial boundary condition or by an initial condition. In the former, we can impose a known value for a given value of $r_{ij}$, \textit{e.g.}, we could use the assumption that the potential energy for a charge interaction approaches zero with increasing particle distance. The alternative (but less likely) approach consists in knowing the potential energy at a given time, \textit{e.g.}, at the beginning of the observation and add the inferred increment to the known initial value. Nevertheless, knowing the absolute value of the potential energy is, in fact, not crucial as only its derivative determines the dynamics of a particle system. This is also confirmed in our experiments by the accurate prediction of the acceleration (see \textsf{MAE\textsubscript{acc}}).

Secondly, similar to the pairwise force prediction, \pignpi also provides a superior performance on satisfying Newton's action-reaction property, while again, the baseline model fails to satisfy the underlying Newton's law. The performance is quantified by the symmetry of the inter-particle potential energies ($\textsf{MAE}_\textsf{symm}^{P}$).  (Fig.~\ref{fig:spring_potential_result}B, Fig.~\ref{fig:potential_summary} and Table~\ref{table:potential-result}).


Finally, we test the generalization ability of the learning algorithms in a similar way as in the pairwise force case study. We apply the  models trained on an eight-particles system to a new particle system comprising 12 particles. The results (see \APPENDIX~\ref{sec:genralization-test-results}) show that \pignpi predicts well the pairwise force and potential energy, and outperforms considerably the baseline model (see Fig.~\ref{fig:spring_potential_result}B and Fig.~\ref{fig:potential_summary}). This demonstrates that the \pignpi model provides a general model for learning particle interactions. 

\subsection{Case study under more realistic conditions: learning pairwise interactions for a LJ-argon system}
\label{sec:results_pignpi_vs_baseline_LJ_argon}
To evaluate the performance of the proposed framework on a more realistic system with a larger particle interaction system (to evaluate the scalability), we apply \pignpi on a large Lennard-Jones system.

We adopt the dataset introduced in \cite{li2022graph}.  This dataset simulates the movements of liquid argon atoms governed by the Lennard-Jones (LJ) potential. The LJ potential, which is given by $V(r) = 4\epsilon \{(\sigma/r)^12 - (\sigma/r)^6\}$, is an extensively used governing law for two non-bonding atoms~\cite{rapaport2004art}. The simulation contains 258 particles in a cubic box whose length is 27.27~\r{A}. The simulation is run at 100~K with periodic boundary conditions. The potential well depth $\epsilon$ is set to 0.238 kilocalorie / mole, the van der Waals radius $\sigma$ is set to 3.4 \r{A}, and the interaction cutoff radius for the argon atoms is set to $3\sigma$. The mass of argon atom is 39.9~dalton. The dataset is run for 10 independent simulations. Each simulation contains 1000 time steps with randomly initialized positions and velocities. The position, velocity and acceleration of all particles are recorded at each time step. 

\begin{figure}[h!]
    \centering
    \includegraphics[width=1.0\linewidth]{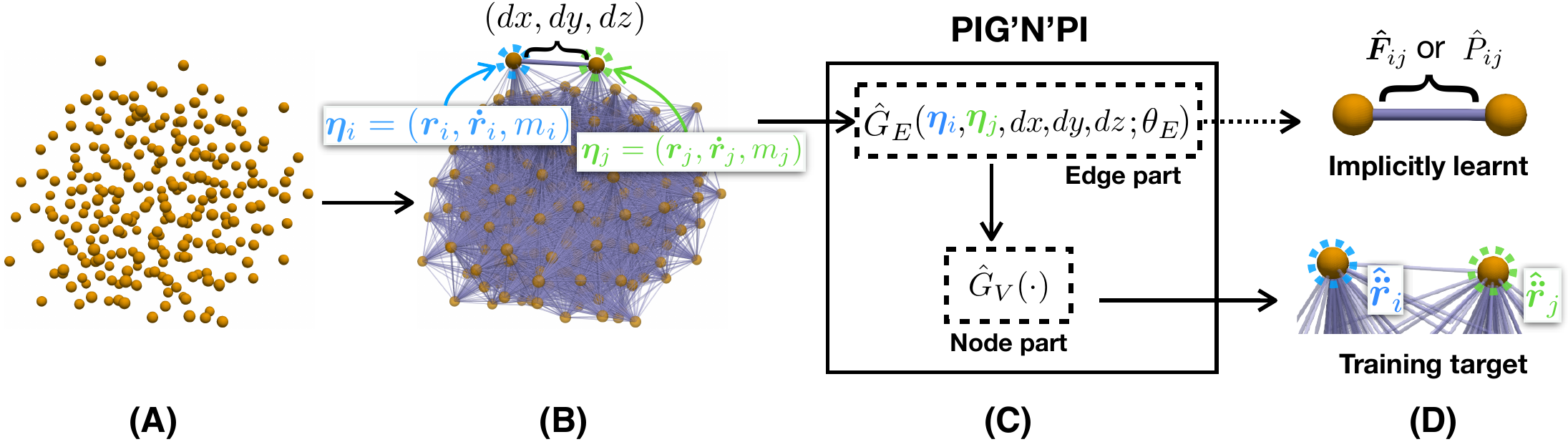}
    \caption{\textbf{Pipeline of \pignpi to learn pairwise force or potential for the LJ-argon particle data. Solid-line arrows indicate the data processing path from the input to the output. The dash-line arrow depicts the intermediate output of every edge corresponding to the inferred pairwise force or potential energy.} (A) Positions of 258 particles at a random time step. (B) Representation of the constructed graph. Node features comprise position, velocity and mass. Edge features comprise the relative position difference under periodic boundary conditions. (C) \pignpi. Edge part takes the concatenation of two nodes' features and the edge feature as input and infers the pairwise force or potential. Node part aggregates the output on every edge and predicts the acceleration. (D) The inferred pairwise force or potential by edge part and the acceleration by node part.}
    \label{fig:LJ-argon-pipeline}
\end{figure}

Fig.~\ref{fig:LJ-argon-pipeline} summarizes the learning pipeline. Contrary to the previous case study where for a small number of particles, a fully connected graph is considered, in this case study, we construct the graph of neighboring particles at every time step (Fig.~\ref{fig:LJ-argon-pipeline}). We connect the particles within the defined interaction cutoff radius while taking the periodic boundary conditions into consideration. Particles in the LJ-argon system are characterized by their position, velocity and mass. The charge is not part of the particle properties, which is different from the particle systems considered in the previous case study (Sec.~\ref{sec:results_pignpi_vs_baseline_force} and Sec.~\ref{sec:results_pignpi_vs_baseline_potential}). Moreover, we compute the position difference under the periodic boundary condition and use it as an edge feature. This edge feature is required because the distance between two particles in this simulation does not correspond to the Euclidean distance in the real world due to the periodic boundary conditions. The node features and the edge features are then concatenated and are used as the input to the edge part of \pignpi (Fig.~\ref{fig:LJ-argon-pipeline}C) or the baseline model. Similarly to the previous case study, the learning target is the accelerations of the particles. The pairwise force and pairwise potential energy are then inferred from the intermediate output of the edge part.

\begin{figure}[h!]
    \centering
    \includegraphics[width=1.0\linewidth]{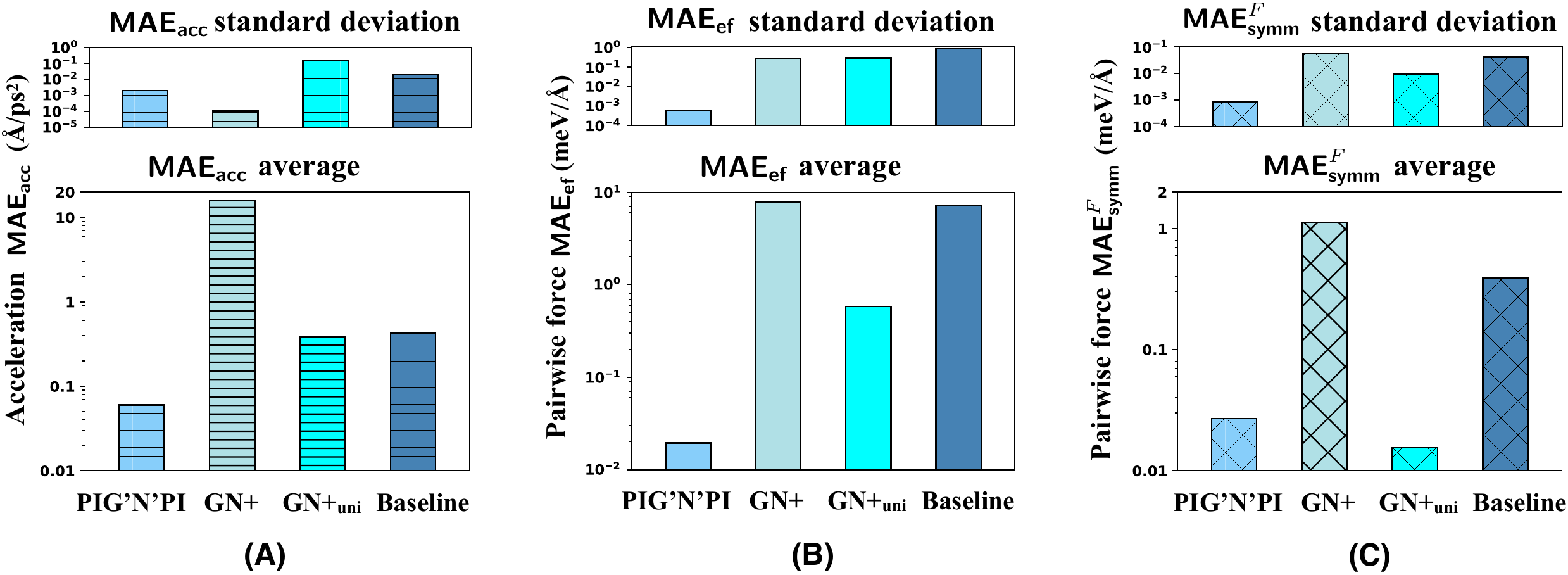}
    \caption{\textbf{Performance of the algorithms on pairwise force predictions on the LJ-argon system.} We report the MAE on the acceleration prediction, which is the target for the learning task  (A), the MAE on the pairwise force inference (indirect inference task) \textsf{MAE\textsubscript{ef}} (B) and the consistency with Newton's action-reaction property: the MAE on pairwise force symmetry $\textsf{MAE}_\textsf{symm}^{F}$ (C). The average (plots at the bottom) on logarithmic scale and standard deviation (plots in the top row) are computed from five experiments.}
    \label{fig:LJ-argon-force}
\end{figure}

We evaluate the performance of \pignpi on inferring pairwise interactions of the LJ-argon particle system with the same performance metrics as in the previous case study. The results are reported in Fig.~\ref{fig:LJ-argon-force} and Table~\ref{table:LJ-argon-force}. {Because the particles in this dataset have the same mass, we also test a variant of \lemos such that we assign all nodes with a unique learnable scalar. We denote this variant as GN+\textsubscript{uni}.} The results confirm the very good performance of \pignpi as observed in the previous case study. {Generally, GN+\textsubscript{uni} outperforms the \lemos, but \pignpi still surpasses GN+\textsubscript{uni} and the baseline.}  On the one hand \pignpi performs better than the baseline{, \lemos, and GN+\textsubscript{uni}} on the supervised prediction task of predicting the acceleration (achieving less than half of the \textsf{MAE\textsubscript{acc}} compared to the baseline {and the GN+\textsubscript{uni}} (Fig.~\ref{fig:LJ-argon-force}A)). On the other hand, \pignpi is also able to infer the learn pairwise force correctly. Again, the baseline model is not able to infer the pairwise force (\pignpi outperforms the baseline by more than two orders of the magnitude on the \textsf{MAE\textsubscript{ef}}).      Moreover, the  particle interactions inferred by \pignpi are consistent with Newton’s action-reaction law ($\textsf{MAE}_\textsf{symm}^{F}$). The bad performance of the baseline model indicates that the learnt interactions do not satisfy the Newton's law.  

To summarize, similar to the cases discussed in Sec.~\ref{sec:results_pignpi_vs_baseline_force}, \pignpi learns the pairwise force well without any direct supervision for this complex and large system.

\begin{figure}[h!]
    \centering
    \includegraphics[width=1.0\linewidth]{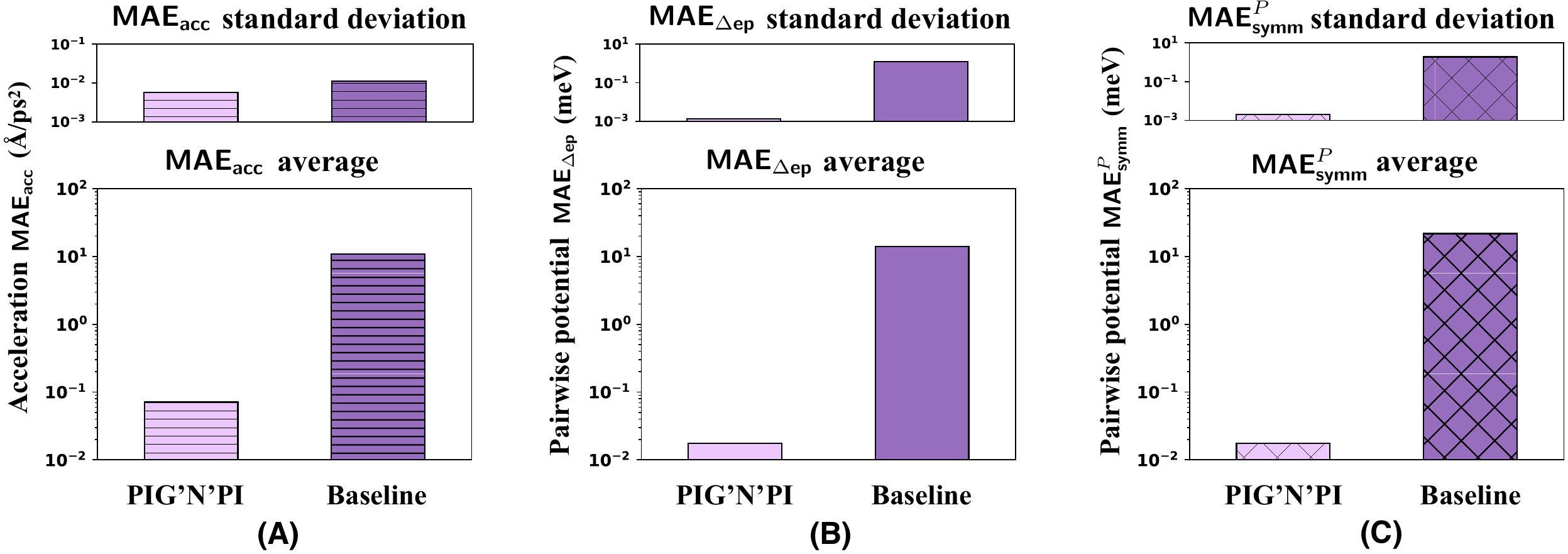}
    \caption{\textbf{Quality of pairwise potential prediction on the LJ-argon data.} We report different errors in \emph{logarithmic} scale. The average and standard deviation are computed from five experiments. (A) Acceleration prediction error \textsf{MAE\textsubscript{acc}}. (B) Pairwise potential incremental error \textsf{MAE\textsubscript{$\Delta$ep}}. (C) Pairwise potential symmetry error $\textsf{MAE}_\textsf{symm}^{P}$.}
    \label{fig:LJ-argon-potential}
\end{figure}

Besides, we test \pignpi to learn the pairwise potential energy for this LJ system. Results are reported in Fig.~\ref{fig:LJ-argon-potential} and Table~\ref{table:LJ-argon-potential}. We first examine the \textsf{MAE\textsubscript{acc}} that is the learning target. The \textsf{MAE\textsubscript{acc}} of \pignpi is similar to that in the force-based version of the algorithm.  \pignpi performs significantly better than the baseline model with more than 2 orders of magnitude (Fig.~\ref{fig:LJ-argon-potential}A). And, similar to the cases in Sec.~\ref{sec:results_pignpi_vs_baseline_potential}, we again observe the performance drop of the baseline model in this potential-based version with the force-based version. Then, we evaluate the \textsf{MAE\textsubscript{$\Delta$ep}} and \textsf{MAE\textsubscript{ef}} that are the two metrics for measuring the quality of the learnt pairwise potential energy. Results show that \pignpi provides again superior performance on inferring the increment of the potential energy \textsf{MAE\textsubscript{$\Delta$ep}} (Fig.~\ref{fig:LJ-argon-potential}B). The bad performance of the baseline model clearly shows that it is not able to infer the potential energy correctly. Finally, we check Newton’s action-reaction property in the potential energy by $\textsf{MAE}_\textsf{symm}^{P}$. Here again, the potential energy inferred by \pignpi follows Newton's laws while the baseline model fails to infer the underlying interaction laws correctly (Fig.~\ref{fig:LJ-argon-potential}D). All evaluations demonstrate that the predicted pairwise potential energy by \pignpi is consistent with the LJ potential used in the simulation, even though \pignpi does not access the ground truth information.

The results on this case study demonstrate the scalability of \pignpi to larger systems and the applicability to more realistic case studies. Moreover, the results confirm the results obtained in the previous case study that \pignpi is able to infer the underlying interaction laws correctly. While the other baseline methods are able to precisely predict the particle dynamics, they fail to infer the underlying pairwise interaction laws correctly.

\subsection{Comparison of \pignpi to alternative hyperparameter choices and an alternative regularized architecture}
\label{sec:results_pignpi_vs_alternatives}
We compare the performance of the proposed approach first to alternative hyperparameter choices, in particular to different activation functions, and, second, to an alternative way of imposing physical consistency in the network architecture. 


First, we evaluate different choices of activation functions following the observations made in previous studies~\cite{hu2021forcenet, niaki2021physics, klicpera2020directional} that confirmed their significant influence on the performance of MLPs in approximating physical quantities. The performance of \pignpi with different activation functions is  reported in \APPENDIX~\ref{sec:activation_function_force}. The results demonstrate that \pignpi with SiLU activation function (which was in fact used in all case studies) performs consistently best on most test datasets compared to \pignpi with other commonly used activation functions, such as ReLU or LeakyReLU. Based on this observation, the performance of the baseline with the SiLU activation function was evaluated (\APPENDIX~\ref{sec:force-potential-test-results}). The results show that the  SiLU activation improves the learning performance of the baseline model to some degree (when only evaluating the prediction performance \textsf{MAE\textsubscript{acc}}). However, it still performs consistently worse than \pignpi and, more importantly, the consistency with underlying physics in terms of the inferred force (or potential) and interaction symmetry worsens even considerably.

Second, we compare the performance of \pignpi to an  alternative way of imposing physical consistency: we add a regularization into the baseline model to enforce the symmetry property onto the output messages of the edge function. The goal of imposing the symmetry regularization term is to ensure that the model satisfies the action-reaction physical consistency requirement. It is expected that by satisfying this symmetry constraint, the model performance on learning physics-consistent pairwise forces and potential energy can be improved. We add a symmetry regularization term on the learnt pairwise corresponding messages to enforce the action-reaction property. The details on this regularization term can be found in Sec.~\ref{sec:method_symmetry_regularization}.

While the performance is improved compared to the baseline model without any regularization, the results demonstrate that \pignpi still performs considerably better on inferring  physical meaningful quantities for the pairwise force and potential energy than the symmetry-regularized baseline model  
(see \APPENDIX~\ref{sec:symmetry_for_baseline} for detailed results and \APPENDIX~\ref{sec:LJ-argon-evaluation} for the evaluation on the LJ-argon system).

\section{Conclusions}
\label{sec:conclusions}
In this paper, we propose the Physics-induced Graph Network for Particle Interaction algorithm to learn  particle interactions that are consistent with the underlying physical laws. The main novelty of the proposed algorithm is in the design of the physics operator in the node part. The designed physics operator on nodes guides the edge neural network to learn the pairwise force or pairwise potential energy exactly. This design also reduces the model complexity for this machine learning algorithm by reducing the number of tunable parameters.

While our method shows a similar performance on the supervised learning task of predicting accelerations compared to the other baseline models (purely data-driven graph networks), it is also able to infer the particle interactions correctly that follow the underlying physical laws. However, while the baseline models are able to be applied as simulators, they fail to infer physically consistent particle interactions that satisfy Newton's laws. Moreover, the proposed \pignpi also outperforms the other baseline models in terms of generalization ability to larger systems and robustness to significant levels of noise. 

The proposed methodology can generalize well to larger particle systems. However, we have to point out that the trained model cannot extrapolate the data arbitrarily far from the training distribution. In our experiments, we found that the edge neural network converges to linear functions outside the training input space. This observation matches the discussion in \cite{xu2021how}, which is an inherent limitation of MLPs.



The developed methodology will help to make a step forward in developing a flexible and robust tool for the discovery of physical laws in material mechanics. Such tools will be able to support, for example, additive manufacturing with heterogeneous materials that are particularly subject to highly varying material properties, \eg, sustainable or recycled materials \cite{wang2021inverse}.

\section{Methods}
\label{sec:method}

\subsection{Notations and formal task description}
\label{sec:notation_task_description}
We use a fully-connected directed graph $G=(V, E)$ to represent the interacting particle system, where nodes $V = \{v_1, v_2, \ldots, v_{|V|}\}$  correspond to the particles and the directed edges $E = \{e_{ij}: v_i, v_j \in V, i\neq j\}$ correspond to particle interactions. Under this notation, $v_i$ refers to the $i$-th particle, and $e_{ij}$ is the directed edge from $v_j$ to $v_i$. We use $\{\bm{\eta}^t_i\}_{i, t}$ to denote the observation of particle states at different time steps, where $\bm{\eta}^t_i$ is a vector describing the state of particle $v_i$ at time $t$. We note that $\bm{\eta}^t_i \in \mathbb{R}^{2d+2}$ ($d$ is the space dimension) includes {position} $\bm{r}^t_i \in \mathbb{R}^d$, {velocity} $\bm{\dot{r}}^t_i \in \mathbb{R}^d$, {electric charge} $q_i \in \mathbb{R}$ and {mass} $m_i \in \mathbb{R}$. The {velocity} $\bm{\dot{r}}^t_i$ and {acceleration} $\bm{\ddot{r}}_i^t$ at time $t$ are computed from the position series of particle $v_i$. We use $\mathcal{M}_{ij}$ to denote the message from $v_j$ to $v_i$ learnt by the neural network $\hat{G}_E(\cdot; \theta_{E})$ with parameters $\theta_{E}$. 
Our goal is to infer the pairwise force $F_{ij}^t$ and the potential energy $P_{ij}^t$ on every edge $e_{ij}$ at each time $t$ given the observation of particle trajectories.


\subsection{PIG’N’PI details}
\label{sec:method_overview}
\pignpi contains an edge part to learn the pairwise interaction and a node part to aggregate the interactions to derive node accelerations (see Fig.~\ref{fig:framework}). In the edge part, we use MLPs as universal approximators~\cite{hornik1989multilayer, hornik1991approximation} to learn the pairwise force or pairwise potential energy. We denote this edge neural network as $\hat{G}_E(\cdot; \theta_{E})$. $\hat{G}_E(\cdot; \theta_{E})$ takes the vectors $\bm{\eta}_i$ and $\bm{\eta}_j$ of two nodes as input. The output $\mathcal{M}_{ij}$ of $\hat{G}_E(\cdot; \theta_{E})$ is the inferred pairwise force $\bm{\hat{F}}_{ij}$ or potential energy $\hat{P}_{ij}$ on edge $e_{ij}$, depending on the operator in the node part. We design the physics operator $G_N(\cdot)$ to aggregate the edge messages in the node part and derive the acceleration $\bm{\hat{\ddot{r}}}^t_i$ for every particle $v_i$ at time $t$. We optimize parameters $\theta_{E}$ by minimizing the mean absolute error between the predicted acceleration and the true acceleration. The objective function is given by:

\begin{equation}
\label{eq:objective_func}
    \arg\min_{\theta_{E}} \mathcal{L} = \frac{1}{|\mathcal{T}_{\text{train}}|} \frac{1}{|V|} \sum_{t \in \mathcal{T}_{\text{train}}} \sum_{i=1}^{|V|} l_1(\bm{\hat{\ddot{r}}}_i^t,\bm{\ddot{r}}_i^t)
\end{equation}

In the following, we explain the design of the edge neural network $\hat{G}_E(\cdot; \theta_{E})$ and the node part operator $G_N(\cdot)$ in two cases: inferring the pairwise force and inferring the pairwise potential energy.


\textbf{Learning pairwise force}

We use an MLP as the edge neural network $\hat{G}_E(\cdot; \theta_{E})$ to learn the pairwise force from $v_j$ to $v_i$ on each edge $e_{ij}$. The output dimension of $\hat{G}_E(\cdot; \theta_{E})$ is the same as the spatial dimension $d$.  We first concatenate $\bm{\eta}_i^t$ and $\bm{\eta}_j^t$ which is the input of $\hat{G}_E(\cdot; \theta_{E})$. We denote the corresponding output as $\mathcal{M}_{ij}^t \in \mathbb{R}^{d}$, \textit{e.g.}, 
\begin{equation}
\label{eq:force_edge_operator}
    \mathcal{M}_{ij}^t \triangleq \hat{G}_E(\text{concat}(\bm{\eta}_i^t, \bm{\eta}_j^t); \theta_{E})
\end{equation}

According to Newton's Second law, the net acceleration of every particle is equal to the net force divided by its mass. Hence, in the node part of \pignpi, we first sum up all incoming messages $\mathcal{M}_{i} = \sum_j^{j\neq i}\mathcal{M}_{ij}$ of every particle $v_i$, and then divide it by the mass of the particle $m_i$. The output of $G_N(\cdot)$ is the predicted acceleration $\bm{\hat{\ddot{r}}}_i$ on particle $v_i$:
\begin{equation}
\label{eq:force_node_operator}
	\begin{split}
		\bm{\hat{\ddot{r}}}_i^t & = G_N(\bm{\eta}_i^t, \textstyle \mathcal{M}_{i}^t) \\
		& = G_N(\bm{\eta}_i^t, \textstyle \sum_{j}^{j \neq i} \mathcal{M}_{ij}^t) \\
        & = \frac{\textstyle \sum_{j}^{j \neq i} \mathcal{M}_{ij}^t}{m_i}\\
    \end{split}
\end{equation}

We optimize the parameters $\theta_{E}$ in $\hat{G}_E(\cdot; \theta_{E})$ by minimizing the objective function Eq.~(\ref{eq:objective_func}). Through this process, the node part  operator $G_N(\cdot)$ guides the edge neural network $\hat{G}_E(\cdot; \theta_{E})$ to predict the pairwise force exactly, \textit{e.g.},  
\begin{equation}
    \label{eq:force-prediction-forceCase}
    \bm{\hat{F}}_{ij}^t = \mathcal{M}_{ij}^t
\end{equation}
{This is illustrated in Block (B) of Fig.~\ref{fig:method_illustration}.}

\textbf{Learning pairwise potential energy}

For the pairwise potential energy case, the edge neural network $\hat{G}_E(\cdot; \theta_{E})$ is designed to output the pairwise potential energy. Here, the output dimension of $\hat{G}_E(\cdot; \theta_{E})$ is one because the potential energy is a scalar. We still first concatenate $\bm{\eta}_i^t$ and $\bm{\eta}_j^t$ as the input of $\hat{G}_E(\cdot; \theta_{E})$ and use MLPs as $\hat{G}_E(\cdot; \theta_{E})$. The corresponding output $\mathcal{M}_{ij}^t \in \mathbb{R}$ is denoted as:
\begin{equation}
\label{eq:potential_edge_operator}
    \mathcal{M}_{ij}^t \triangleq \hat{G}_E(\text{concat}(\bm{\eta}_i^t, \bm{\eta}_j^t); \theta_{E})
\end{equation}

We know that the net force of every particle equals to the negative partial derivative of the potential energy with respect to its position. Hence, in the node part, we first sum up all incoming messages $\mathcal{M}_{i} = \sum_j^{j\neq i}\mathcal{M}_{ij}$ for every particle $i$, then compute the negative derivative with respect to the input position and finally divide it by the mass. The final output corresponds then to the predicted acceleration. The node part operator $G_N(\cdot)$ for the potential energy case is given by:
\begin{equation}
\begin{split}
\label{eq:potential_node_operator}
	\bm{\hat{\ddot{r}}}_i^t & = G_N(\bm{\eta}_i^t, \textstyle \mathcal{M}_{i}^t) \\
	& = G_N(\bm{\eta}_i^t, \textstyle \sum_{j}^{j\neq i} \mathcal{M}_{ij}^t) \\
	& = - \frac{ \partial (\sum_{j}^{j \neq i} \mathcal{M}_{ij}^t)/ \partial \bm{r}_i^t }{m_i}
\end{split}
\end{equation}

Analogously to the force-based case, we optimize for the parameters $\theta_{E}$ in $\hat{G}_E(\cdot; \theta_{E})$ by minimizing the acceleration loss (Eq.~(\ref{eq:objective_func})). The node part  operator $G_N(\cdot)$ here guides the edge neural network $\hat{G}_E(\cdot; \theta_{E})$ to learn the pairwise potential energy exactly. The learnt message on each edge corresponds to the predicted pairwise potential energy, and the negative partial derivative is the predicted pairwise force, \textit{e.g.},
\begin{equation}
\begin{split}
\label{eq:potential-force-prediction-potentialCase}
    \hat{P}_{ij} &= \mathcal{M}_{ij}\\
    \bm{\hat{F}}_{ij} &= - \partial \hat{P}_{ij} / \partial \bm{r}_{i} = - \partial \mathcal{M}_{ij} / \partial \bm{r}_{i}
\end{split}
\end{equation}
{This is illustrated in Block (C) of Fig.~\ref{fig:method_illustration}.}


We note that the commonly used ReLU activation function is not suitable as activation function in $\hat{G}_E(\cdot; \theta_{E})$ for learning the potential energy. The reason is that we compute the partial derivative of $\mathcal{M}_{ij} = \hat{G}_E(\text{concat}(\bm{\eta}_i, \bm{\eta}_j); \theta_{E})$ to derive the predicted accelerations for every particle. The derivative should be continuous and even smooth considering physical forces. However, ReLU approximates the underlying function by piece-wise linear hyper-planes with sharp boundaries. The first-order derivative is, thus, piece-wise constant that does not change with input~\cite{hu2021forcenet}.
Details on selecting the activation function in $\hat{G}_E(\cdot; \theta_{E})$ are explained in Sec.~\ref{sec:experiment_setting}.

\subsection{Details on imposing a symmetry regularization on the baseline model}
\label{sec:method_symmetry_regularization}
As mentioned in Sec.~\ref{sec:results_pignpi_vs_alternatives}, to ensure that the model satisfies the action-reaction physical consistency requirement, we also test an extension of the baseline model by imposing a symmetry regularization on the corresponding pairwise messages in the baseline model. This can be considered as an alternative way of imposing physical consistency. In detail, let $\mathcal{M}_{ij}$ be the message from $v_{j}$ to $v_{i}$ which is the output of the edge neural network of the baseline model. In our experimental setup, the message $\mathcal{M}_{ij}$ corresponds to the force from $v_{j}$ to $v_{i}$. We impose the symmetry regularization by adding a regularization term on the learnt messages in the objective function (Eq.~(\ref{eq:objective_func})). This results in the following objective function:
\begin{equation}
    \arg\min_{\theta_{E}} \mathcal{L} = \frac{1}{|\mathcal{T}_{\text{train}}|} \sum_{t \in \mathcal{T}_{\text{train}}} \left (\underbrace{\frac{1}{|V|}\sum_{i=1}^{|V|} \abs{\bm{\hat{\ddot{r}}}_i^t - \bm{\ddot{r}}_i^t}}_{\text{Acceleration loss on nodes}} + \underbrace{\alpha \frac{1}{|E|}\sum_{i,j}^{i\neq j}\abs{\mathcal{M}_{ij}^t + \mathcal{M}_{ji}^t}}_{\text{Symmetry regularization loss on edges}} \right )
    \label{eq:symmetry_regularization_objective}
\end{equation}
where $\alpha$ is a weight parameter. The original baseline model can be considered as the special case with $\alpha=0$ in Eq.~(\ref{eq:symmetry_regularization_objective}). In our experiments, we evaluate the impact of the regularization term with different weights  ($\alpha=0.1$, $1.0$, $10$, $100$). The results are reported in \APPENDIX~Table~\ref{tab:symmetry_regularization_result}.

\subsection{Details of the method to learn pairwise force introduced by \cite{lemos2021rediscovering}}
\label{sec:method_lemos2021rediscovering}

\citet{lemos2021rediscovering} proposed a method that has a similar goal to the proposed \pignpi applied to pairwise force prediction. The authors also impose Newton's second law in the standard GN block by dividing the aggregated messages by the node property. We denote this method as \lemos. The main difference between \lemos and \pignpi is that \lemos treats the node property as a learnable parameter. It assigns an individual learnable scalar $w_i$ for each particle $v_i$ and predicts the acceleration of $v_i$ by dividing the aggregated incoming messages by $10^{w_i}$. The learnable scalars on all nodes representing the pairwise force are learnt together with all other parameters. It is important to point out that \lemos was designed solely for learning the pairwise force while \pignpi can be applied both: to infer the pairwise forces and also the pairwise potential energy. The detailed results of \lemos are reported in Table~\ref{table:pairwise-force-result}, Table~\ref{table:transfer_generalization_force}, Table~\ref{table:LJ-argon-force}, Fig.~\ref{fig:spring_force_results}D, Fig.~\ref{fig:force_summary} and Fig.~\ref{fig:LJ-argon-force}.


\subsection{Details about simulation and experiments}
\label{sec:experiment_setting}

Here, we summarize the different force functions used in our simulation. Please note that in this work, we used the same case studies as in previous work~\cite{cranmer2020discovering}. However, we adapted the parameters of the particle systems slightly to make the learning more challenging. 
\begin{itemize}
    \item \textbf{Spring force} We denote the spring constant as $k$ and balance length as $L$. The pairwise force from $v_i$ to $v_j$ is $k(r_{ij} - L)\bm{n}_{ij}$ and its potential energy is $0.5k(r_{ij} - L)^2$, where $r_{ij} = \norm{\bm{r}_j - \bm{r}_i}$ is the Euclidean distance and $\bm{n}_{ij}= \frac{\bm{r}_j - \bm{r}_i}{\norm{\bm{r}_j - \bm{r}_i}}$ is the unit vector pointing from $v_i$ to $v_j$. We set $k=2.0$ and $L=1.0$ in our simulations.
    \item \textbf{Charge force} The electric charge force from $v_i$ to $v_j$ is $-c q_i q_j \bm{n}_{ij} / r_{ij}^2$ and the potential energy is $c q_i q_j / r_{ij} $, where $c$ is the charge constant, and $q_i, q_j$ are the electric charges. We set $c=1.0$ in the simulation. Furthermore, to prevent any zeros in the denominator, we add a small number $\delta$ ($\delta=0.01$) when computing distances.
    \item \textbf{Orbital force} The orbital force from $v_i$ to $v_j$ equals to $m_i m_j\bm{n}_{ij}/r_{ij}$ and the potential energy is $m_i m_j \text{ln}(r_{ij})$, where $m_i, m_j$ are the masses of $v_i$ and $v_j$. We again add a small number $\delta$ ($\delta=0.01$) when computing distances to prevent zeros in the denominator and logarithm.
    \item \textbf{Discontinuous force} We set threshold constant $\Theta  =2.0$ such that the pairwise force is $\bm{0}$ if the Euclidean distance $r_{ij}$ is strictly smaller than this threshold and $(r_{ij} - 1)\bm{n}_{ij}$ otherwise. The corresponding potential is 0 if $r_{ij}$ is strictly smaller than this threshold and $0.5(r_{ij} - 1)^2$ otherwise. 
\end{itemize}

We intentionally omit units for variables because the simulation data can be at arbitrary scale. Moreover, the presented cases serve as proof of concept to learn the input–output relation.
Further, we note that $m_i$ is sampled from the log-uniform distribution within the range $[- 1]$ ($\text{ln}(m_i) \sim \mathcal{U}(-1, 1)$). $q_i$ is uniformly sampled from the range $[-1, 1]$. Initial location and velocity of particles are both sampled from the normal Gaussian distribution $\mathcal{N}(0, 1)$. 
Each simulation contains eight particles. Each particle is associated with the corresponding features including position, velocity, mass and charge. The target for prediction is node accelerations. Every simulation contains 10,000 time steps with step size $0.01$. We randomly split the simulation steps into training dataset, validation dataset and testing dataset with the ratio $7:1.5:1.5$. We use $\mathcal{T}_{\text{train}}$, $\mathcal{T}_{\text{valid}}$ and $\mathcal{T}_{\text{test}}$ to indicate the simulation time steps corresponding to training split, validation split and testing split.  We train the model on the training dataset (optimizing the parameters $\theta_{E}$ in $\hat{G}_E( \cdot ; \theta_{E} )$) by optimizing Eq.~(\ref{eq:objective_func}), fine-tune
hyperparameters and select the best trained model on the validation dataset and report the performance of the selected trained model on the testing dataset. For generalization tests, we re-run each simulation on 12 particles with 1500 time steps (same size as original testing dataset). The previously selected trained model with eight particles is tested on the new testing dataset.

We only fine-tune hyperparameters on the spring validation dataset and use the same hyperparameters in all experiments. We set the learning rate to  $0.001$, the number of hidden layers in the edge neural network  to four, the units of hidden layers to $300$, max training epochs to $200$. The dimension of the output layer in the edge neural network is $d$ to learn the force or one to learn the potential energy. We use the Adam optimizer with the mini-batch size of 32 for the force case study and eight for the potential  case study to train the model. 
The SiLU activation function is used in all \pignpi evaluations.  



\section{Data availability}
The data used in the experiments are generated by the numerical simulator. All data used for the experiments are included in the associated Gitlab repository: \url{https://gitlab.ethz.ch/cmbm-public/pignpi/-/tree/main/simulation}.

\section{Code availability}
The implementation of \pignpi is based on PyTorch~\cite{paszke2017automatic} and pytorch-geometric~\cite{fey2019fast} libraries. The source code is available on Gitlab: \url{https://gitlab.ethz.ch/cmbm-public/pignpi}.

\section{Acknowledgements}
{This project has been funded by the ETH Grant office. The contribution of Olga Fink to this project was funded by the Swiss National Science Foundation (SNSF) Grant no. PP00P2\_176878.}

\section{Author contributions}
{DK and OF conceptualized the idea, ZH developed the methodology, ZH, DK and OF conceived the experiments, ZH conducted the experiments, ZH, DK and OF analysed the results, and all authors wrote the manuscript.}

\section{Competing interests.}
{The authors declare no competing interest.}

\bibliographystyle{unsrtnat}
\bibliography{ref}

\begin{thebibliography}{42}
\providecommand{\natexlab}[1]{#1}
\providecommand{\url}[1]{\texttt{#1}}
\expandafter\ifx\csname urlstyle\endcsname\relax
  \providecommand{\doi}[1]{doi: #1}\else
  \providecommand{\doi}{doi: \begingroup \urlstyle{rm}\Url}\fi

\bibitem[Murray and Dermott(1999)]{murray1999solar}
Carl~D Murray and Stanley~F Dermott.
\newblock \emph{Solar system dynamics}.
\newblock Cambridge university press, 1999.

\bibitem[Dramis(2016)]{dramis2016mass}
Francesco Dramis.
\newblock Mass movement processes and landforms.
\newblock \emph{International Encyclopedia of Geography: People, the Earth,
  Environment and Technology: People, the Earth, Environment and Technology},
  pages 1--9, 2016.

\bibitem[Sawyer et~al.(2014)Sawyer, Argibay, Burris, and
  Krick]{sawyer2014mechanistic}
W~Gregory Sawyer, Nicolas Argibay, David~L Burris, and Brandon~A Krick.
\newblock Mechanistic studies in friction and wear of bulk materials.
\newblock \emph{Annual Review of Materials Research}, 44:\penalty0 395--427,
  2014.

\bibitem[Furlani(2010)]{furlani2010magnetic}
Edward~P Furlani.
\newblock Magnetic biotransport: analysis and applications.
\newblock \emph{Materials}, 3\penalty0 (4):\penalty0 2412--2446, 2010.

\bibitem[Wang and G{\'o}mez-Bombarelli(2019)]{wang2019coarse}
Wujie Wang and Rafael G{\'o}mez-Bombarelli.
\newblock Coarse-graining auto-encoders for molecular dynamics.
\newblock \emph{npj Computational Materials}, 5\penalty0 (1):\penalty0 1--9,
  2019.

\bibitem[Angioletti-Uberti(2017)]{angioletti2017theory}
Stefano Angioletti-Uberti.
\newblock Theory, simulations and the design of functionalized nanoparticles
  for biomedical applications: a soft matter perspective.
\newblock \emph{npj Computational Materials}, 3\penalty0 (1):\penalty0 1--15,
  2017.

\bibitem[Radovic et~al.(2018)Radovic, Williams, Rousseau, Kagan, Bonacorsi,
  Himmel, Aurisano, Terao, and Wongjirad]{radovic2018machine}
Alexander Radovic, Mike Williams, David Rousseau, Michael Kagan, Daniele
  Bonacorsi, Alexander Himmel, Adam Aurisano, Kazuhiro Terao, and Taritree
  Wongjirad.
\newblock Machine learning at the energy and intensity frontiers of particle
  physics.
\newblock \emph{Nature}, 560\penalty0 (7716):\penalty0 41--48, 2018.

\bibitem[Carleo et~al.(2019)Carleo, Cirac, Cranmer, Daudet, Schuld, Tishby,
  Vogt-Maranto, and Zdeborov{\'a}]{carleo2019machine}
Giuseppe Carleo, Ignacio Cirac, Kyle Cranmer, Laurent Daudet, Maria Schuld,
  Naftali Tishby, Leslie Vogt-Maranto, and Lenka Zdeborov{\'a}.
\newblock Machine learning and the physical sciences.
\newblock \emph{Reviews of Modern Physics}, 91\penalty0 (4):\penalty0 045002,
  2019.

\bibitem[Zhou et~al.(2020)Zhou, Cui, Hu, Zhang, Yang, Liu, Wang, Li, and
  Sun]{zhou2020graph}
Jie Zhou, Ganqu Cui, Shengding Hu, Zhengyan Zhang, Cheng Yang, Zhiyuan Liu,
  Lifeng Wang, Changcheng Li, and Maosong Sun.
\newblock Graph neural networks: A review of methods and applications.
\newblock \emph{AI Open}, 1:\penalty0 57--81, 2020.

\bibitem[Shlomi et~al.(2020)Shlomi, Battaglia, and Vlimant]{shlomi2020graph}
Jonathan Shlomi, Peter Battaglia, and Jean-Roch Vlimant.
\newblock Graph neural networks in particle physics.
\newblock \emph{Machine Learning: Science and Technology}, 2\penalty0
  (2):\penalty0 021001, 2020.

\bibitem[Atz et~al.(2021)Atz, Grisoni, and Schneider]{atz2021geometric}
Kenneth Atz, Francesca Grisoni, and Gisbert Schneider.
\newblock Geometric deep learning on molecular representations.
\newblock \emph{Nature Machine Intelligence}, 3\penalty0 (12):\penalty0
  1023--1032, 2021.

\bibitem[M{\'e}ndez-Lucio et~al.(2021)M{\'e}ndez-Lucio, Ahmad, del Rio-Chanona,
  and Wegner]{mendez2021geometric}
Oscar M{\'e}ndez-Lucio, Mazen Ahmad, Ehecatl~Antonio del Rio-Chanona, and
  J{\"o}rg~Kurt Wegner.
\newblock A geometric deep learning approach to predict binding conformations
  of bioactive molecules.
\newblock \emph{Nature Machine Intelligence}, 3\penalty0 (12):\penalty0
  1033--1039, 2021.

\bibitem[Park et~al.(2021)Park, Kornbluth, Vandermause, Wolverton, Kozinsky,
  and Mailoa]{park2021accurate}
Cheol~Woo Park, Mordechai Kornbluth, Jonathan Vandermause, Chris Wolverton,
  Boris Kozinsky, and Jonathan~P. Mailoa.
\newblock Accurate and scalable graph neural network force field and molecular
  dynamics with direct force architecture.
\newblock \emph{npj Computational Materials}, 7\penalty0 (1):\penalty0 73,
  2021.

\bibitem[Sanchez-Gonzalez et~al.(2020)Sanchez-Gonzalez, Godwin, Pfaff, Ying,
  Leskovec, and Battaglia]{sanchez2020learning}
Alvaro Sanchez-Gonzalez, Jonathan Godwin, Tobias Pfaff, Rex Ying, Jure
  Leskovec, and Peter Battaglia.
\newblock Learning to simulate complex physics with graph networks.
\newblock In \emph{International Conference on Machine Learning (ICML)}, pages
  8459--8468. PMLR, 2020.

\bibitem[Matsumoto et~al.(2021)Matsumoto, Ishida, Araki, Kato, Terayama, and
  Okuno]{matsumoto2021extraction}
Shigeyuki Matsumoto, Shoichi Ishida, Mitsugu Araki, Takayuki Kato, Kei
  Terayama, and Yasushi Okuno.
\newblock Extraction of protein dynamics information from cryo-em maps using
  deep learning.
\newblock \emph{Nature Machine Intelligence}, 3\penalty0 (2):\penalty0
  153--160, 2021.

\bibitem[Sch{\"u}tt et~al.(2017)Sch{\"u}tt, Kindermans, Sauceda, Chmiela,
  Tkatchenko, and M{\"u}ller]{schutt2017schnet}
Kristof~T Sch{\"u}tt, Pieter-Jan Kindermans, Huziel~E Sauceda, Stefan Chmiela,
  Alexandre Tkatchenko, and Klaus-Robert M{\"u}ller.
\newblock Schnet: A continuous-filter convolutional neural network for modeling
  quantum interactions.
\newblock In \emph{Proceedings of the 31st International Conference on Neural
  Information Processing Systems (NeurIPS)}, page 992–1002, 2017.

\bibitem[Sch{\"u}tt et~al.(2018)Sch{\"u}tt, Sauceda, Kindermans, Tkatchenko,
  and M{\"u}ller]{schutt2018schnet}
Kristof~T Sch{\"u}tt, Huziel~E Sauceda, P-J Kindermans, Alexandre Tkatchenko,
  and K-R M{\"u}ller.
\newblock Schnet--a deep learning architecture for molecules and materials.
\newblock \emph{The Journal of Chemical Physics}, 148\penalty0 (24):\penalty0
  241722, 2018.

\bibitem[Kipf et~al.(2018)Kipf, Fetaya, Wang, Welling, and
  Zemel]{kipf2018neural}
Thomas Kipf, Ethan Fetaya, Kuan-Chieh Wang, Max Welling, and Richard Zemel.
\newblock Neural relational inference for interacting systems.
\newblock In \emph{International Conference on Machine Learning (ICML)}, pages
  2688--2697. PMLR, 2018.

\bibitem[Unke and Meuwly(2019)]{unke2019physnet}
Oliver~T Unke and Markus Meuwly.
\newblock Physnet: A neural network for predicting energies, forces, dipole
  moments, and partial charges.
\newblock \emph{Journal of chemical theory and computation}, 15\penalty0
  (6):\penalty0 3678--3693, 2019.

\bibitem[Klicpera et~al.(2020{\natexlab{a}})Klicpera, Gro{\ss}, and
  G{\"u}nnemann]{klicpera2020directional}
Johannes Klicpera, Janek Gro{\ss}, and Stephan G{\"u}nnemann.
\newblock Directional message passing for molecular graphs.
\newblock In \emph{International Conference on Learning Representations
  (ICLR)}, 2020{\natexlab{a}}.

\bibitem[Klicpera et~al.(2020{\natexlab{b}})Klicpera, Giri, Margraf, and
  G{\"u}nnemann]{klicpera2020fast}
Johannes Klicpera, Shankari Giri, Johannes~T Margraf, and Stephan
  G{\"u}nnemann.
\newblock Fast and uncertainty-aware directional message passing for
  non-equilibrium molecules.
\newblock In \emph{Machine Learning for Molecules Workshop @ NeurIPS},
  2020{\natexlab{b}}.

\bibitem[Bapst et~al.(2020)Bapst, Keck, Grabska-Barwi{\'n}ska, Donner, Cubuk,
  Schoenholz, Obika, Nelson, Back, Hassabis, et~al.]{bapst2020unveiling}
Victor Bapst, Thomas Keck, A~Grabska-Barwi{\'n}ska, Craig Donner, Ekin~Dogus
  Cubuk, Samuel~S Schoenholz, Annette Obika, Alexander~WR Nelson, Trevor Back,
  Demis Hassabis, et~al.
\newblock Unveiling the predictive power of static structure in glassy systems.
\newblock \emph{Nature Physics}, 16\penalty0 (4):\penalty0 448--454, 2020.

\bibitem[Hu et~al.(2021)Hu, Shuaibi, Das, Goyal, Sriram, Leskovec, Parikh, and
  Zitnick]{hu2021forcenet}
Weihua Hu, Muhammed Shuaibi, Abhishek Das, Siddharth Goyal, Anuroop Sriram,
  Jure Leskovec, Devi Parikh, and C~Lawrence Zitnick.
\newblock Forcenet: A graph neural network for large-scale quantum
  calculations.
\newblock \emph{arXiv preprint arXiv:2103.01436}, 2021.

\bibitem[Cranmer et~al.(2020)Cranmer, Sanchez~Gonzalez, Battaglia, Xu, Cranmer,
  Spergel, and Ho]{cranmer2020discovering}
Miles Cranmer, Alvaro Sanchez~Gonzalez, Peter Battaglia, Rui Xu, Kyle Cranmer,
  David Spergel, and Shirley Ho.
\newblock Discovering symbolic models from deep learning with inductive biases.
\newblock In \emph{Advances in Neural Information Processing Systems
  (NeurIPS)}, volume~33, pages 17429--17442, 2020.

\bibitem[Battaglia et~al.(2018)Battaglia, Hamrick, Bapst, Sanchez-Gonzalez,
  Zambaldi, Malinowski, Tacchetti, Raposo, Santoro, Faulkner,
  et~al.]{battaglia2018relational}
Peter~W Battaglia, Jessica~B Hamrick, Victor Bapst, Alvaro Sanchez-Gonzalez,
  Vinicius Zambaldi, Mateusz Malinowski, Andrea Tacchetti, David Raposo, Adam
  Santoro, Ryan Faulkner, et~al.
\newblock Relational inductive biases, deep learning, and graph networks.
\newblock \emph{arXiv preprint arXiv:1806.01261}, 2018.

\bibitem[Greydanus et~al.(2019)Greydanus, Dzamba, and
  Yosinski]{greydanus2019hamiltonian}
Samuel Greydanus, Misko Dzamba, and Jason Yosinski.
\newblock Hamiltonian neural networks.
\newblock In \emph{Advances in Neural Information Processing Systems
  (NeurIPS)}, volume~32, 2019.

\bibitem[Sanchez-Gonzalez et~al.(2019)Sanchez-Gonzalez, Bapst, Cranmer, and
  Battaglia]{sanchez2019hamiltonian}
Alvaro Sanchez-Gonzalez, Victor Bapst, Kyle Cranmer, and Peter Battaglia.
\newblock Hamiltonian graph networks with ode integrators.
\newblock In \emph{Machine Learning and the Physical Sciences Workshop @
  NeurIPS}, 2019.

\bibitem[Lutter et~al.(2019)Lutter, Ritter, and Peters]{lutter2019deep}
Michael Lutter, Christian Ritter, and Jan Peters.
\newblock Deep lagrangian networks: Using physics as model prior for deep
  learning.
\newblock In \emph{International Conference on Learning Representations
  (ICLR)}, 2019.

\bibitem[Wang et~al.(2019)Wang, Olsson, Wehmeyer, P{\'e}rez, Charron,
  De~Fabritiis, No{\'e}, and Clementi]{wang2019machine}
Jiang Wang, Simon Olsson, Christoph Wehmeyer, Adri{\`a} P{\'e}rez, Nicholas~E
  Charron, Gianni De~Fabritiis, Frank No{\'e}, and Cecilia Clementi.
\newblock Machine learning of coarse-grained molecular dynamics force fields.
\newblock \emph{ACS central science}, 5\penalty0 (5):\penalty0 755--767, 2019.

\bibitem[Finzi et~al.(2020)Finzi, Wang, and Wilson]{finzi2020simplifying}
Marc Finzi, Ke~Alexander Wang, and Andrew~G Wilson.
\newblock Simplifying hamiltonian and lagrangian neural networks via explicit
  constraints.
\newblock In \emph{Advances in neural information processing systems
  (NeurIPS)}, volume~33, pages 13880--13889, 2020.

\bibitem[Karniadakis et~al.(2021)Karniadakis, Kevrekidis, Lu, Perdikaris, Wang,
  and Yang]{karniadakis2021physics}
George~Em Karniadakis, Ioannis~G Kevrekidis, Lu~Lu, Paris Perdikaris, Sifan
  Wang, and Liu Yang.
\newblock Physics-informed machine learning.
\newblock \emph{Nature Reviews Physics}, 3\penalty0 (6):\penalty0 422--440,
  2021.

\bibitem[Goodfellow et~al.(2016)Goodfellow, Bengio, and
  Courville]{goodfellow2016deep}
Ian Goodfellow, Yoshua Bengio, and Aaron Courville.
\newblock \emph{Deep learning}.
\newblock MIT press, 2016.

\bibitem[Lemos et~al.(2021)Lemos, Jeffrey, Cranmer, Battaglia, and
  Ho]{lemos2021rediscovering}
Pablo Lemos, Niall Jeffrey, Miles Cranmer, Peter Battaglia, and Shirley Ho.
\newblock Rediscovering newton’s gravity and solar system properties using
  deep learning and inductive biases.
\newblock In \emph{SimDL Workshop @ ICLR}, 2021.

\bibitem[Li et~al.(2022)Li, Meidani, Yadav, and Barati~Farimani]{li2022graph}
Zijie Li, Kazem Meidani, Prakarsh Yadav, and Amir Barati~Farimani.
\newblock Graph neural networks accelerated molecular dynamics.
\newblock \emph{The Journal of Chemical Physics}, 156\penalty0 (14):\penalty0
  144103, 2022.

\bibitem[Rapaport and Rapaport(2004)]{rapaport2004art}
Dennis~C Rapaport and Dennis C~Rapaport Rapaport.
\newblock \emph{The art of molecular dynamics simulation}.
\newblock Cambridge university press, 2004.

\bibitem[Niaki et~al.(2021)Niaki, Haghighat, Campbell, Poursartip, and
  Vaziri]{niaki2021physics}
Sina~Amini Niaki, Ehsan Haghighat, Trevor Campbell, Anoush Poursartip, and Reza
  Vaziri.
\newblock Physics-informed neural network for modelling the thermochemical
  curing process of composite-tool systems during manufacture.
\newblock \emph{Computer Methods in Applied Mechanics and Engineering},
  384:\penalty0 113959, 2021.

\bibitem[Xu et~al.(2021)Xu, Zhang, Li, Du, Kawarabayashi, and
  Jegelka]{xu2021how}
Keyulu Xu, Mozhi Zhang, Jingling Li, Simon~Shaolei Du, Ken-Ichi Kawarabayashi,
  and Stefanie Jegelka.
\newblock How neural networks extrapolate: From feedforward to graph neural
  networks.
\newblock In \emph{International Conference on Learning Representations
  (ICLR)}, 2021.

\bibitem[Wang and Zhang(2021)]{wang2021inverse}
Qi~Wang and Longfei Zhang.
\newblock Inverse design of glass structure with deep graph neural networks.
\newblock \emph{Nature Communications}, 12\penalty0 (1):\penalty0 5359, 2021.

\bibitem[Hornik et~al.(1989)Hornik, Stinchcombe, and
  White]{hornik1989multilayer}
Kurt Hornik, Maxwell Stinchcombe, and Halbert White.
\newblock Multilayer feedforward networks are universal approximators.
\newblock \emph{Neural networks}, 2\penalty0 (5):\penalty0 359--366, 1989.

\bibitem[Hornik(1991)]{hornik1991approximation}
Kurt Hornik.
\newblock Approximation capabilities of multilayer feedforward networks.
\newblock \emph{Neural networks}, 4\penalty0 (2):\penalty0 251--257, 1991.

\bibitem[Paszke et~al.(2017)Paszke, Gross, Chintala, Chanan, Yang, DeVito, Lin,
  Desmaison, Antiga, and Lerer]{paszke2017automatic}
Adam Paszke, Sam Gross, Soumith Chintala, Gregory Chanan, Edward Yang, Zachary
  DeVito, Zeming Lin, Alban Desmaison, Luca Antiga, and Adam Lerer.
\newblock Automatic differentiation in pytorch.
\newblock In \emph{Workshop on Autodiff @ NeurIPS}, 2017.

\bibitem[Fey and Lenssen(2019)]{fey2019fast}
Matthias Fey and Jan~E. Lenssen.
\newblock Fast graph representation learning with {PyTorch Geometric}.
\newblock In \emph{Representation Learning on Graphs and Manifolds Workshop @
  ICLR}, 2019.

\end{thebibliography}

\appendix

\clearpage
\section{Supplementary information}

\subsection{Symbol table}
The variable notations used in the paper are summarized in Table~\ref{tab:symbol_notations}.
\begin{table}[h!]
\centering
\caption{Symbol notations and their meanings}
\label{tab:symbol_notations}

{ \small
\begin{tabularx}{\textwidth}{|c X|} 
 \hline
    \textbf{notation} & \textbf{meaning}\\[0.75ex]
\hline \rule{0pt}{2ex}
$G=(V, E)$ & graph representation of the interacting particle system \Tstrut\Bstrut \\
\hline
$V = \{v_1, v_2, \ldots, v_{|V|}\}$ &  set of nodes corresponding to particles  \Tstrut\Bstrut\\
\hline
$E = \{e_{ij}: v_i, v_j \in V, i\neq j\}$ & set of edges  corresponding to interactions between particles  \Tstrut\Bstrut\\
\hline
$v_i \in V$ & $i$-th particle  \Tstrut\Bstrut\\
\hline
$e_{ij} \in E$ & directed edge from particle $v_j$ to particle $v_i$  \Tstrut\Bstrut\\
\hline
$d$ & spatial dimension (2 or 3)  \Tstrut\Bstrut\\
\hline
$\bm{r}_i^t \in \mathbb{R}^d$ & position of $v_i$ at time $t$  \Tstrut\Bstrut\\
\hline
$\bm{n}_{ij} \in \mathbb{R}^d$ & unit vector pointing from $v_i$ to $v_j$, $\bm{n}_{ij} = \frac{\bm{r}_j - \bm{r}_i}{\norm{\bm{r}_j - \bm{r}_i}}$ \Tstrut\Bstrut\\
\hline
$\bm{\dot{r}}_i^t \in \mathbb{R}^d$ & velocity of $v_i$ at time $t$  \Tstrut\Bstrut\\
\hline
$q_i \in \mathbb{R} $ & electric charge of particle $v_i$, it is a constant  \Tstrut\Bstrut\\
\hline
$m_i \in \mathbb{R} $ & mass of particle $v_i$, it is a constant  \Tstrut\Bstrut\\
\hline
$\bm{\eta}_i^t \in \mathbb{R}^{2d+2}$ & feature vector of particle $v_i$ at time $t$, $\bm{\eta}_i^t = [\bm{r}_i^t, \bm{\dot{r}}_i^t, q_i, m_i]$ \Tstrut\Bstrut\\
\hline
$\bm{\ddot{r}}_i^t \in \mathbb{R}^{d}$ & true acceleration of particle $v_i$ at time $t$  \Tstrut\Bstrut\\
\hline
$\bm{\hat{\ddot{r}}}_i^t \in \mathbb{R}^{d}$ & predicted acceleration of particle $v_i$ at time $t$  \Tstrut\Bstrut\\
\hline
$\bm{F}_{ij}^t \in \mathbb{R}^{d}$ & true force from  $v_j$ to $v_i$ at time $t$  \Tstrut\Bstrut\\
\hline
$\bm{\hat{F}}_{ij}^t \in \mathbb{R}^{d}$ & predicted force from  $v_j$ to $v_i$ at time $t$  \Tstrut\Bstrut\\
\hline
$P_{ij}^t \in \mathbb{R}$ & true potential energy incurred by $v_j$ on  $v_i$  at time $t$  \Tstrut\Bstrut\\
\hline
$\hat{P}_{ij}^t \in \mathbb{R}$ & predicted potential energy incurred by $v_j$ on  $v_i$  at time $t$  \Tstrut\Bstrut\\
\hline
$\hat{G}_E( \cdot ; \theta_{E} )$ & edge part neural network of \pignpi with learnable parameters $\theta_{E}$ \Tstrut\Bstrut\\
\hline
$G_N(\cdot)$ & proposed deterministic node part operator of \pignpi  \Tstrut\Bstrut\\
\hline
$\theta_{E}$ & learnable parameters in the edge neural network $\hat{G}_E( \cdot ; \theta_{E} )$  \Tstrut\Bstrut\\
\hline
$\mathcal{M}_{ij}$ & learnt message from $v_{j}$ to $v_{i}$ output by edge neural network $\hat{G}_E( \cdot ; \theta_{E} )$, $\mathcal{M}_{ij}\in \mathbb{R}^d$ in learning force and $\mathcal{M}_{ij}\in \mathbb{R}$ in learning potential\Tstrut\Bstrut \\
\hline
$\mathcal{M}_{i}$ & sum of all incoming message on particle $v_{i}$, $\mathcal{M}_{i}=\sum_i^{j\neq i}\mathcal{M}_{ij}$  \Tstrut\Bstrut\\
\hline
$\mathcal{T}_{\text{train}}$ 
& set of time steps corresponding to the training split of simulation data  \Tstrut\Bstrut\\
\hline
$\mathcal{T}_{\text{valid}}$ 
& set of time steps corresponding to the validation split of simulation data\Tstrut\Bstrut\\
\hline
$\mathcal{T}_{\text{test}}$ 
& set of time steps corresponding to the testing split of simulation data  \Tstrut\Bstrut\\
\hline
$l_1(x, y)$ & sum of absolute differences between each element in $x$ and $y$, $l_1(x, y) = \sum_i \abs{x_i - y_i}$, if $x$ and $y$ are vectors; or the absolute difference, $l_1(x, y) = \abs{x-y}$, if if $x$ and $y$ are scalars \Tstrut\Bstrut\\
\hline
$k$
& stiffness constant in spring simulation, we set $k=2$\Tstrut\Bstrut\\
\hline
$L$ & balance length constant in spring simulation, we set $L=1$\Tstrut\Bstrut\\
\hline
$c$ & constant in charge simulation, we set $c=1$\Tstrut\Bstrut\\
\hline
$\Theta$ & threshold constant in discontinuous dataset simulation, we set $\Theta=2$\Tstrut\Bstrut\\
\hline
\end{tabularx}}
\end{table}

\clearpage

\subsection{Performance evaluation of learning physics-consistent particle interactions (force and potential energy)}
\label{sec:force-potential-test-results}
Two different performance characteristics are evaluated. First, the learning performance is evaluated and, second, the ability of the algorithms to learn the particle interactions that are consistent with the underlying physical laws. 

We compute the following metrics for evaluating the performance of \pignpi and the baseline model to learn the pairwise force:
\begin{align}
&\textsf{MAE\textsubscript{acc}}= \textsf{MAE}^{\mathrm{part}}(\bm{\hat{\ddot{r}}},\bm{\ddot{r}}) =  \frac{1}{|\mathcal{T}_{\text{test}}|} \frac{1}{|V|}  \sum_{t\in \mathcal{T}_{\text{test}}}\sum_{i\in V} l_1(\bm{\hat{\ddot{r}}}_{i}^t, \bm{\ddot{r}}_{i}^t)
 \\
&\textsf{MAE\textsubscript{ef}} = \textsf{MAE}^{\mathrm{inter}}(\bm{\hat{F}},\bm{F}) = \frac{1}{|\mathcal{T}_{\text{test}}|} \frac{1}{|E|}  \sum_{t\in \mathcal{T}_{\text{test}}}\sum_{i,j \in V}^{i\neq j} l_1(\bm{\hat{F}}_{ij}^t, \bm{F}_{ij}^t)\\
&\textsf{MAE\textsubscript{nf}}= \textsf{MAE}^{\mathrm{part}}(\bm{\hat{F}}, \bm{F}) = \frac{1}{|\mathcal{T}_{\text{test}}|} \frac{1}{|V|}  \sum_{t\in \mathcal{T}_{\text{test}}}\sum_{i\in V} l_1(\bm{\hat{F}}_{i}^t, \bm{F}_{i}^t), \text{ where } \bm{\hat{F}}_{i}^t = \sum_j^{j\neq i}\bm{\hat{F}}_{ij}^t\\
&\textsf{MAE}_\textsf{symm}^{F} = \frac{1}{|\mathcal{T}_{\text{test}}|} \frac{1}{|E|}  \sum_{t\in \mathcal{T}_{\text{test}}}\sum_{i,j\in V}^{i\neq j} l_1(\bm{\hat{F}}_{ij}^t, - \bm{\hat{F}}_{ji}^t)
\end{align}
where $\bm{\ddot{r}}$ and $\bm{F}$ are the ground-truth acceleration and force, and $\bm{\hat{\ddot{r}}}$ and $\bm{\hat{F}}$ are the predicted  acceleration and force. 
Table~\ref{table:pairwise-force-result} reports the performance of the baseline model and \pignpi to the learn pairwise force in terms of metrics listed above (learning performance and the ability of the algorithms to learn physics-consistent particle interactions). 

\begin{table}[h!]
  \centering
    \caption{Performance of \pignpi and the baseline model on pairwise force prediction. Baseline\textsubscript{SiLU} denotes the baseline with the SiLU activation function. \lemos is the method to learn pairwise force introduced by \cite{lemos2021rediscovering}. Results averaged across five experiments.
    }
  \label{table:pairwise-force-result}
  \setlength\extrarowheight{-6pt}
  \scalebox{1.0}{
  \begin{tabularx}{\textwidth}{ccXXXXXXXX}
    \toprule
  & &\makecell[Xt]{Spring \\ dim=2} &\makecell[Xt]{Spring \\ dim=3} & \makecell[Xt]{Charge\\ dim=2} & \makecell[Xt]{Charge\\ dim=3} &\makecell[Xt]{Orbital \\ dim=2} &\makecell[Xt]{Orbital \\ dim=3} & \makecell[Xt]{{Discnt} \\ dim=2} &\makecell[Xt]{{Discnt} \\ dim=3}\\
    \midrule
        \multirow{8}{*}{\textsf{MAE\textsubscript{acc}}}

        &\multirow{2}{*}{Baseline}
        &0.0565 & 0.1076 & 0.2521 & 0.3824 & 0.0437 & 0.0439 & 0.0592 & 0.1171\\ 
        &&\scriptsize $\pm$0.0023 & \scriptsize $\pm$0.0012 & \scriptsize $\pm$0.0173 & \scriptsize $\pm$0.0559 & \scriptsize $\pm$0.0026 & \scriptsize $\pm$0.0014 & \scriptsize \scriptsize $\pm$0.0015 & \scriptsize $\pm$0.0010\\ 
        \cline{3-10}\rule{0pt}{2.3ex}

        &\multirow{2}{*}{{Baseline\textsubscript{SiLU}}}
        &0.0258 & 0.0476 & 1.0326 & 0.2092 & 0.0187 & 0.0196 & 0.0249 & 0.0508\\
        &&\scriptsize$\pm$0.0011 & \scriptsize $\pm$0.0025 & \scriptsize $\pm$1.3788 & \scriptsize $\pm$0.0060 & \scriptsize $\pm$0.0005 & \scriptsize $\pm$0.0002 & \scriptsize $\pm$0.0002 & \scriptsize $\pm$0.0010\\
        \cline{3-10}\rule{0pt}{2.3ex}
        
        &\multirow{2}{*}{\lemos}
        & 0.0246 & 0.0542 & 0.1216 & 0.1890 & 0.0255 & 0.0581 & 0.0667 & 0.2280\\
        &&\scriptsize$\pm$0.0047&\scriptsize$\pm$0.0047&\scriptsize$\pm$0.0099&\scriptsize$\pm$0.0111&\scriptsize$\pm$0.0023&\scriptsize$\pm$0.0004&\scriptsize$\pm$0.0174&\scriptsize$\pm$0.0951\\
        \cline{3-10}\rule{0pt}{2.3ex}

        &\multirow{2}{*}{\pignpi}
        &\textbf{0.0206} & \textbf{0.0278} & \textbf{0.0425} & \textbf{0.1191} & \textbf{0.0202} & \textbf{0.0182} & \textbf{0.0227} & \textbf{0.0399}\\
        &&\scriptsize\textbf{$\pm$0.0009} & \scriptsize \textbf{$\pm$0.0021} & \scriptsize \textbf{$\pm$0.0053} & \scriptsize \textbf{$\pm$0.0027} & \scriptsize \textbf{$\pm$0.0003} & \scriptsize \textbf{$\pm$0.0003} & \scriptsize \textbf{$\pm$0.0019} & \scriptsize \textbf{$\pm$0.0011}\\
        \hline\rule{0pt}{2.3ex}
    
        \multirow{8}{*}{\textsf{MAE\textsubscript{ef}}}
        &\multirow{2}{*}{Baseline}
        &2.3979 & 3.8952 & 1.1832 & 0.6447 & 4.1010 & 3.5379 & 1.6536 & 2.5803\\
        &&\scriptsize$\pm$0.2095 & \scriptsize $\pm$0.7178 & \scriptsize $\pm$0.0955 & \scriptsize $\pm$0.1118 & \scriptsize $\pm$0.1467 & \scriptsize $\pm$0.7571 & \scriptsize $\pm$0.0640 & \scriptsize $\pm$0.2886\\
        \cline{3-10}\rule{0pt}{2.3ex}

        &\multirow{2}{*}{{Baseline\textsubscript{SiLU}}}
        &4.2027 & 5.5185 & 2.1581 & 1.3842 & 3.1097 & 1.9863 & 2.6576 & 4.4222\\
        &&\scriptsize$\pm$1.1242 & \scriptsize $\pm$1.1452 & \scriptsize $\pm$0.9572 & \scriptsize $\pm$0.1411 & \scriptsize $\pm$0.7148 & \scriptsize $\pm$0.1434 & \scriptsize $\pm$0.4146 & \scriptsize $\pm$0.6116\\
        \cline{3-10}\rule{0pt}{2.3ex}
        
        &\multirow{2}{*}{\lemos}
        &0.5724 & 0.3638 & 1.0248 & 0.3137 & 0.9372 & 0.6943 & 0.3714 & 0.6974\\
        &&\scriptsize$\pm$0.2321&\scriptsize$\pm$0.3133&\scriptsize$\pm$0.0182&\scriptsize$\pm$0.0051&\scriptsize$\pm$0.0294&\scriptsize$\pm$0.0661&\scriptsize$\pm$0.2711&\scriptsize$\pm$0.4143\\
        \cline{3-10}\rule{0pt}{2.3ex}

        &\multirow{2}{*}{\pignpi}
        &\textbf{0.0063} & \textbf{0.0101} & \textbf{0.0136} & \textbf{0.0363} & \textbf{0.0093} & \textbf{0.0095} & \textbf{0.0040} & \textbf{0.0079}\\
        &&\scriptsize\textbf{$\pm$0.0002} & \scriptsize \textbf{$\pm$0.0007} & \scriptsize \textbf{$\pm$0.0023} & \scriptsize \textbf{$\pm$0.0015} & \scriptsize \textbf{$\pm$0.0002} & \scriptsize \textbf{$\pm$0.0001} & \scriptsize \textbf{$\pm$0.0004} & \scriptsize \textbf{$\pm$0.0002}\\
        \hline\rule{0pt}{2.3ex}
        
        \multirow{8}{*}{\textsf{MAE\textsubscript{nf}}}
        &\multirow{2}{*}{Baseline}
        &11.652 & 20.967 & 6.8310 & 3.8038 & 18.194 & 16.677 & 10.786 & 15.651\\
        &&\scriptsize$\pm$0.9890 & \scriptsize $\pm$3.8552 & \scriptsize $\pm$0.5548 & \scriptsize $\pm$0.7523 & \scriptsize $\pm$0.6884 & \scriptsize $\pm$3.5212 & \scriptsize $\pm$0.3764 & \scriptsize $\pm$1.7983\\
        \cline{3-10}\rule{0pt}{2.3ex}

        &\multirow{2}{*}{{Baseline\textsubscript{SiLU}}}
        &20.685 & 29.824 & 12.480 & 8.7533 & 13.644 & 9.2546 & 17.430 & 27.149\\
        &&\scriptsize$\pm$5.0491 & \scriptsize $\pm$6.2007 & \scriptsize $\pm$5.4145 & \scriptsize $\pm$0.9595 & \scriptsize $\pm$3.1699 & \scriptsize $\pm$0.6675 & \scriptsize $\pm$2.7127 & \scriptsize $\pm$3.8191\\
        \cline{3-10}\rule{0pt}{2.3ex}
        
        &\multirow{2}{*}{\lemos}
        & 2.7639 & 1.9370 & 5.9332 & 1.6546 & 3.9950 & 3.1841 & 2.4280 & 4.2270\\
        &&\scriptsize$\pm$1.1198&\scriptsize$\pm$1.6941&\scriptsize$\pm$0.1038&\scriptsize$\pm$0.0299&\scriptsize$\pm$0.1166&\scriptsize$\pm$0.2858&\scriptsize$\pm$1.8009&\scriptsize$\pm$2.6246\\
        \cline{3-10}\rule{0pt}{2.3ex}

        &\multirow{2}{*}{\pignpi}
        &\textbf{0.0219} & \textbf{0.0292} & \textbf{0.0488} & \textbf{0.1317} & \textbf{0.0260} & \textbf{0.0233} & \textbf{0.0239} & \textbf{0.0419}\\
        &&\scriptsize\textbf{$\pm$0.0010} & \scriptsize \textbf{$\pm$0.0022} & \scriptsize \textbf{$\pm$0.0059} & \scriptsize \textbf{$\pm$0.0033} & \scriptsize \textbf{$\pm$0.0005} & \scriptsize \textbf{$\pm$0.0004} & \scriptsize \textbf{$\pm$0.0020} & \scriptsize \textbf{$\pm$0.0011}\\
        \hline\rule{0pt}{2.3ex}

        \multirow{8}{*}{$\textsf{MAE}_\textsf{symm}^{F}$}
        &\multirow{2}{*}{Baseline}
        &1.1099 & 1.7452 & 0.1248 & 0.6938 & 2.2074 & 1.7684 & 0.9399 & 1.4118\\
        &&\scriptsize$\pm$0.0785 & \scriptsize $\pm$0.0467 & \scriptsize $\pm$0.0137 & \scriptsize $\pm$0.2670 & \scriptsize $\pm$0.1852 & \scriptsize $\pm$0.0941 & \scriptsize $\pm$0.0257 & \scriptsize $\pm$0.0722\\
        \cline{3-10}\rule{0pt}{2.3ex}

        &\multirow{2}{*}{{Baseline\textsubscript{SiLU}}}
        &2.1473 & 3.1809 & 2.1585 & 2.1378 & 0.8103 & 0.8877 & 1.6121 & 2.4231\\
        &&\scriptsize$\pm$0.1366 & \scriptsize $\pm$0.5156 & \scriptsize $\pm$1.8080 & \scriptsize $\pm$0.2803 & \scriptsize $\pm$0.1062 & \scriptsize $\pm$0.0584 & \scriptsize $\pm$0.2357 & \scriptsize $\pm$0.3908\\
        \cline{3-10}\rule{0pt}{2.3ex}
        
        &\multirow{2}{*}{\lemos}
        & 0.0400 & 0.0756 & 0.1013 & 0.0315 & 1.0831 & 0.8068 & 0.0102 & 0.0975\\
        &&\scriptsize$\pm$0.0437&\scriptsize$\pm$0.0284&\scriptsize$\pm$0.0082&\scriptsize$\pm$0.0033&\scriptsize$\pm$0.0733&\scriptsize$\pm$0.1251&\scriptsize$\pm$0.0041&\scriptsize$\pm$0.0927\\
        \cline{3-10}\rule{0pt}{2.3ex}

        &\multirow{2}{*}{\pignpi}
        &\textbf{0.0075} & \textbf{0.0133} & \textbf{0.0185} & \textbf{0.0345} & \textbf{0.0136} & \textbf{0.0134} & \textbf{0.0026} & \textbf{0.0066}\\
        &&\scriptsize\textbf{$\pm$0.0003} & \scriptsize \textbf{$\pm$0.0008} & \scriptsize \textbf{$\pm$0.0036} & \scriptsize \textbf{$\pm$0.0017} & \scriptsize \textbf{$\pm$0.0004} & \scriptsize \textbf{$\pm$0.0004} & \scriptsize \textbf{$\pm$0.0004} & \scriptsize \textbf{$\pm$0.0001}\\

      \bottomrule
\end{tabularx}
}
\end{table}

We use the following metrics for evaluating the performance to learn pairwise potential energy:
\begin{align}
&\textsf{MAE\textsubscript{acc}}= \textsf{MAE}^{\mathrm{part}}(\bm{\hat{\ddot{r}}},\bm{\ddot{r}}) =  \frac{1}{|\mathcal{T}_{\text{test}}|} \frac{1}{|V|}  \sum_{t\in \mathcal{T}_{\text{test}}}\sum_{i\in V} l_1(\bm{\hat{\ddot{r}}}_{i}^t, \bm{\ddot{r}}_{i}^t)
 \\
&\textsf{MAE\textsubscript{$\Delta$ep}} = \textsf{MAE}^\mathrm{inter} (\hat{P}-\hat{P}^0, P-P^0) = \frac{1}{|\mathcal{T}_{\text{test}}|} \frac{1}{|E|} \sum_{t \in \mathcal{T}_{\text{test}}} \sum_{i,j\in V}^{i \neq j} l_1(\hat{P}_{ij}^t - \hat{P}_{ij}^0, P_{ij}^t - P_{ij}^0)\\
&\textsf{MAE\textsubscript{$\Delta$np}} = \textsf{MAE}^\mathrm{part} (\hat{P} - \hat{P}^0, P - P^0) =\frac{1}{|\mathcal{T}_{\text{test}}|} \frac{1}{|V|} \sum_{t \in \mathcal{T}_{\text{test}}} \sum_{i \in |V|} l_1(\sum_{j}^{j\neq i}\hat{P}_{ij}^t - \sum_{j}^{j\neq i}\hat{P}_{ij}^0, P_{i}^t - P_{i}^0 )\\
&\textsf{MAE\textsubscript{ef}} = \textsf{MAE}^{\mathrm{inter}}(\bm{\hat{F}},\bm{F}) = \frac{1}{|\mathcal{T}_{\text{test}}|} \frac{1}{|E|}  \sum_{t\in \mathcal{T}_{\text{test}}}\sum_{i,j \in V}^{i\neq j} l_1(\bm{\hat{F}}_{ij}^t, \bm{F}_{ij}^t) \text{, where } \bm{\hat{F}}_{ij}^t = -\frac{\partial \hat{P}_{ij}^t}{\partial \bm{r}_{i}^t}\\
&\textsf{MAE\textsubscript{nf}}= \textsf{MAE}^{\mathrm{part}}(\bm{\hat{F}}, \bm{F}) = \frac{1}{|\mathcal{T}_{\text{test}}|} \frac{1}{|V|}  \sum_{t\in \mathcal{T}_{\text{test}}}\sum_{i\in V} l_1(\bm{\hat{F}}_{i}^t, \bm{F}_{i}^t) \text{, where } \bm{\hat{F}}_{i}^t=-\frac{\partial \sum_j^{j\neq i}\hat{P}_{ij}^t}{\partial \bm{r}_{i}^t}\\
&\textsf{MAE}_\textsf{symm}^{P}= \frac{1}{|\mathcal{T}_{\text{test}}|} \frac{1}{|E|}  \sum_{t\in \mathcal{T}_{\text{test}}}\sum_{i,j\in V}^{i\neq j} l_1(\hat{P}_{ij}^t, \hat{P}_{ji}^t)
\end{align}
where $\bm{\ddot{r}}$, $\bm{F}$ and $P$ are the ground-truth accelerations, forces and potentials, $\bm{\hat{\ddot{r}}}$, $\bm{\hat{F}}$ and $\hat{P}$ are the predictions computed from Eq.~(\ref{eq:potential_node_operator})-(\ref{eq:potential-force-prediction-potentialCase}). 
Table~\ref{table:potential-result} reports the  performance of baseline model and \pignpi to learn pairwise potential energy. 

\begin{table}[h!]
  \centering
    \caption{Performance evaluation of \pignpi and the baseline model on the pairwise potential energy learning task. {Baseline\textsubscript{SiLU}} denotes the baseline model with SiLU activation function. We report the error of predicting the potential energy and its first-order derivative which corresponds to the inter-particle force. Results averaged across five experiments.}
  \label{table:potential-result}
  \setlength\extrarowheight{-6pt}
  \scalebox{1.0}{
  \begin{tabularx}{\textwidth}{ccXXXXXXXX}
    \toprule
  & &\makecell[Xt]{Spring \\ dim=2} &\makecell[Xt]{Spring \\ dim=3} & \makecell[Xt]{Charge\\ dim=2} & \makecell[Xt]{Charge\\ dim=3} &\makecell[Xt]{Orbital \\ dim=2} &\makecell[Xt]{Orbital \\ dim=3} & \makecell[Xt]{{Discnt} \\ dim=2} &\makecell[Xt]{{Discnt} \\ dim=3}\\
    \midrule
        \multirow{6}{*}{\textsf{MAE\textsubscript{acc}}}

        &\multirow{2}{*}{Baseline}
        & 1.4841 & 1.9996 & 2.9127 & 0.5959 & 2.4585 & 1.0113 & 0.4532 & 0.7222\\
        &&\scriptsize$\pm$0.0064 & \scriptsize $\pm$0.0253 & \scriptsize $\pm$0.0844 & \scriptsize $\pm$0.0087 & \scriptsize $\pm$0.0399 & \scriptsize $\pm$0.0100 & \scriptsize $\pm$0.0186 & \scriptsize $\pm$0.0168\\
        \cline{3-10}\rule{0pt}{2.3ex}
        &\multirow{2}{*}{{Baseline\textsubscript{SiLU}}}
        & 1.4094 & 1.8721 & 4.5466 & 0.6047 & 2.2880 & 0.9044 & 0.3923 & 0.6554 \\
        &&\scriptsize$\pm$0.0842 & \scriptsize $\pm$0.0567 & \scriptsize $\pm$0.1088 & \scriptsize $\pm$0.0169 & \scriptsize $\pm$0.0231 & \scriptsize $\pm$0.0125 & \scriptsize $\pm$0.0058 & \scriptsize $\pm$0.0132\\
        \cline{3-10}\rule{0pt}{2.3ex}
        &\multirow{2}{*}{\pignpi}
        & \textbf{0.0076} & \textbf{0.0099} & \textbf{0.0225} & \textbf{0.1088} & \textbf{0.0090} & \textbf{0.0091} & \textbf{0.0089} & \textbf{0.0150}\\
        && \scriptsize \textbf{$\pm$0.0003} & \scriptsize \textbf{$\pm$0.0007} & \scriptsize \textbf{$\pm$0.0012} & \scriptsize \textbf{$\pm$0.0079} & \scriptsize \textbf{$\pm$0.0004} & \scriptsize \textbf{$\pm$0.0004} & \scriptsize \textbf{$\pm$0.0002} & \scriptsize \textbf{$\pm$0.0022}\\
        \hline\rule{0pt}{2.3ex}
        
        \multirow{6}{*}{\textsf{MAE\textsubscript{ef}}}
        &\multirow{2}{*}{Baseline}
        & 1.7644 & 2.4864 & 1.5492 & 0.5105 & 3.0739 & 1.9313 & 0.7243 & 1.1630 \\
        &&\scriptsize$\pm$0.0104 & \scriptsize $\pm$0.0104 & \scriptsize $\pm$0.0553 & \scriptsize $\pm$0.1459 & \scriptsize $\pm$0.0580 & \scriptsize $\pm$0.0076 & \scriptsize $\pm$0.0111 & \scriptsize $\pm$0.0063\\
        \cline{3-10}\rule{0pt}{2.3ex}
        &\multirow{2}{*}{{Baseline\textsubscript{SiLU}}}
        & 2.2647 & 2.7155 & 2.2720 & 1.0911 & 3.4747 & 2.1853 & 1.1977 & 1.6008 \\
        &&\scriptsize$\pm$0.0420 & \scriptsize $\pm$0.0333 & \scriptsize $\pm$0.2375 & \scriptsize $\pm$0.1684 & \scriptsize $\pm$0.0963 & \scriptsize $\pm$0.0321 & \scriptsize $\pm$0.0359 & \scriptsize $\pm$0.0311\\
        \cline{3-10}\rule{0pt}{2.3ex}
        &\multirow{2}{*}{\pignpi}
        & \textbf{0.0023} & \textbf{0.0037} & \textbf{0.0080} & \textbf{0.0223} & \textbf{0.0058} & \textbf{0.0053} & \textbf{0.0016} & \textbf{0.0030} \\
        && \scriptsize \textbf{$\pm$0.0001} & \scriptsize \textbf{$\pm$0.0003} & \scriptsize \textbf{$\pm$0.0006} & \scriptsize \textbf{$\pm$0.0013} & \scriptsize \textbf{$\pm$0.0011} & \scriptsize \textbf{$\pm$0.0005} & \tiny\textbf{$\pm$3.4E-5} & \scriptsize \textbf{$\pm$0.0004}\\
        \hline\rule{0pt}{2.3ex}
        
        \multirow{6}{*}{\textsf{MAE\textsubscript{nf}}}
        &\multirow{2}{*}{Baseline}
        & 8.3353 & 13.1721 & 9.3034 & 2.6295 & 14.1222 & 9.1354 & 4.6927 & 6.7873 \\
        &&\scriptsize$\pm$0.0627 & \scriptsize $\pm$0.0716 & \scriptsize $\pm$0.3807 & \scriptsize $\pm$0.6315 & \scriptsize $\pm$0.3045 & \scriptsize $\pm$0.0546 & \scriptsize $\pm$0.0584 & \scriptsize $\pm$0.0566\\
        \cline{3-10}\rule{0pt}{2.3ex}
        &\multirow{2}{*}{{Baseline\textsubscript{SiLU}}}
        & 10.3496 & 14.1069 & 10.6447 & 5.1109 & 16.2250 & 10.5310 & 7.9723 & 9.6495 \\
        &&\scriptsize$\pm$0.1832 & \scriptsize $\pm$0.1588 & \scriptsize $\pm$1.0175 & \scriptsize $\pm$0.6467 & \scriptsize $\pm$0.5242 & \scriptsize $\pm$0.1533 & \scriptsize $\pm$0.2558 & \scriptsize $\pm$0.1291\\
        \cline{3-10}\rule{0pt}{2.3ex}
        &\multirow{2}{*}{\pignpi}
        & \textbf{0.0080} & \textbf{0.0104} & \textbf{0.0261} & \textbf{0.1212} & \textbf{0.0115} & \textbf{0.0118} & \textbf{0.0098} & \textbf{0.0160} \\
        && \scriptsize \textbf{$\pm$0.0003} & \scriptsize \textbf{$\pm$0.0007} & \scriptsize \textbf{$\pm$0.0014} & \scriptsize \textbf{$\pm$0.0085} & \scriptsize \textbf{$\pm$0.0006} & \scriptsize \textbf{$\pm$0.0005} & \scriptsize \textbf{$\pm$0.0002} & \scriptsize \textbf{$\pm$0.0023}\\
        \hline\rule{0pt}{2.3ex}
        
        \multirow{6}{*}{\textsf{MAE\textsubscript{$\Delta$ep}}}
        &\multirow{2}{*}{Baseline}
        & 0.9588 & 1.1007 & 0.4656 & 0.3734 & 1.3174 & 1.1298 & 0.5979 & 0.8568 \\
        &&\scriptsize$\pm$0.0048 & \scriptsize $\pm$0.0058 & \scriptsize $\pm$0.0046 & \scriptsize $\pm$0.2349 & \scriptsize $\pm$0.0442 & \scriptsize $\pm$0.0088 & \scriptsize $\pm$0.0210 & \scriptsize $\pm$0.0167 \\
        \cline{3-10}\rule{0pt}{2.3ex}
        &\multirow{2}{*}{{Baseline\textsubscript{SiLU}}}
        & 1.2192 & 1.2418 & 1.6590 & 1.3053 & 1.4173 & 1.1852 & 0.9872 & 1.0427 \\
        &&\scriptsize$\pm$0.0296 & \scriptsize $\pm$0.0226 & \scriptsize $\pm$0.3484 & \scriptsize $\pm$0.2559 & \scriptsize $\pm$0.0648 & \scriptsize $\pm$0.0376 & \scriptsize $\pm$0.0395 & \scriptsize $\pm$0.0415 \\
        \cline{3-10}\rule{0pt}{2.3ex}
        &\multirow{2}{*}{\pignpi}
        & \textbf{0.0005} & \textbf{0.0016} & \textbf{0.0096} & \textbf{0.0156} & \textbf{0.0048} & \textbf{0.0031} & \textbf{0.2197} & \textbf{0.2344} \\
        && \tiny \textbf{$\pm$1.4E-5} & \scriptsize \textbf{$\pm$0.0003} & \scriptsize \textbf{$\pm$0.0009} & \scriptsize \textbf{$\pm$0.0015} & \scriptsize \textbf{$\pm$0.0017} & \scriptsize \textbf{$\pm$0.0006} & \scriptsize \textbf{$\pm$0.0001} & \scriptsize \textbf{$\pm$0.0001}\\
        \hline\rule{0pt}{2.3ex}
        
        \multirow{6}{*}{\textsf{MAE\textsubscript{$\Delta$np}}} 
        &\multirow{2}{*}{Baseline}
        & 4.0389 & 5.5378 & 1.5875 & 1.6498 & 4.9960 & 4.7247 & 2.9314 & 3.7998 \\
        &&\scriptsize$\pm$0.2410 & \scriptsize $\pm$0.1083 & \scriptsize $\pm$0.0363 & \scriptsize $\pm$1.2898 & \scriptsize $\pm$0.3848 & \scriptsize $\pm$0.0448 & \scriptsize $\pm$0.2362 & \scriptsize $\pm$0.2532\\
        \cline{3-10}\rule{0pt}{2.3ex}
        &\multirow{2}{*}{{Baseline\textsubscript{SiLU}}}
        & 5.3748 & 6.0889 & 6.3113 & 6.2894 & 5.4304 & 5.1887 & 5.2514 & 4.9127 \\
        &&\scriptsize$\pm$0.6218 & \scriptsize $\pm$0.2634 & \scriptsize $\pm$1.7498 & \scriptsize $\pm$0.8594 & \scriptsize $\pm$0.5369 & \scriptsize $\pm$0.1327 & \scriptsize $\pm$0.3821 & \scriptsize $\pm$0.5259\\
        \cline{3-10}\rule{0pt}{2.3ex}
        &\multirow{2}{*}{\pignpi}
        & \textbf{0.0016} & \textbf{0.0062} & \textbf{0.0179} & \textbf{0.0381} & \textbf{0.0129} & \textbf{0.0074} & \textbf{0.9319} & \textbf{0.9322} \\
        && \scriptsize \textbf{$\pm$0.0001} & \scriptsize \textbf{$\pm$0.0012} & \scriptsize \textbf{$\pm$0.0011} & \scriptsize \textbf{$\pm$0.0037} & \scriptsize \textbf{$\pm$0.0046} & \scriptsize \textbf{$\pm$0.0009} & \scriptsize \textbf{$\pm$0.0005} & \scriptsize \textbf{$\pm$0.0004}\\
        \hline\rule{0pt}{2.3ex}
        
        \multirow{6}{*}{$\textsf{MAE}_\textsf{symm}^{P}$} 
        &\multirow{2}{*}{Baseline}
        & 0.5641 & 0.4931 & 0.1094 & 0.2663 & 0.8017 & 0.4261 & 0.3538 & 0.3474 \\
        &&\scriptsize$\pm$0.0119 & \scriptsize $\pm$0.0056 & \scriptsize $\pm$0.0055 & \scriptsize $\pm$0.3068 & \scriptsize $\pm$0.0528 & \scriptsize $\pm$0.0121 & \scriptsize $\pm$0.0222 & \scriptsize $\pm$0.0106\\
        \cline{3-10}\rule{0pt}{2.3ex}
        &\multirow{2}{*}{{Baseline\textsubscript{SiLU}}}
        & 1.1668 & 0.9798 & 1.8888 & 1.2961 & 1.0689 & 0.7108 & 0.8908 & 0.8612 \\
        &&\scriptsize$\pm$0.0281 & \scriptsize $\pm$0.0735 & \scriptsize $\pm$0.6032 & \scriptsize $\pm$0.3374 & \scriptsize $\pm$0.0443 & \scriptsize $\pm$0.0394 & \scriptsize $\pm$0.0287 & \scriptsize $\pm$0.0305\\
        \cline{3-10}\rule{0pt}{2.3ex}
        &\multirow{2}{*}{\pignpi}
        & \textbf{0.0007} & \textbf{0.0025} & \textbf{0.0074} & \textbf{0.0062} & \textbf{0.0252} & \textbf{0.0422} & \textbf{0.0003} & \textbf{0.0005}\\
        && \scriptsize \textbf{$\pm$0.0001} & \scriptsize \textbf{$\pm$0.0008} & \scriptsize \textbf{$\pm$0.0010} & \scriptsize \textbf{$\pm$0.0005} & \scriptsize \textbf{$\pm$0.0065} & \scriptsize \textbf{$\pm$0.0144} & \tiny \textbf{$\pm$2.1E-5} & \tiny \textbf{$\pm$2.6E-5}\\
      \bottomrule
\end{tabularx}
}
\end{table}

\clearpage

\subsection{Evaluation of the generalization ability on learning the pairwise force and potential energy}
\label{sec:genralization-test-results}

We evaluate the generalization ability of the baseline model, \lemos and \pignpi by first training the models on an eight-particle system and then evaluating their performance on a 12-particle system. 
We evaluate the performance of baseline model, \lemos and \pignpi on the pairwise force learning task (Table~\ref{table:transfer_generalization_force}) and baseline model and \pignpi on the pairwise potential energy learning task (Table~\ref{table:transfer_generalization_potential}) because \lemos is only designed for learning the pairwise force. 
Furthermore, a limitation of \lemos is, after training, it cannot be generalized to predict the acceleration for a new system. The reason is the learnt node property is specifically associated to the system used for training. We need to train \lemos from scratch again to predict the acceleration for a new system.

\begin{table}[h!]
  \centering
  \caption{Evaluation of the generalization ability on the pairwise force learning task. Models are trained on a eight-particle system and then tested on a 12-particle system. Results averaged across five experiments. Note that \lemos cannot be generalized to predict the acceleration because of the learnt node property.}
  \label{table:transfer_generalization_force}
  \setlength\extrarowheight{-6pt}
  \scalebox{1.0}{
  \begin{tabularx}{\textwidth}{ccXXXXXXXX}
    \toprule
  & &\makecell[Xt]{Spring \\ dim=2} &\makecell[Xt]{Spring \\ dim=3} & \makecell[Xt]{Charge\\ dim=2} & \makecell[Xt]{Charge\\ dim=3} &\makecell[Xt]{Orbital \\ dim=2} &\makecell[Xt]{Orbital \\ dim=3} & \makecell[Xt]{{Discnt} \\ dim=2} &\makecell[Xt]{{Discnt} \\ dim=3}\\
    \midrule
        \multirow{6}{*}{\textsf{MAE\textsubscript{acc}}}        

        &\multirow{2}{*}{Baseline}
        & 0.2790 & 0.5664 & 1.0363 & 2.2038 & 0.1007 & 0.1497 & 0.2067 & 0.3705\\
        &&\scriptsize$\pm$0.0402 & \scriptsize $\pm$0.0630 & \scriptsize $\pm$0.0780 & \scriptsize $\pm$0.3393 & \scriptsize $\pm$0.0096 & \scriptsize $\pm$0.0166 & \scriptsize $\pm$0.0217 & \scriptsize $\pm$0.0274\\
        \cline{3-10}\rule{0pt}{2.3ex}
        
        &\multirow{2}{*}{\lemos}
        &- & - & - & - & - & - & - & -\\
        &&- & - & - & - & - & - & - & -\\
        \cline{3-10}\rule{0pt}{2.3ex}

        &\multirow{2}{*}{\pignpi}
        & \textbf{0.0449} & \textbf{0.0680} & \textbf{0.3561} & \textbf{0.5467} & \textbf{0.0413} & \textbf{0.0407} & \textbf{0.0489} & \textbf{0.0726}\\
        && \scriptsize \textbf{$\pm$0.0014} & \scriptsize \textbf{$\pm$0.0034} & \scriptsize \textbf{$\pm$0.0481} & \scriptsize \textbf{$\pm$0.0441} & \scriptsize \textbf{$\pm$0.0020} & \scriptsize \textbf{$\pm$0.0010} & \scriptsize \textbf{$\pm$0.0042} & \scriptsize \textbf{$\pm$0.0014}\\
        \hline\rule{0pt}{2.3ex}
        
        \multirow{6}{*}{\textsf{MAE\textsubscript{ef}}}

        &\multirow{2}{*}{Baseline}
        & 2.1514 & 4.2343 & 0.6920 & 0.6283 & 3.3921 & 3.3837 & 1.6555 & 2.7026\\
        &&\scriptsize$\pm$0.1950 & \scriptsize $\pm$0.7791 & \scriptsize $\pm$0.0616 & \scriptsize $\pm$0.1043 & \scriptsize $\pm$0.1321 & \scriptsize $\pm$0.7063 & \scriptsize $\pm$0.1018 & \scriptsize $\pm$0.3773\\
        \cline{3-10}\rule{0pt}{2.3ex}
        
        &\multirow{2}{*}{\lemos}
        &0.5563 & 0.3990 & 0.5907 & 0.3177 & 0.6381 & 0.5792 & 0.3717 & 0.7442\\
        &&\scriptsize$\pm$0.2231&\scriptsize$\pm$0.3266&\scriptsize$\pm$0.0102&\scriptsize$\pm$0.0053&\scriptsize$\pm$0.0030&\scriptsize$\pm$0.0328&\scriptsize$\pm$0.2698&\scriptsize$\pm$0.5024\\
        \cline{3-10}\rule{0pt}{2.3ex}

        &\multirow{2}{*}{\pignpi}
        & \textbf{0.0087} & \textbf{0.0149} & \textbf{0.0481} & \textbf{0.0697} & \textbf{0.0111} & \textbf{0.0113} & \textbf{0.0052} & \textbf{0.0092}\\
        && \scriptsize \textbf{$\pm$0.0002} & \scriptsize \textbf{$\pm$0.0008} & \scriptsize \textbf{$\pm$0.0065} & \scriptsize \textbf{$\pm$0.0043} & \scriptsize \textbf{$\pm$0.0008} & \scriptsize \textbf{$\pm$0.0004} & \scriptsize \textbf{$\pm$0.0005} & \scriptsize \textbf{$\pm$0.0002}\\
        \hline\rule{0pt}{2.3ex}

        \multirow{6}{*}{\textsf{MAE\textsubscript{nf}}}

        &\multirow{2}{*}{Baseline}
        &14.7789 & 34.1458 & 5.8505 & 5.2545 & 21.1451 & 23.0175 & 16.5313 & 25.0296\\
        &&\scriptsize$\pm$1.4639 & \scriptsize $\pm$6.3003 & \scriptsize $\pm$0.5123 & \scriptsize $\pm$1.0215 & \scriptsize $\pm$0.9028 & \scriptsize $\pm$4.7677 & \scriptsize $\pm$1.0373 & \scriptsize $\pm$3.6511 \\
        \cline{3-10}\rule{0pt}{2.3ex}
        
        &\multirow{2}{*}{\lemos}
        &3.8516 & 3.2087 & 4.9887 & 2.3699 & 4.0284 & 3.8214 & 3.7130 & 6.5094\\
        &&\scriptsize$\pm$1.5444&\scriptsize$\pm$2.6770&\scriptsize$\pm$0.0858&\scriptsize$\pm$0.0388&\scriptsize$\pm$0.0148&\scriptsize$\pm$0.2360&\scriptsize$\pm$2.7434&\scriptsize$\pm$4.1421\\
        \cline{3-10}\rule{0pt}{2.3ex}

        &\multirow{2}{*}{\pignpi}
        & \textbf{0.0443} & \textbf{0.0665} & \textbf{0.4178} & \textbf{0.5564} & \textbf{0.0476} & \textbf{0.0451} & \textbf{0.0477} & \textbf{0.0730}\\
        && \scriptsize \textbf{$\pm$0.0012} & \scriptsize \textbf{$\pm$0.0033} & \scriptsize \textbf{$\pm$0.0614} & \scriptsize \textbf{$\pm$0.0425} & \scriptsize \textbf{$\pm$0.0029} & \scriptsize \textbf{$\pm$0.0011} & \scriptsize \textbf{$\pm$0.0039} & \scriptsize \textbf{$\pm$0.0016}\\
        \hline\rule{0pt}{2.3ex}
        
        \multirow{6}{*}{$\textsf{MAE}_\textsf{symm}^{F}$}
        &\multirow{2}{*}{Baseline} 
        & 1.0060 & 1.6034 & 0.1059 & 0.6677 & 1.6018 & 1.6047 & 0.8586 & 1.2154 \\
        &&\scriptsize$\pm$0.0711 & \scriptsize $\pm$0.0494 & \scriptsize $\pm$0.0158 & \scriptsize $\pm$0.2549 & \scriptsize $\pm$0.1370 & \scriptsize $\pm$0.0858 & \scriptsize $\pm$0.0239 & \scriptsize $\pm$0.0622\\
        \cline{3-10}\rule{0pt}{2.3ex}
        
        &\multirow{2}{*}{\lemos}
        &0.0452 & 0.0731 & 0.0775 & 0.0357 & 0.7903 & 0.7328 & 0.0114 & 0.2427\\
        &&\scriptsize$\pm$0.0372&\scriptsize$\pm$0.0250&\scriptsize$\pm$0.0065&\scriptsize$\pm$0.0032&\scriptsize$\pm$0.0542&\scriptsize$\pm$0.1140&\scriptsize$\pm$0.0053&\scriptsize$\pm$0.3709\\
        \cline{3-10}\rule{0pt}{2.3ex}

        &\multirow{2}{*}{\pignpi}
        & \textbf{0.0108} & \textbf{0.0197} & \textbf{0.0733} & \textbf{0.0614} & \textbf{0.0158} & \textbf{0.0149} & \textbf{0.0039} & \textbf{0.0086}\\
        && \scriptsize \textbf{$\pm$0.0003} & \scriptsize \textbf{$\pm$0.0008} & \scriptsize \textbf{$\pm$0.0125} & \scriptsize \textbf{$\pm$0.0021} & \scriptsize \textbf{$\pm$0.0013} & \scriptsize \textbf{$\pm$0.0005} & \scriptsize \textbf{$\pm$0.0006} & \scriptsize \textbf{$\pm$0.0003}\\
        
      \bottomrule
\end{tabularx}
}
\end{table}

\begin{table}[H]
  \centering
  \caption{Evaluation of the generalization ability on the potential energy learning task. Models are trained on a eight-particle system and then tested on a 12-particle system. Results averaged across five experiments. Here, the comparison model does not contain \lemos because it is only designed for learning force.}
  \label{table:transfer_generalization_potential}
  \setlength\extrarowheight{-6pt}
  \scalebox{1.0}{
  \begin{tabularx}{\textwidth}{ccXXXXXXXX}
    \toprule
  & &\makecell[Xt]{Spring \\ dim=2} &\makecell[Xt]{Spring \\ dim=3} & \makecell[Xt]{Charge\\ dim=2} & \makecell[Xt]{Charge\\ dim=3} &\makecell[Xt]{Orbital \\ dim=2} &\makecell[Xt]{Orbital \\ dim=3} & \makecell[Xt]{{Discnt} \\ dim=2} &\makecell[Xt]{{Discnt} \\ dim=3}\\
    \midrule
        \multirow{4}{*}{\textsf{MAE\textsubscript{acc}}}
        & \multirow{2}{*}{Baseline} 
        & 6.7336 & 14.697 & 6.2643 & 3.5436 & 5.5236 & 6.0802 & 2.6173 & 5.4259\\
        &&\scriptsize$\pm$0.0626 & \scriptsize $\pm$0.3470 & \scriptsize $\pm$0.7284 & \scriptsize $\pm$0.3228 & \scriptsize $\pm$0.0863 & \scriptsize $\pm$0.0930 & \scriptsize $\pm$0.1238 & \scriptsize $\pm$0.3450\\
        \cline{3-10}\rule{0pt}{2.3ex}

        & \multirow{2}{*}{\pignpi}
        & \textbf{0.0180} & \textbf{0.0238} & \textbf{1.1900} & \textbf{0.5542} & \textbf{0.0760} & \textbf{0.0995} & \textbf{0.0215} & \textbf{0.0311}\\
        && \scriptsize \textbf{$\pm$0.0018} & \scriptsize \textbf{$\pm$0.0025} & \scriptsize \textbf{$\pm$0.3611} & \scriptsize \textbf{$\pm$0.0644} & \scriptsize \textbf{$\pm$0.0167} & \scriptsize \textbf{$\pm$0.0198} & \scriptsize \textbf{$\pm$0.0017} & \scriptsize \textbf{$\pm$0.0020}\\
        \hline\rule{0pt}{2.3ex}
        
        \multirow{4}{*}{\textsf{MAE\textsubscript{ef}}}
        & \multirow{2}{*}{Baseline} 
        & 1.7397 & 2.6430 & 0.8552 & 0.5264 & 2.3356 & 1.7652 & 0.7328 & 1.1855\\
        &&\scriptsize$\pm$0.0114 & \scriptsize $\pm$0.0120 & \scriptsize $\pm$0.0267 & \scriptsize $\pm$0.1457 & \scriptsize $\pm$0.0487 & \scriptsize $\pm$0.0116 & \scriptsize $\pm$0.0148 & \scriptsize $\pm$0.0213\\
        \cline{3-10}\rule{0pt}{2.3ex}

        & \multirow{2}{*}{\pignpi}
        & \textbf{0.0034} & \textbf{0.0049} & \textbf{0.1385} & \textbf{0.0631} & \textbf{0.0144} & \textbf{0.0186} & \textbf{0.0022} & \textbf{0.0040}\\
        && \scriptsize \textbf{$\pm$0.0002} & \scriptsize \textbf{$\pm$0.0004} & \scriptsize \textbf{$\pm$0.0357} & \scriptsize \textbf{$\pm$0.0065} & \scriptsize \textbf{$\pm$0.0020} & \scriptsize \textbf{$\pm$0.0039} & \scriptsize \textbf{$\pm$0.0001} & \scriptsize \textbf{$\pm$0.0003}\\
        \hline\rule{0pt}{2.3ex}
        
        \multirow{4}{*}{\textsf{MAE\textsubscript{nf}}}
        & \multirow{2}{*}{Baseline} 
        & 11.819 & 21.083 & 7.5812 & 3.7500 & 15.014 & 12.254 & 7.2370 & 10.400\\
        &&\scriptsize$\pm$0.0599 & \scriptsize $\pm$0.1454 & \scriptsize $\pm$0.2963 & \scriptsize $\pm$0.7783 & \scriptsize $\pm$0.3101 & \scriptsize $\pm$0.1030 & \scriptsize $\pm$0.1482 & \scriptsize $\pm$0.0918\\
        \cline{3-10}\rule{0pt}{2.3ex}

        & \multirow{2}{*}{\pignpi}
        & \textbf{0.0173} & \textbf{0.0233} & \textbf{1.3505} & \textbf{0.5486} & \textbf{0.0704} & \textbf{0.0930} & \textbf{0.0204} & \textbf{0.0317}\\
        && \scriptsize \textbf{$\pm$0.0013} & \scriptsize \textbf{$\pm$0.0022} & \scriptsize \textbf{$\pm$0.3867} & \scriptsize \textbf{$\pm$0.0646} & \scriptsize \textbf{$\pm$0.0138} & \scriptsize \textbf{$\pm$0.0171} & \scriptsize \textbf{$\pm$0.0013} & \scriptsize \textbf{$\pm$0.0022}\\
        \hline\rule{0pt}{2.3ex}
        
        \multirow{4}{*}{\textsf{MAE\textsubscript{$\Delta$ep}}}
        & \multirow{2}{*}{Baseline} 
        & 2.1921 & 3.1820 & 0.5311 & 0.4822 & 0.9830 & 0.9161 & 0.7449 & 1.7985\\
        &&\scriptsize$\pm$0.0082 & \scriptsize $\pm$0.0284 & \scriptsize $\pm$0.0044 & \scriptsize $\pm$0.1159 & \scriptsize $\pm$0.0112 & \scriptsize $\pm$0.0032 & \scriptsize $\pm$0.0165 & \scriptsize $\pm$0.0182\\
        \cline{3-10}\rule{0pt}{2.3ex}

        & \multirow{2}{*}{\pignpi}
        & \textbf{0.0022} & \textbf{0.0033} & \textbf{0.0516} & \textbf{0.0428} & \textbf{0.0086} & \textbf{0.0106} & \textbf{0.2238} & \textbf{0.2415}\\
        && \scriptsize \textbf{$\pm$0.0002} & \scriptsize \textbf{$\pm$0.0004} & \scriptsize \textbf{$\pm$0.0072} & \scriptsize \textbf{$\pm$0.0037} & \scriptsize \textbf{$\pm$0.0007} & \scriptsize \textbf{$\pm$0.0023} & \scriptsize \textbf{$\pm$0.0001} & \scriptsize \textbf{$\pm$0.0003}\\
    \hline\rule{0pt}{2.3ex}
        
        \multirow{4}{*}{\textsf{MAE\textsubscript{$\Delta$np}}}
        & \multirow{2}{*}{Baseline} 
        & 19.256 & 28.514 & 2.0988 & 1.9149 & 7.1296 & 7.4346 & 5.6309 & 15.478\\
        &&\scriptsize$\pm$0.0530 & \scriptsize $\pm$0.4680 & \scriptsize $\pm$0.0294 & \scriptsize $\pm$1.0773 & \scriptsize $\pm$0.2341 & \scriptsize $\pm$0.1342 & \scriptsize $\pm$0.2235 & \scriptsize $\pm$0.2467\\
        \cline{3-10}\rule{0pt}{2.3ex}

        & \multirow{2}{*}{\pignpi}
        & \textbf{0.0137} & \textbf{0.0166} & \textbf{0.2937} & \textbf{0.1668} & \textbf{0.0415} & \textbf{0.0503} & \textbf{1.3273} & \textbf{1.8850}\\
        && \scriptsize \textbf{$\pm$0.0022} & \scriptsize \textbf{$\pm$0.0049} & \scriptsize \textbf{$\pm$0.0485} & \scriptsize \textbf{$\pm$0.0173} & \scriptsize \textbf{$\pm$0.0063} & \scriptsize \textbf{$\pm$0.0105} & \scriptsize \textbf{$\pm$0.0009} & \scriptsize \textbf{$\pm$0.0028}\\
    \hline\rule{0pt}{2.3ex}
        
        \multirow{4}{*}{$\textsf{MAE}_\textsf{symm}^{P}$}
        & \multirow{2}{*}{Baseline} 
        & 0.4309 & 0.5237 & 0.0481 & 0.2303 & 0.6652 & 0.3765 & 0.3753 & 0.3894\\
        &&\scriptsize$\pm$0.0094 & \scriptsize $\pm$0.0743 & \scriptsize $\pm$0.0048 & \scriptsize $\pm$0.2488 & \scriptsize $\pm$0.0404 & \scriptsize $\pm$0.0188 & \scriptsize $\pm$0.0190 & \scriptsize $\pm$0.0914\\
        \cline{3-10}\rule{0pt}{2.3ex}

        & \multirow{2}{*}{\pignpi}
        & \textbf{0.0010} & \textbf{0.0033} & \textbf{0.0159} & \textbf{0.0124} & \textbf{0.0219} & \textbf{0.0336} & \textbf{0.0004} & \textbf{0.0009}\\
        && \scriptsize \textbf{$\pm$0.0001} & \scriptsize \textbf{$\pm$0.0007} & \scriptsize \textbf{$\pm$0.0027} & \scriptsize \textbf{$\pm$0.0011} & \scriptsize \textbf{$\pm$0.0065} & \scriptsize \textbf{$\pm$0.0099} & \scriptsize \textbf{$\pm$0.0001} & \scriptsize \textbf{$\pm$0.0001}\\
        
      \bottomrule
\end{tabularx}
}
\end{table}

\clearpage

\subsection{Potential energy prediction in the discontinuous dataset}
\label{sec:discontinuous_potential_prediction}

\begin{figure}[h]
\centering
\includegraphics[width=\linewidth]{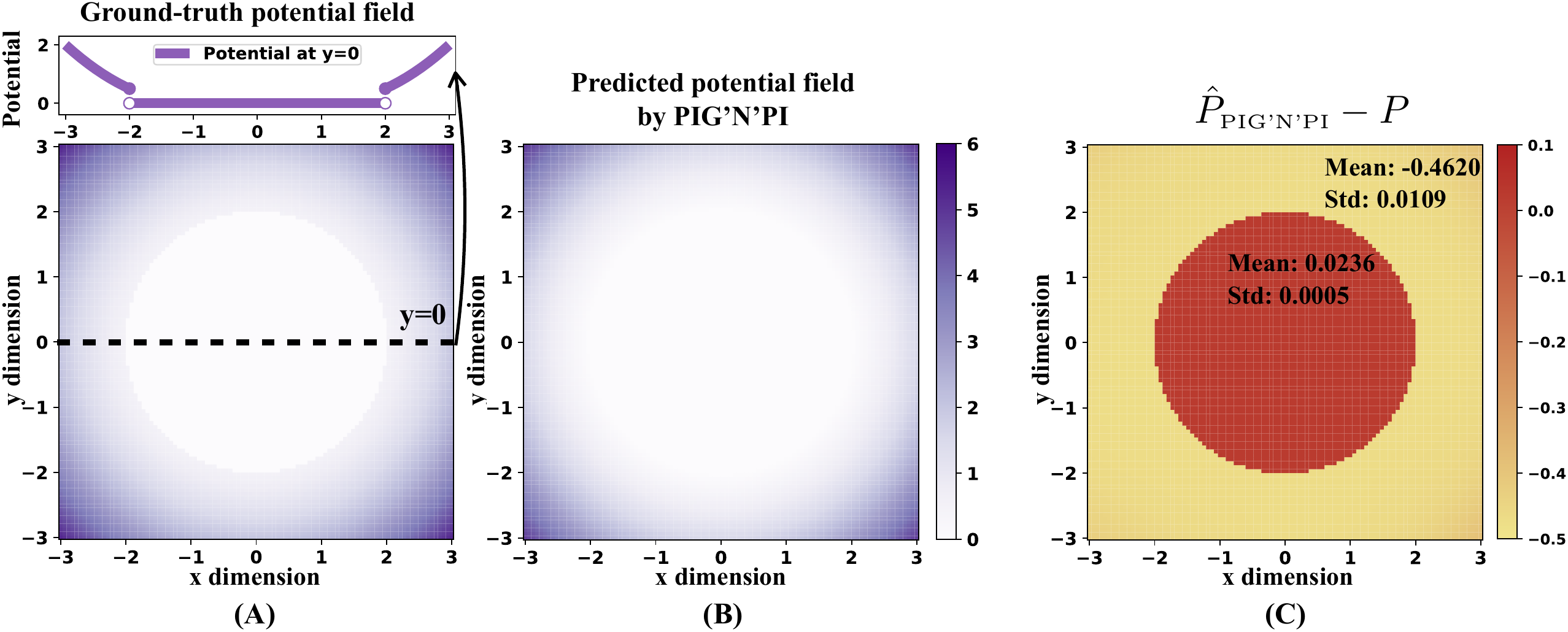}
\caption{\textbf{Ground-truth potential energy and predicted potential energy of \pignpi for the Discontinuous dataset.} (A) The ground-truth discontinuous potential field around a fixed particle  at center. The potential between two particles is discontinuous at distance $r=2$. (top) Cross-section of the potential at $y=0$. (B) The predicted potential field by \pignpi. (C) Difference between the potential field predicted by \pignpi and the ground-truth: $\hat{P}_{\text{\pignpi}}-P$. The mean value and standard deviation are computed separately for the two areas limited by the position of the discontinuity in the potential, $r=2$. }
\label{fig:discontinuous_data}
\end{figure}

Here, we take a closer look at the discontinuous dataset as it presented a particularly large \textsf{MAE\textsubscript{$\Delta$ep}} for \pignpi predictions compared to the other (continuous) datasets (Fig.~\ref{fig:potential_summary}). The potential energy field $P$ presents a discontinuity at $r=2$ (see Fig.~\ref{fig:discontinuous_data}(A)), where $P=0$ for $r<2$ and $P\geq 0.5$ for $r \geq 2$. \pignpi, however, appears to infer a continuous potential function $P_\text{\pignpi}$ (see Fig.~\ref{fig:discontinuous_data}(B)) that presents similar trades to the ground-truth but without the discontinuity. In fact, \pignpi infers the shape of the potential energy function independently in the two areas separated by $r=2$ without learning the absolute value of the potential energy (see $P_\text{\pignpi}-P$ in Fig.~\ref{fig:discontinuous_data}(C)). The reported mean values of $P_\text{\pignpi}-P$ for each area (see Fig.~\ref{fig:discontinuous_data}(C)) are relatively large indicating the error in the absolute value, whereas the values for the standard deviation are small in both areas showing that \pignpi infers well the shape of the potential (\textit{i.e.}, the derivative of the potential). 

Note that the difference in the mean values between the two areas suggests that the absolute value is differently incorrect in the two areas. This explains why the \textsf{MAE\textsubscript{$\Delta$ep}} of \pignpi is larger on the discontinuous dataset (Fig.~\ref{fig:potential_summary}) compared to other datasets. Here, \pignpi learns the shape of the potential energy function in two ranges separately, and hence introduces a different discontinuity, which leads to an arbitrary constant that is integrated into the \textsf{MAE\textsubscript{$\Delta$ep}} computation over the \emph{entire} space. Therefore, the increased value of \textsf{MAE\textsubscript{$\Delta$ep}} simply indicates that the discontinuity in the potential cannot be normalized-out with a measure of the relative potential energy as for the continuous datasets.

\clearpage

\subsection{Performance evaluation for the LJ-argon dataset}
\label{sec:LJ-argon-evaluation}
Table~\ref{table:LJ-argon-force} and Table~\ref{table:LJ-argon-potential} report the performance of \pignpi to learn pairwise force and pairwise potential energy. $\textsf{MAE\textsubscript{acc}}$, $\textsf{MAE\textsubscript{$\Delta$ep}}$, $\textsf{MAE\textsubscript{$\Delta$np}}$, $\textsf{MAE\textsubscript{ef}}$, $\textsf{MAE\textsubscript{nf}}$, $\textsf{MAE}_\textsf{symm}^{F}$ and $\textsf{MAE}_\textsf{symm}^{P}$ are defined same as before (see Sec.~\ref{sec:performance_evaluation} and Sec.~\ref{sec:force-potential-test-results}). We also compute the division between each error and the average of its corresponding ground-truth as the relative error. For example, the relative $\textsf{MAE\textsubscript{acc}}=\textsf{MAE\textsubscript{acc}}/\frac{1}{N}\frac{1}{T}\sum_{i=1}^{N}\sum_{t=1}^{T}|\bm{\ddot{r}}_i^t|$. Note that the relative $\textsf{MAE\textsubscript{acc}}$ equals to the relative $\textsf{MAE\textsubscript{nf}}$ because all particles have the same mass. Baseline$_{\alpha=*}$ refers to the baseline model with symmetry regularization. See Sec.~\ref{sec:method_symmetry_regularization} for the details of imposing the symmetry regularization into baseline. We can find that symmetry regularization makes the baseline model perform better in terms of $\textsf{MAE\textsubscript{ef}}$. However, \pignpi is still significantly better than the extended baseline. Furthermore, when evaluate the models to learn pairwise force, we also test the method \lemos proposed by \cite{lemos2021rediscovering} to learn pairwise force (see Sec.~\ref{sec:method_lemos2021rediscovering}). Considering particles in this dataset have the same mass, we also test a variant of \lemos such that we assign all nodes with a unique learnable scalar. We denote this variant as GN+\textsubscript{uni}. We can see that GN+\textsubscript{uni} is better than the baseline and \lemos. However, \pignpi still outperforms GN+\textsubscript{uni} by more than one order of magnitude, especially if we look at the \textsf{MAE\textsubscript{ef}} which measures the quality of the predicted pariwise force.
\begin{table}[H]
  \centering
  \caption{Evaluation of the performance to learn pairwise force for the LJ-argon dataset. Results
averaged across five experiments.}
  \label{table:LJ-argon-force}
  \setlength\extrarowheight{-6pt}
  \scalebox{0.85}{
  \begin{tabularx}{1.2\textwidth}{cXXXXXXXX}
    \toprule
  &\makecell[Xt]{\textsf{MAE\textsubscript{acc}}\\(\r{A}/ps$^2$)} &\makecell[Xt]{Relative\\ \textsf{MAE\textsubscript{acc}}} &\makecell[Xt]{\textsf{MAE\textsubscript{nf}}\\ \footnotesize(meV/\r{A})} &\makecell[Xt]{Relative\\ \textsf{MAE\textsubscript{nf}}} & \makecell[Xt]{\textsf{MAE\textsubscript{ef}}\\ \footnotesize(meV/\r{A})} & \makecell[Xt]{Relative\\ \textsf{MAE\textsubscript{ef}}} & \makecell[Xt]{$\textsf{MAE}_\textsubscript{symm}^{F}$\\ \footnotesize(meV/\r{A})} &\makecell[Xt]{Relative\\$\textsf{MAE}_\textsubscript{symm}^{F}$}\\
    \midrule
         \multirow{2}{*}{Baseline} 
        & 0.4230 & 2.66\% & 1.7493 & 2.66\% & 7.2635 & 269.41\% & 0.3885 & 14.41\%\\
        &\scriptsize$\pm$0.0206 &\scriptsize$\pm$0.13\% &\scriptsize$\pm$0.0850 &\scriptsize$\pm$0.13\% &\scriptsize$\pm$0.8811 &\scriptsize$\pm$32.68\% &\scriptsize$\pm$0.0425 &\scriptsize$\pm$1.58\% \\
        
        \cline{2-9}\rule{0pt}{2.3ex}

        \multirow{2}{*}{Baseline$_{\alpha=0.1}$}
        & 0.6326 & 3.97\% & 2.6160 & 3.97\% & 3.0155 & 111.85\% & 0.1263 & 4.68\% \\
        &\scriptsize$\pm$0.0381 &\scriptsize$\pm$0.24\% &\scriptsize$\pm$0.1576 &\scriptsize$\pm$0.24\% &\scriptsize$\pm$0.3660 &\scriptsize$\pm$13.57\% &\scriptsize$\pm$0.0050 &\scriptsize$\pm$0.19\% \\
        
        \cline{2-9}\rule{0pt}{2.3ex}

        \multirow{2}{*}{Baseline$_{\alpha=1}$} 
        & 0.8022 & 5.04\% & 3.3171 & 5.04\% & 2.7034 & 100.27\% & 0.0879 & 3.26\% \\
        &\scriptsize$\pm$0.0706 &\scriptsize$\pm$0.44\% &\scriptsize$\pm$0.2918 &\scriptsize$\pm$0.44\% &\scriptsize$\pm$0.2164 &\scriptsize$\pm$8.02\% &\scriptsize$\pm$0.0126 &\scriptsize$\pm$0.47\% \\

        \cline{2-9}\rule{0pt}{2.3ex}

        \multirow{2}{*}{Baseline$_{\alpha=10}$} 
        & 15.1633 & 95.24\% & 62.7044 & 95.24\% & 2.6961 & 100.00\% & 0.0069 & 0.26\% \\
        &\scriptsize$\pm$1.7219 &\scriptsize$\pm$10.81\% &\scriptsize$\pm$7.1206 &\scriptsize$\pm$10.81\% &\scriptsize$\pm$0.0001 &\scriptsize$\pm$2.94E-5 &\scriptsize$\pm$0.0048 &\scriptsize$\pm$0.18\% \\

        \cline{2-9}\rule{0pt}{2.3ex}

        \multirow{2}{*}{Baseline$_{\alpha=100}$}
        & 15.9832 & 100.39\% & 66.0951 & 100.39\% & 2.6961 & 100.00\% & \textbf{1.64E-8} & \textbf{6.07E-9} \\
        &\scriptsize$\pm$0.0027 &\scriptsize$\pm$0.02\% &\scriptsize$\pm$0.0110 &\scriptsize$\pm$0.02\% &\scriptsize$\pm$0.0000 &\scriptsize$\pm$0.00E0 &\scriptsize\textbf{$\pm$1.18E-8} &\scriptsize\textbf{$\pm$4.37E-9} \\
        
        \cline{1-9}\rule{0pt}{2.3ex}
        
        \multirow{2}{*}{GN+}
        & 15.7991 & 99.23\% & 65.3337 & 99.23\% & 7.7990 & 289.27\% & 1.1210 & 41.58\%\\
        &\scriptsize$\pm$0.0001 &\scriptsize$\pm$3.54E-06 &\scriptsize$\pm$0.0002 &\scriptsize$\pm$3.54E-06 &\scriptsize$\pm$0.2822 &\scriptsize$\pm$10.47\% &\scriptsize$\pm$0.0599 &\scriptsize$\pm$2.22\%\\

        \cline{2-9}\rule{0pt}{2.3ex}

        \multirow{2}{*}{GN+\textsubscript{uni}}
        & 0.3832 & 2.41\% & 1.5848 & 2.41\% & 0.5799 & 21.51\% & {0.0155} & {0.58\%}\\
        &\scriptsize$\pm$0.1498 &\scriptsize$\pm$0.94\% &\scriptsize$\pm$0.6196 &\scriptsize$\pm$0.94\% &\scriptsize$\pm$0.3035 &\scriptsize$\pm$11.26\% &\scriptsize{$\pm$0.0092} &\scriptsize{$\pm$0.34\%}\\        
        
        \cline{1-9}\rule{0pt}{2.3ex}

         \multirow{2}{*}{\pignpi}
        & \textbf{0.0600} & \textbf{0.38\%} & \textbf{0.2483} & \textbf{0.38\%} & \textbf{0.0194} & \textbf{0.72\%} & {0.0270} & {1.00\%}\\
        & \scriptsize \textbf{$\pm$0.0020} &\scriptsize \textbf{$\pm$0.01\%} &\scriptsize \textbf{$\pm$0.0081} &\scriptsize \textbf{$\pm$0.01\%} &\scriptsize \textbf{$\pm$0.0006} &\scriptsize \textbf{$\pm$0.02\%} &\scriptsize {$\pm$0.0008} &\scriptsize {$\pm$0.03\%} \\      
      \bottomrule
\end{tabularx}
}
\end{table}

\begin{table}[H]
  \centering
  \caption{Evaluation of the performance to learn pairwise potential energy for the LJ-argon dataset. Results
averaged across five experiments.}
  \label{table:LJ-argon-potential}
  \setlength\extrarowheight{-6pt}
  \scalebox{0.85}{
  \begin{tabularx}{1.2\textwidth}{cXXXXXXXXXXXX}
    \toprule
  &\makecell[Xt]{\small \textsf{MAE\textsubscript{acc}}\\ \footnotesize (\r{A}/ps$^2$)} &\makecell[Xt]{\small Relative\\ \small \textsf{MAE\textsubscript{acc}}} &\makecell[Xt]{\small \textsf{MAE\textsubscript{nf}}\\ \scriptsize(meV/\r{A})} &\makecell[Xt]{\small Relative\\ \small \textsf{MAE\textsubscript{nf}}} & \makecell[Xt]{\small \textsf{MAE\textsubscript{ef}}\\ \scriptsize(meV/\r{A})} & \makecell[Xt]{\small Relative\\ \small \textsf{MAE\textsubscript{ef}}} & \makecell[Xt]{\small \textsf{MAE\textsubscript{ep}}\\ \footnotesize(meV)} & \makecell[Xt]{\small Relative\\ \small \textsf{MAE\textsubscript{ep}}} & \makecell[Xt]{\small \textsf{MAE\textsubscript{np}}\\ \footnotesize(meV)} & \makecell[Xt]{\small Relative\\ \small \textsf{MAE\textsubscript{np}}} & \makecell[Xt]{\small $\textsf{MAE}_\textsubscript{symm}^{P}$\\ \footnotesize(meV)} &\makecell[Xt]{\small Relative\\ \small $\textsf{MAE}_\textsubscript{symm}^{P}$}\\
    \midrule
         \multirow{2}{0.9cm}{\small Baseline} 
        & 10.8064 & 67.87\% & 44.6875 & 67.87\% & 73.4575 & 2725\% & 13.9532 & 1051\% & 403.866 & 510\% & 21.6873 & 1633\%\\
        &\scriptsize$\pm$0.0113 &\scriptsize$\pm$0.07\% &\scriptsize$\pm$0.0467 &\scriptsize$\pm$0.07\% &\scriptsize$\pm$6.3486 &\scriptsize$\pm$236\% &\scriptsize$\pm$1.1882 &\scriptsize$\pm$89.47\% &\scriptsize$\pm$34.9743 &\scriptsize$\pm$44.2\% &\scriptsize$\pm$1.8809 &\scriptsize$\pm$141.6\%\\
        
        \cline{1-13}\rule{0pt}{2.3ex}

         \multirow{2}{0.9cm}{\small \pignpi}
        & \textbf{0.0714} & \textbf{0.45\%} & \textbf{0.2951} & \textbf{0.45\%} & \textbf{0.0217} & \textbf{0.81\%} & \textbf{0.0176} & \textbf{1.33\%} & \textbf{0.4428} & \textbf{0.56\%} & \textbf{0.0174} & \textbf{1.31\%}\\

        & \scriptsize\textbf{$\pm$0.0057} & \scriptsize\textbf{$\pm$0.04\%} & \scriptsize\textbf{$\pm$0.0238} & \scriptsize\textbf{$\pm$0.04\%} & \scriptsize\textbf{$\pm$0.0016} & \scriptsize\textbf{$\pm$0.06\%} & \scriptsize\textbf{$\pm$0.0014} & \scriptsize\textbf{$\pm$0.11\%} & \scriptsize\textbf{$\pm$0.1061} & \scriptsize\textbf{$\pm$0.13\%} & \scriptsize\textbf{$\pm$0.0020} & \scriptsize\textbf{$\pm$0.15\%}\\      
      \bottomrule
\end{tabularx}
}
\end{table}

\clearpage


\subsection{Evaluation of \pignpi with different activation functions to learn force}
\label{sec:activation_function_force}
Table~\ref{table:activation_functions_force} reports the performance of \pignpi with different activation functions to learn pairwise force. 
\begin{table}[h!]
  \centering
    \caption{Quality of pairwise force prediction of \pignpi with different activation functions. Results averaged across five experiments.}
    \label{table:activation_functions_force}
    \setlength\extrarowheight{-6pt}
  \scalebox{0.9}{
  \begin{tabularx}{\textwidth}{ccXXXXXXXX}
    \toprule
  & &\makecell[Xt]{Spring \\ dim=2} &\makecell[Xt]{Spring \\ dim=3} & \makecell[Xt]{Charge\\ dim=2} & \makecell[Xt]{Charge\\ dim=3} &\makecell[Xt]{Orbital \\ dim=2} &\makecell[Xt]{Orbital \\ dim=3} & \makecell[Xt]{{Discnt} \\ dim=2} &\makecell[Xt]{{Discnt} \\ dim=3}\\
    \midrule
        \multirow{14}{*}{\textsf{MAE\textsubscript{acc}}} 

        & \multirow{2}{*}{SiLU}
        & 0.0206 & 0.0278 & 0.0425 & 0.1191 & 0.0202 & 0.0182 & 0.0227 & 0.0399\\
        &&\scriptsize $\pm$0.0009 & \scriptsize $\pm$0.0021 & \scriptsize $\pm$0.0053 & \scriptsize $\pm$0.0027 & \scriptsize $\pm$0.0003 & \scriptsize $\pm$0.0003 & \scriptsize $\pm$0.0019 & \scriptsize $\pm$0.0011\\
        \cline{3-10}\rule{0pt}{2.3ex}

        & \multirow{2}{*}{ReLU}
        & 0.0339 & 0.0524 & 0.1528 & 0.2058 & 0.0402 & 0.0399 & 0.0463 & 0.0868\\
        &&\scriptsize $\pm$0.0007 & \scriptsize $\pm$0.0009 & \scriptsize $\pm$0.0039 & \scriptsize $\pm$0.0066 & \scriptsize $\pm$0.0028 & \scriptsize $\pm$0.0003 & \scriptsize $\pm$0.0020 & \scriptsize $\pm$0.0026\\
        \cline{3-10}\rule{0pt}{2.3ex}

        & \multirow{2}{*}{GELU}
        & 0.0171 & 0.0189 & 0.0401 & 0.1247 & 0.0212 & 0.0191 & 0.0232 & 0.0388\\
        &&\scriptsize $\pm$0.0009 & \scriptsize $\pm$0.0007 & \scriptsize $\pm$0.0017 & \scriptsize $\pm$0.0077 & \scriptsize $\pm$0.0008 & \scriptsize $\pm$0.0004 & \scriptsize $\pm$0.0030 & \scriptsize $\pm$0.0013\\
        \cline{3-10}\rule{0pt}{2.3ex}

        & \multirow{2}{*}{tanh}
        & 0.0234 & 0.0645 & 0.1713 & 0.3252 & 0.0415 & 0.0860 & 0.0646 & 0.1046\\
        &&\scriptsize $\pm$0.0004 & \scriptsize $\pm$0.0003 & \scriptsize $\pm$0.0388 & \scriptsize $\pm$0.0267 & \scriptsize $\pm$0.0002 & \scriptsize $\pm$0.0015 & \scriptsize $\pm$0.0148 & \scriptsize $\pm$0.0007\\
        \cline{3-10}\rule{0pt}{2.3ex}

        & \multirow{2}{*}{sigmoid}
        & 0.0597 & 0.1618 & 1.1555 & 0.2747 & 0.0381 & 0.0421 & 0.1053 & 0.2588\\
        &&\scriptsize $\pm$0.0046 & \scriptsize $\pm$0.0173 & \scriptsize $\pm$0.0121 & \scriptsize $\pm$0.0455 & \scriptsize $\pm$0.0042 & \scriptsize $\pm$0.0024 & \scriptsize $\pm$0.0014 & \scriptsize $\pm$0.0235\\
        \cline{3-10}\rule{0pt}{2.3ex}

        & \multirow{2}{*}{softplus}
        & 0.0228 & 0.0354 & 0.0647 & 0.0933 & 0.0293 & 0.0302 & 0.0508 & 0.1720\\
        &&\scriptsize $\pm$0.0011 & \scriptsize $\pm$0.0018 & \scriptsize $\pm$0.0071 & \scriptsize $\pm$0.0045 & \scriptsize $\pm$0.0018 & \scriptsize $\pm$0.0012 & \scriptsize $\pm$0.0017 & \scriptsize $\pm$0.0145\\
        \cline{3-10}\rule{0pt}{2.3ex}

        & \multirow{2}{*}{LeakyReLU}
        & 0.0326 & 0.0545 & 0.1477 & 0.2212 & 0.0387 & 0.0396 & 0.0494 & 0.0910\\
        &&\scriptsize $\pm$0.0009 & \scriptsize $\pm$0.0016 & \scriptsize $\pm$0.0036 & \scriptsize $\pm$0.0059 & \scriptsize $\pm$0.0018 & \scriptsize $\pm$0.0005 & \scriptsize $\pm$0.0014 & \scriptsize $\pm$0.0032\\

        \hline\rule{0pt}{2.3ex}

        \multirow{14}{*}{\textsf{MAE\textsubscript{ef}}} 

        & \multirow{2}{*}{SiLU}
        & 0.0063 & 0.0101 & 0.0136 & 0.0363 & 0.0093 & 0.0095 & 0.0040 & 0.0079\\
        &&\scriptsize $\pm$0.0002 & \scriptsize $\pm$0.0007 & \scriptsize $\pm$0.0023 & \scriptsize $\pm$0.0015 & \scriptsize $\pm$0.0002 & \scriptsize $\pm$0.0001 & \scriptsize $\pm$0.0004 & \scriptsize $\pm$0.0002\\
        \cline{3-10}\rule{0pt}{2.3ex}

        & \multirow{2}{*}{ReLU}
        & 0.0146 & 0.0247 & 0.0379 & 0.0574 & 0.0179 & 0.0201 & 0.0088 & 0.0202\\
        &&\scriptsize $\pm$0.0004 & \scriptsize $\pm$0.0006 & \scriptsize $\pm$0.0013 & \scriptsize $\pm$0.0022 & \scriptsize $\pm$0.0012 & \scriptsize $\pm$0.0004 & \scriptsize $\pm$0.0003 & \scriptsize $\pm$0.0006\\
        \cline{3-10}\rule{0pt}{2.3ex}

        & \multirow{2}{*}{GELU}
        & 0.0059 & 0.0079 & 0.0120 & 0.0347 & 0.0097 & 0.0108 & 0.0041 & 0.0077\\
        &&\scriptsize $\pm$0.0003 & \scriptsize $\pm$0.0003 & \scriptsize $\pm$0.0005 & \scriptsize $\pm$0.0020 & \scriptsize $\pm$0.0004 & \scriptsize $\pm$0.0003 & \scriptsize $\pm$0.0006 & \scriptsize $\pm$0.0002\\
        \cline{3-10}\rule{0pt}{2.3ex}

        & \multirow{2}{*}{tanh}
        & 0.0096 & 0.0279 & 0.0363 & 0.0920 & 0.0171 & 0.0350 & 0.0137 & 0.0257\\
        &&\scriptsize $\pm$0.0003 & \scriptsize $\pm$0.0002 & \scriptsize $\pm$0.0068 & \scriptsize $\pm$0.0075 & \scriptsize $\pm$0.0003 & \scriptsize $\pm$0.0007 & \scriptsize $\pm$0.0033 & \scriptsize $\pm$0.0003\\
        \cline{3-10}\rule{0pt}{2.3ex}

        & \multirow{2}{*}{sigmoid}
        & 0.0166 & 0.0477 & 0.2129 & 0.0607 & 0.0160 & 0.0172 & 0.0211 & 0.0654\\
        &&\scriptsize $\pm$0.0014 & \scriptsize $\pm$0.0040 & \scriptsize $\pm$0.0020 & \scriptsize $\pm$0.0079 & \scriptsize $\pm$0.0018 & \scriptsize $\pm$0.0011 & \scriptsize $\pm$0.0003 & \scriptsize $\pm$0.0075\\
        \cline{3-10}\rule{0pt}{2.3ex}

        & \multirow{2}{*}{softplus}
        & 0.0067 & 0.0118 & 0.0181 & 0.0257 & 0.0121 & 0.0141 & 0.0098 & 0.0363\\
        &&\scriptsize $\pm$0.0002 & \scriptsize $\pm$0.0005 & \scriptsize $\pm$0.0021 & \scriptsize $\pm$0.0011 & \scriptsize $\pm$0.0006 & \scriptsize $\pm$0.0003 & \scriptsize $\pm$0.0004 & \scriptsize $\pm$0.0028\\
        \cline{3-10}\rule{0pt}{2.3ex}

        & \multirow{2}{*}{LeakyReLU}
        & 0.0139 & 0.0258 & 0.0352 & 0.0578 & 0.0170 & 0.0197 & 0.0096 & 0.0215\\
        &&\scriptsize $\pm$0.0006 & \scriptsize $\pm$0.0011 & \scriptsize $\pm$0.0008 & \scriptsize $\pm$0.0023 & \scriptsize $\pm$0.0009 & \scriptsize $\pm$0.0006 & \scriptsize $\pm$0.0004 & \scriptsize $\pm$0.0008\\        

        \hline\rule{0pt}{2.3ex}

        \multirow{14}{*}{\textsf{MAE\textsubscript{nf}}} 
        & \multirow{2}{*}{SiLU}
        & 0.0219 & 0.0292 & 0.0488 & 0.1317 & 0.0260 & 0.0233 & 0.0239 & 0.0419\\
        &&\scriptsize $\pm$0.0010 & \scriptsize $\pm$0.0022 & \scriptsize $\pm$0.0059 & \scriptsize $\pm$0.0033 & \scriptsize $\pm$0.0005 & \scriptsize $\pm$0.0004 & \scriptsize $\pm$0.0020 & \scriptsize $\pm$0.0011\\
        \cline{3-10}\rule{0pt}{2.3ex}

        & \multirow{2}{*}{ReLU}
        & 0.0358 & 0.0552 & 0.1694 & 0.2246 & 0.0483 & 0.0494 & 0.0489 & 0.0911\\
        &&\scriptsize $\pm$0.0007 & \scriptsize $\pm$0.0009 & \scriptsize $\pm$0.0046 & \scriptsize $\pm$0.0070 & \scriptsize $\pm$0.0033 & \scriptsize $\pm$0.0004 & \scriptsize $\pm$0.0019 & \scriptsize $\pm$0.0023\\
        \cline{3-10}\rule{0pt}{2.3ex}

        & \multirow{2}{*}{GELU}
        & 0.0182 & 0.0202 & 0.0460 & 0.1366 & 0.0270 & 0.0249 & 0.0244 & 0.0402\\
        &&\scriptsize $\pm$0.0009 & \scriptsize $\pm$0.0007 & \scriptsize $\pm$0.0017 & \scriptsize $\pm$0.0078 & \scriptsize $\pm$0.0010 & \scriptsize $\pm$0.0006 & \scriptsize $\pm$0.0032 & \scriptsize $\pm$0.0012\\
        \cline{3-10}\rule{0pt}{2.3ex}

        & \multirow{2}{*}{tanh}
        & 0.0249 & 0.0682 & 0.1975 & 0.3719 & 0.0510 & 0.1166 & 0.0682 & 0.1097\\
        &&\scriptsize $\pm$0.0006 & \scriptsize $\pm$0.0003 & \scriptsize $\pm$0.0444 & \scriptsize $\pm$0.0303 & \scriptsize $\pm$0.0003 & \scriptsize $\pm$0.0018 & \scriptsize $\pm$0.0157 & \scriptsize $\pm$0.0009\\
        \cline{3-10}\rule{0pt}{2.3ex}

        & \multirow{2}{*}{sigmoid}
        & 0.0629 & 0.1673 & 1.3848 & 0.3233 & 0.0496 & 0.0553 & 0.1111 & 0.2722\\
        &&\scriptsize $\pm$0.0048 & \scriptsize $\pm$0.0179 & \scriptsize $\pm$0.0148 & \scriptsize $\pm$0.0533 & \scriptsize $\pm$0.0054 & \scriptsize $\pm$0.0038 & \scriptsize $\pm$0.0014 & \scriptsize $\pm$0.0243\\
        \cline{3-10}\rule{0pt}{2.3ex}

        & \multirow{2}{*}{softplus}
        & 0.0239 & 0.0367 & 0.0753 & 0.1038 & 0.0366 & 0.0390 & 0.0541 & 0.1829\\
        &&\scriptsize $\pm$0.0012 & \scriptsize $\pm$0.0018 & \scriptsize $\pm$0.0086 & \scriptsize $\pm$0.0051 & \scriptsize $\pm$0.0016 & \scriptsize $\pm$0.0013 & \scriptsize $\pm$0.0018 & \scriptsize $\pm$0.0161\\
        \cline{3-10}\rule{0pt}{2.3ex}

        & \multirow{2}{*}{LeakyReLU}
        & 0.0344 & 0.0573 & 0.1634 & 0.2411 & 0.0464 & 0.0489 & 0.0520 & 0.0951\\
        &&\scriptsize $\pm$0.0010 & \scriptsize $\pm$0.0018 & \scriptsize $\pm$0.0044 & \scriptsize $\pm$0.0058 & \scriptsize $\pm$0.0022 & \scriptsize $\pm$0.0005 & \scriptsize $\pm$0.0016 & \scriptsize $\pm$0.0031\\

        \hline\rule{0pt}{2.3ex}

        \multirow{14}{*}{$\textsf{MAE}_\textsf{symm}^{F}$} 
        & \multirow{2}{*}{SiLU}
        & 0.0075 & 0.0133 & 0.0185 & 0.0345 & 0.0136 & 0.0134 & 0.0026 & 0.0066\\
        &&\scriptsize $\pm$0.0003 & \scriptsize $\pm$0.0008 & \scriptsize $\pm$0.0036 & \scriptsize $\pm$0.0017 & \scriptsize $\pm$0.0004 & \scriptsize $\pm$0.0004 & \scriptsize $\pm$0.0004 & \scriptsize $\pm$0.0001\\
        \cline{3-10}\rule{0pt}{2.3ex}

        & \multirow{2}{*}{ReLU}
        & 0.0205 & 0.0350 & 0.0459 & 0.0477 & 0.0256 & 0.0285 & 0.0104 & 0.0248\\
        &&\scriptsize $\pm$0.0006 & \scriptsize $\pm$0.0009 & \scriptsize $\pm$0.0013 & \scriptsize $\pm$0.0018 & \scriptsize $\pm$0.0016 & \scriptsize $\pm$0.0007 & \scriptsize $\pm$0.0004 & \scriptsize $\pm$0.0008\\
        \cline{3-10}\rule{0pt}{2.3ex}

        & \multirow{2}{*}{GELU}
        & 0.0074 & 0.0108 & 0.0151 & 0.0311 & 0.0138 & 0.0155 & 0.0031 & 0.0071\\
        &&\scriptsize $\pm$0.0003 & \scriptsize $\pm$0.0003 & \scriptsize $\pm$0.0007 & \scriptsize $\pm$0.0013 & \scriptsize $\pm$0.0006 & \scriptsize $\pm$0.0005 & \scriptsize $\pm$0.0003 & \scriptsize $\pm$0.0003\\
        \cline{3-10}\rule{0pt}{2.3ex}

        & \multirow{2}{*}{tanh}
        & 0.0128 & 0.0367 & 0.0242 & 0.0580 & 0.0223 & 0.0363 & 0.0106 & 0.0265\\
        &&\scriptsize $\pm$0.0005 & \scriptsize $\pm$0.0002 & \scriptsize $\pm$0.0013 & \scriptsize $\pm$0.0121 & \scriptsize $\pm$0.0003 & \scriptsize $\pm$0.0008 & \scriptsize $\pm$0.0018 & \scriptsize $\pm$0.0004\\
        \cline{3-10}\rule{0pt}{2.3ex}

        & \multirow{2}{*}{sigmoid}
        & 0.0108 & 0.0337 & 0.0386 & 0.0344 & 0.0194 & 0.0193 & 0.0094 & 0.0318\\
        &&\scriptsize $\pm$0.0006 & \scriptsize $\pm$0.0018 & \scriptsize $\pm$0.0055 & \scriptsize $\pm$0.0013 & \scriptsize $\pm$0.0026 & \scriptsize $\pm$0.0016 & \scriptsize $\pm$0.0004 & \scriptsize $\pm$0.0055\\
        \cline{3-10}\rule{0pt}{2.3ex}

        & \multirow{2}{*}{softplus}
        & 0.0072 & 0.0145 & 0.0217 & 0.0252 & 0.0163 & 0.0195 & 0.0068 & 0.0402\\
        &&\scriptsize $\pm$0.0003 & \scriptsize $\pm$0.0007 & \scriptsize $\pm$0.0021 & \scriptsize $\pm$0.0014 & \scriptsize $\pm$0.0003 & \scriptsize $\pm$0.0004 & \scriptsize $\pm$0.0008 & \scriptsize $\pm$0.0050\\
        \cline{3-10}\rule{0pt}{2.3ex}

        & \multirow{2}{*}{LeakyReLU}
        & 0.0194 & 0.0363 & 0.0463 & 0.0494 & 0.0242 & 0.0279 & 0.0109 & 0.0257\\
        &&\scriptsize $\pm$0.0008 & \scriptsize $\pm$0.0016 & \scriptsize $\pm$0.0014 & \scriptsize $\pm$0.0026 & \scriptsize $\pm$0.0013 & \scriptsize $\pm$0.0008 & \scriptsize $\pm$0.0004 & \scriptsize $\pm$0.0008\\

      \bottomrule
\end{tabularx}
}
\end{table}

\clearpage

\subsection{Performance evaluation of \pignpi with different activation functions  for pairwise potential energy prediction}
\label{sec:activation_function_potential}
Table~\ref{table:activation_functions_potential} reports the performance of \pignpi with different activation functions to learn pairwise potential energy.


\begingroup 
\setlength{\LTleft}{0pt minus \textwidth}
\setlength{\LTright}{0pt minus \textwidth}
\setlength{\LTcapwidth}{\textwidth}

\begin{longtable}[h!]{cccccccccc}
  \caption{Performance evaluation of \pignpi with different activation functions for pairwise potential energy prediction. Results averaged across five experiments.}
  \label{table:activation_functions_potential}\\\toprule
  \endfirsthead
  \caption*{Continued: Performance evaluation of \pignpi with different activation functions for pairwise potential energy prediction..}\\\toprule
  \endhead
  \multicolumn{10}{r}{{(Continued on next page)}}\\ 
  \endfoot
  \multicolumn{10}{r}{{(The end)}}\\ 
  \endlastfoot
  & & \makecell{Spring \\ dim=2} & \makecell{Spring \\ dim=3} & \makecell{Charge\\ dim=2} & \makecell{Charge\\ dim=3} &\makecell{Orbital \\ dim=2} &\makecell{Orbital \\ dim=3} & \makecell{{Discnt} \\ dim=2} &\makecell{{Discnt} \\ dim=3}\\
    \midrule
        \multirow{14}{*}{\textsf{MAE\textsubscript{acc}}} 

        & \multirow{2}{*}{SiLU}
        & {0.0076} & {0.0099} & {0.0225} & {0.1088} & {0.0090} & {0.0091} & {0.0089} & {0.0150}\\
        && \scriptsize {$\pm$0.0003} & \scriptsize {$\pm$0.0007} & \scriptsize {$\pm$0.0012} & \scriptsize {$\pm$0.0079} & \scriptsize {$\pm$0.0004} & \scriptsize {$\pm$0.0004} & \scriptsize {$\pm$0.0002} & \scriptsize {$\pm$0.0022}\\
        \cline{3-10}\rule{0pt}{2.3ex}

        & \multirow{2}{*}{ReLU}
        &3.2521 & 5.0524 & 6.2996 & 1.8127 & 5.6495 & 3.8274 & 1.5069 & 2.8088\\
        &&\scriptsize $\pm$0.0818&\scriptsize$\pm$0.0796&\scriptsize$\pm$0.0043&\scriptsize$\pm$0.0017&\scriptsize$\pm$0.0788&\scriptsize$\pm$0.0224&\scriptsize$\pm$0.0501&\scriptsize$\pm$0.0224\\
        \cline{3-10}\rule{0pt}{2.3ex}

        & \multirow{2}{*}{GELU}
        &0.0063 & 0.0054 & 0.0298 & 0.1586 & 0.0089 & 0.0098 & 0.0104 & 0.0154\\
        &&\scriptsize $\pm$0.0004&\scriptsize$\pm$0.0003&\scriptsize$\pm$0.0010&\scriptsize$\pm$0.0061&\scriptsize$\pm$0.0001&\scriptsize$\pm$0.0004&\scriptsize$\pm$0.0009&\scriptsize$\pm$0.0005\\
        \cline{3-10}\rule{0pt}{2.3ex}

        & \multirow{2}{*}{tanh}
        &0.0223 & 0.0366 & 0.0499 & 0.1949 & 0.0139 & 0.0187 & 0.0330 & 0.0889\\
        &&\scriptsize $\pm$0.0023&\scriptsize$\pm$0.0054&\scriptsize$\pm$0.0016&\scriptsize$\pm$0.0112&\scriptsize$\pm$0.0001&\scriptsize$\pm$0.0006&\scriptsize$\pm$0.0010&\scriptsize$\pm$0.0104\\
        \cline{3-10}\rule{0pt}{2.3ex}

        & \multirow{2}{*}{sigmoid}
        &0.0949 & 0.0432 & 0.0631 & 0.1022 & 0.0206 & 0.0290 & 0.0299 & 0.0638\\
        &&\scriptsize $\pm$0.0915&\scriptsize$\pm$0.0038&\scriptsize$\pm$0.0029&\scriptsize$\pm$0.0092&\scriptsize$\pm$0.0012&\scriptsize$\pm$0.0015&\scriptsize$\pm$0.0014&\scriptsize$\pm$0.0026\\
        \cline{3-10}\rule{0pt}{2.3ex}

        & \multirow{2}{*}{softplus}
        &0.0259 & 0.0271 & 0.0516 & 0.0870 & 0.0113 & 0.0161 & 0.0274 & 0.0430\\
        &&\scriptsize $\pm$0.0029&\scriptsize$\pm$0.0019&\scriptsize$\pm$0.0034&\scriptsize$\pm$0.0049&\scriptsize$\pm$0.0005&\scriptsize$\pm$0.0018&\scriptsize$\pm$0.0023&\scriptsize$\pm$0.0021\\
        \cline{3-10}\rule{0pt}{2.3ex}

        & \multirow{2}{*}{LeakyReLU}
        &3.3248 & 5.0494 & 6.2987 & 1.8135 & 5.6056 & 3.8106 & 1.5031 & 2.7777\\
        &&\scriptsize $\pm$0.0626&\scriptsize$\pm$0.0265&\scriptsize$\pm$0.0070&\scriptsize$\pm$0.0017&\scriptsize$\pm$0.0270&\scriptsize$\pm$0.0647&\scriptsize$\pm$0.0369&\scriptsize$\pm$0.0408\\
        
        \hline\rule{0pt}{2.3ex}

        \multirow{14}{*}{\textsf{MAE\textsubscript{ef}}} 
        & \multirow{2}{*}{SiLU}
        & {0.0023} & {0.0037} & {0.0080} & {0.0223} & {0.0058} & {0.0053} & {0.0016} & {0.0030} \\
        && \scriptsize {$\pm$0.0001} & \scriptsize {$\pm$0.0003} & \scriptsize {$\pm$0.0006} & \scriptsize {$\pm$0.0013} & \scriptsize {$\pm$0.0011} & \scriptsize {$\pm$0.0005} & \tiny{$\pm$3.4E-5} & \scriptsize {$\pm$0.0004}\\
        \cline{3-10}\rule{0pt}{2.3ex}

        & \multirow{2}{*}{ReLU}
        &1.1132 & 1.5625 & 1.2551 & 0.3854 & 1.9323 & 1.3612 & 0.5471 & 0.9503\\
        &&\scriptsize $\pm$0.0243&\scriptsize$\pm$0.0287&\scriptsize$\pm$0.0027&\scriptsize$\pm$0.0007&\scriptsize$\pm$0.0265&\scriptsize$\pm$0.0129&\scriptsize$\pm$0.0178&\scriptsize$\pm$0.0067\\
        \cline{3-10}\rule{0pt}{2.3ex}

        & \multirow{2}{*}{GELU}
        &0.0020 & 0.0029 & 0.0114 & 0.0312 & 0.0054 & 0.0062 & 0.0019 & 0.0032\\
        &&\scriptsize $\pm$0.0001&\scriptsize$\pm$0.0007&\scriptsize$\pm$0.0001&\scriptsize$\pm$0.0016&\scriptsize$\pm$0.0008&\scriptsize$\pm$0.0004&\scriptsize$\pm$0.0002&\scriptsize$\pm$0.0001\\
        \cline{3-10}\rule{0pt}{2.3ex}

        & \multirow{2}{*}{tanh}
        &0.0081 & 0.0136 & 0.0214 & 0.0719 & 0.0109 & 0.0150 & 0.0069 & 0.0223\\
        &&\scriptsize $\pm$0.0011&\scriptsize$\pm$0.0014&\scriptsize$\pm$0.0006&\scriptsize$\pm$0.0016&\scriptsize$\pm$0.0015&\scriptsize$\pm$0.0011&\scriptsize$\pm$0.0002&\scriptsize$\pm$0.0027\\
        \cline{3-10}\rule{0pt}{2.3ex}

        & \multirow{2}{*}{sigmoid}
        &0.0267 & 0.0123 & 0.0210 & 0.0292 & 0.0113 & 0.0139 & 0.0058 & 0.0143\\
        &&\scriptsize $\pm$0.0290&\scriptsize$\pm$0.0010&\scriptsize$\pm$0.0013&\scriptsize$\pm$0.0026&\scriptsize$\pm$0.0007&\scriptsize$\pm$0.0009&\scriptsize$\pm$0.0003&\scriptsize$\pm$0.0006\\
        \cline{3-10}\rule{0pt}{2.3ex}

        & \multirow{2}{*}{softplus}
        &0.0068 & 0.0103 & 0.0202 & 0.0260 & 0.0073 & 0.0084 & 0.0054 & 0.0097\\
        &&\scriptsize $\pm$0.0007&\scriptsize$\pm$0.0009&\scriptsize$\pm$0.0023&\scriptsize$\pm$0.0022&\scriptsize$\pm$0.0012&\scriptsize$\pm$0.0009&\scriptsize$\pm$0.0005&\scriptsize$\pm$0.0006\\
        \cline{3-10}\rule{0pt}{2.3ex}

        & \multirow{2}{*}{LeakyReLU}
        &1.1325 & 1.5550 & 1.2511 & 0.3857 & 1.9077 & 1.3596 & 0.5444 & 0.9422\\
        &&\scriptsize $\pm$0.0143&\scriptsize$\pm$0.0120&\scriptsize$\pm$0.0026&\scriptsize$\pm$0.0011&\scriptsize$\pm$0.0125&\scriptsize$\pm$0.0250&\scriptsize$\pm$0.0174&\scriptsize$\pm$0.0112\\

        \hline\rule{0pt}{2.3ex}

        \multirow{14}{*}{\textsf{MAE\textsubscript{nf}}} 
        & \multirow{2}{*}{SiLU}
        & {0.0080} & {0.0104} & {0.0261} & {0.1212} & {0.0115} & {0.0118} & {0.0098} & {0.0160} \\
        && \scriptsize {$\pm$0.0003} & \scriptsize {$\pm$0.0007} & \scriptsize {$\pm$0.0014} & \scriptsize {$\pm$0.0085} & \scriptsize {$\pm$0.0006} & \scriptsize {$\pm$0.0005} & \scriptsize {$\pm$0.0002} & \scriptsize {$\pm$0.0023}\\
        \cline{3-10}\rule{0pt}{2.3ex}

        & \multirow{2}{*}{ReLU}
        &3.3210 & 5.1395 & 7.1663 & 1.9481 & 7.2113 & 4.9279 & 1.5787 & 2.8956\\
        &&\scriptsize $\pm$0.0732&\scriptsize$\pm$0.0846&\scriptsize$\pm$0.0054&\scriptsize$\pm$0.0020&\scriptsize$\pm$0.0851&\scriptsize$\pm$0.0347&\scriptsize$\pm$0.0533&\scriptsize$\pm$0.0199\\
        \cline{3-10}\rule{0pt}{2.3ex}

        & \multirow{2}{*}{GELU}
        &0.0067 & 0.0059 & 0.0347 & 0.1757 & 0.0114 & 0.0128 & 0.0111 & 0.0162\\
        &&\scriptsize $\pm$0.0005&\scriptsize$\pm$0.0003&\scriptsize$\pm$0.0011&\scriptsize$\pm$0.0071&\scriptsize$\pm$0.0003&\scriptsize$\pm$0.0006&\scriptsize$\pm$0.0009&\scriptsize$\pm$0.0004\\
        \cline{3-10}\rule{0pt}{2.3ex}

        & \multirow{2}{*}{tanh}
        &0.0235 & 0.0392 & 0.0578 & 0.2202 & 0.0176 & 0.0236 & 0.0355 & 0.0944\\
        &&\scriptsize $\pm$0.0023&\scriptsize$\pm$0.0056&\scriptsize$\pm$0.0018&\scriptsize$\pm$0.0131&\scriptsize$\pm$0.0003&\scriptsize$\pm$0.0007&\scriptsize$\pm$0.0010&\scriptsize$\pm$0.0110\\
        \cline{3-10}\rule{0pt}{2.3ex}

        & \multirow{2}{*}{sigmoid}
        &0.0993 & 0.0447 & 0.0738 & 0.1137 & 0.0268 & 0.0369 & 0.0320 & 0.0672\\
        &&\scriptsize $\pm$0.0957&\scriptsize$\pm$0.0038&\scriptsize$\pm$0.0034&\scriptsize$\pm$0.0095&\scriptsize$\pm$0.0016&\scriptsize$\pm$0.0017&\scriptsize$\pm$0.0015&\scriptsize$\pm$0.0029\\
        \cline{3-10}\rule{0pt}{2.3ex}

        & \multirow{2}{*}{softplus}
        &0.0268 & 0.0283 & 0.0599 & 0.0964 & 0.0148 & 0.0206 & 0.0292 & 0.0454\\
        &&\scriptsize $\pm$0.0030&\scriptsize$\pm$0.0019&\scriptsize$\pm$0.0037&\scriptsize$\pm$0.0059&\scriptsize$\pm$0.0006&\scriptsize$\pm$0.0019&\scriptsize$\pm$0.0023&\scriptsize$\pm$0.0025\\
        \cline{3-10}\rule{0pt}{2.3ex}

        & \multirow{2}{*}{LeakyReLU}
        &3.3932 & 5.1328 & 7.1647 & 1.9487 & 7.1476 & 4.8920 & 1.5798 & 2.8733\\
        &&\scriptsize $\pm$0.0640&\scriptsize$\pm$0.0205&\scriptsize$\pm$0.0082&\scriptsize$\pm$0.0022&\scriptsize$\pm$0.0364&\scriptsize$\pm$0.0795&\scriptsize$\pm$0.0339&\scriptsize$\pm$0.0387\\

        \hline\rule{0pt}{2.3ex} 


        \multirow{14}{*}{\textsf{MAE\textsubscript{$\Delta$ep}}} 
        & \multirow{2}{*}{SiLU}
        & {0.0005} & {0.0016} & {0.0096} & {0.0156} & {0.0048} & {0.0031} & {0.2197} & {0.2344} \\
        && \tiny {$\pm$1.4E-5} & \scriptsize {$\pm$0.0003} & \scriptsize {$\pm$0.0009} & \scriptsize {$\pm$0.0015} & \scriptsize {$\pm$0.0017} & \scriptsize {$\pm$0.0006} & \scriptsize {$\pm$0.0001} & \scriptsize {$\pm$0.0001}\\
        \cline{3-10}\rule{0pt}{2.3ex}

        & \multirow{2}{*}{ReLU}
        &1.8798 & 5.8125 & 0.4592 & 0.2061 & 1.3700 & 1.0480 & 0.6844 & 1.4246\\
        &&\scriptsize $\pm$0.2632&\scriptsize$\pm$0.2222&\scriptsize$\pm$0.0115&\scriptsize$\pm$0.0042&\scriptsize$\pm$0.1386&\scriptsize$\pm$0.0589&\scriptsize$\pm$0.0786&\scriptsize$\pm$0.1339\\
        \cline{3-10}\rule{0pt}{2.3ex}

        & \multirow{2}{*}{GELU}
        &0.0006 & 0.0017 & 0.0145 & 0.0156 & 0.0034 & 0.0037 & 0.2197 & 0.2344\\
        &&\scriptsize $\pm$0.0001&\scriptsize$\pm$0.0007&\scriptsize$\pm$0.0007&\scriptsize$\pm$0.0020&\scriptsize$\pm$0.0011&\scriptsize$\pm$0.0006&\scriptsize$\pm$0.0001&\scriptsize$\pm$0.0001\\
        \cline{3-10}\rule{0pt}{2.3ex}

        & \multirow{2}{*}{tanh}
        &0.0030 & 0.0037 & 0.0312 & 0.0803 & 0.0080 & 0.0102 & 0.2202 & 0.2353\\
        &&\scriptsize $\pm$0.0013&\scriptsize$\pm$0.0008&\scriptsize$\pm$0.0008&\scriptsize$\pm$0.0040&\scriptsize$\pm$0.0019&\scriptsize$\pm$0.0010&\scriptsize$\pm$0.0001&\scriptsize$\pm$0.0002\\
        \cline{3-10}\rule{0pt}{2.3ex}

        & \multirow{2}{*}{sigmoid}
        &0.0068 & 0.0025 & 0.0271 & 0.0298 & 0.0068 & 0.0074 & 0.2199 & 0.2349\\
        &&\scriptsize $\pm$0.0094&\scriptsize$\pm$0.0002&\scriptsize$\pm$0.0033&\scriptsize$\pm$0.0048&\scriptsize$\pm$0.0012&\scriptsize$\pm$0.0014&\scriptsize$\pm$0.0002&\scriptsize$\pm$0.0005\\
        \cline{3-10}\rule{0pt}{2.3ex}

        & \multirow{2}{*}{softplus}
        &0.0015 & 0.0080 & 0.0315 & 0.0369 & 0.0079 & 0.0065 & 0.2199 & 0.2349\\
        &&\scriptsize $\pm$0.0005&\scriptsize$\pm$0.0016&\scriptsize$\pm$0.0059&\scriptsize$\pm$0.0041&\scriptsize$\pm$0.0016&\scriptsize$\pm$0.0013&\scriptsize$\pm$0.0001&\scriptsize$\pm$0.0002\\
        \cline{3-10}\rule{0pt}{2.3ex}

        & \multirow{2}{*}{LeakyReLU}
        &2.0263 & 5.2505 & 0.4646 & 0.2081 & 1.5073 & 1.1100 & 0.7408 & 1.3386\\
        &&\scriptsize $\pm$0.3663&\scriptsize$\pm$0.6543&\scriptsize$\pm$0.0083&\scriptsize$\pm$0.0034&\scriptsize$\pm$0.1474&\scriptsize$\pm$0.0630&\scriptsize$\pm$0.1149&\scriptsize$\pm$0.1336\\

        \hline\rule{0pt}{2.3ex}

        \multirow{14}{*}{\textsf{MAE\textsubscript{$\Delta$np}}} 
        & \multirow{2}{*}{SiLU}
        & {0.0016} & {0.0062} & {0.0179} & {0.0381} & {0.0129} & {0.0074} & {0.9319} & {0.9322} \\
        && \scriptsize {$\pm$0.0001} & \scriptsize {$\pm$0.0012} & \scriptsize {$\pm$0.0011} & \scriptsize {$\pm$0.0037} & \scriptsize {$\pm$0.0046} & \scriptsize {$\pm$0.0009} & \scriptsize {$\pm$0.0005} & \scriptsize {$\pm$0.0004}\\
        \cline{3-10}\rule{0pt}{2.3ex}

        & \multirow{2}{*}{ReLU}
        &8.6263 & 22.7924 & 1.5514 & 0.6727 & 5.3176 & 4.0300 & 3.2903 & 6.6232\\
        &&\scriptsize $\pm$1.4047&\scriptsize$\pm$2.6079&\scriptsize$\pm$0.0545&\scriptsize$\pm$0.0154&\scriptsize$\pm$1.0821&\scriptsize$\pm$0.2800&\scriptsize$\pm$0.4927&\scriptsize$\pm$1.1136\\
        \cline{3-10}\rule{0pt}{2.3ex}

        & \multirow{2}{*}{GELU}
        &0.0015 & 0.0044 & 0.0263 & 0.0441 & 0.0085 & 0.0099 & 0.9322 & 0.9319\\
        &&\scriptsize $\pm$0.0002&\scriptsize$\pm$0.0020&\scriptsize$\pm$0.0013&\scriptsize$\pm$0.0053&\scriptsize$\pm$0.0026&\scriptsize$\pm$0.0020&\scriptsize$\pm$0.0003&\scriptsize$\pm$0.0002\\
        \cline{3-10}\rule{0pt}{2.3ex}

        & \multirow{2}{*}{tanh}
        &0.0068 & 0.0097 & 0.0469 & 0.1859 & 0.0173 & 0.0234 & 0.9338 & 0.9338\\
        &&\scriptsize $\pm$0.0016&\scriptsize$\pm$0.0009&\scriptsize$\pm$0.0023&\scriptsize$\pm$0.0073&\scriptsize$\pm$0.0019&\scriptsize$\pm$0.0029&\scriptsize$\pm$0.0009&\scriptsize$\pm$0.0014\\
        \cline{3-10}\rule{0pt}{2.3ex}

        & \multirow{2}{*}{sigmoid}
        &0.0220 & 0.0076 & 0.0398 & 0.0689 & 0.0203 & 0.0211 & 0.9324 & 0.9331\\
        &&\scriptsize $\pm$0.0312&\scriptsize$\pm$0.0007&\scriptsize$\pm$0.0040&\scriptsize$\pm$0.0111&\scriptsize$\pm$0.0044&\scriptsize$\pm$0.0024&\scriptsize$\pm$0.0011&\scriptsize$\pm$0.0017\\
        \cline{3-10}\rule{0pt}{2.3ex}

        & \multirow{2}{*}{softplus}
        &0.0051 & 0.0339 & 0.0463 & 0.0794 & 0.0331 & 0.0307 & 0.9323 & 0.9330\\
        &&\scriptsize $\pm$0.0017&\scriptsize$\pm$0.0095&\scriptsize$\pm$0.0069&\scriptsize$\pm$0.0108&\scriptsize$\pm$0.0107&\scriptsize$\pm$0.0097&\scriptsize$\pm$0.0005&\scriptsize$\pm$0.0012\\
        \cline{3-10}\rule{0pt}{2.3ex}

        & \multirow{2}{*}{LeakyReLU}
        &10.3190 & 22.1011 & 1.5917 & 0.6743 & 6.1927 & 4.2487 & 3.4278 & 6.1117\\
        &&\scriptsize $\pm$2.5713&\scriptsize$\pm$4.2016&\scriptsize$\pm$0.0667&\scriptsize$\pm$0.0250&\scriptsize$\pm$1.0489&\scriptsize$\pm$0.3877&\scriptsize$\pm$1.0505&\scriptsize$\pm$1.1812\\

        \hline\rule{0pt}{2.3ex}

        \multirow{14}{*}{$\textsf{MAE}_\textsf{symm}^{P}$} 
        & \multirow{2}{*}{SiLU}
        & {0.0007} & {0.0025} & {0.0074} & {0.0062} & {0.0252} & {0.0422} & {0.0003} & {0.0005}\\
        && \scriptsize {$\pm$0.0001} & \scriptsize {$\pm$0.0008} & \scriptsize {$\pm$0.0010} & \scriptsize {$\pm$0.0005} & \scriptsize {$\pm$0.0065} & \scriptsize {$\pm$0.0144} & \tiny {$\pm$2.1E-5} & \tiny {$\pm$2.6E-5}\\
        \cline{3-10}\rule{0pt}{2.3ex}

        & \multirow{2}{*}{ReLU}
        &1.8823 & 3.6725 & 0.1467 & 0.0702 & 0.9496 & 0.7473 & 0.5693 & 1.1960\\
        &&\scriptsize $\pm$0.4517&\scriptsize$\pm$0.8805&\scriptsize$\pm$0.0339&\scriptsize$\pm$0.0076&\scriptsize$\pm$0.0576&\scriptsize$\pm$0.0270&\scriptsize$\pm$0.1220&\scriptsize$\pm$0.2500\\
        \cline{3-10}\rule{0pt}{2.3ex}

        & \multirow{2}{*}{GELU}
        &0.0009 & 0.0047 & 0.0090 & 0.0078 & 0.0521 & 0.0460 & 0.0004 & 0.0008\\
        &&\scriptsize $\pm$0.0005&\scriptsize$\pm$0.0022&\scriptsize$\pm$0.0005&\scriptsize$\pm$0.0009&\scriptsize$\pm$0.0149&\scriptsize$\pm$0.0176&\scriptsize$\pm$0.0000&\scriptsize$\pm$0.0000\\
        \cline{3-10}\rule{0pt}{2.3ex}

        & \multirow{2}{*}{tanh}
        &0.0202 & 0.0056 & 0.0155 & 0.0215 & 0.1968 & 0.1172 & 0.0010 & 0.0024\\
        &&\scriptsize $\pm$0.0327&\scriptsize$\pm$0.0022&\scriptsize$\pm$0.0019&\scriptsize$\pm$0.0007&\scriptsize$\pm$0.0592&\scriptsize$\pm$0.0242&\scriptsize$\pm$0.0001&\scriptsize$\pm$0.0001\\
        \cline{3-10}\rule{0pt}{2.3ex}

        & \multirow{2}{*}{sigmoid}
        &0.0059 & 0.0028 & 0.0237 & 0.0130 & 0.0294 & 0.0751 & 0.0006 & 0.0017\\
        &&\scriptsize $\pm$0.0076&\scriptsize$\pm$0.0003&\scriptsize$\pm$0.0071&\scriptsize$\pm$0.0010&\scriptsize$\pm$0.0127&\scriptsize$\pm$0.0181&\scriptsize$\pm$0.0000&\scriptsize$\pm$0.0001\\
        \cline{3-10}\rule{0pt}{2.3ex}

        & \multirow{2}{*}{softplus}
        &0.0017 & 0.0112 & 0.0151 & 0.0115 & 0.0481 & 0.0328 & 0.0007 & 0.0015\\
        &&\scriptsize $\pm$0.0007&\scriptsize$\pm$0.0041&\scriptsize$\pm$0.0070&\scriptsize$\pm$0.0012&\scriptsize$\pm$0.0427&\scriptsize$\pm$0.0257&\scriptsize$\pm$0.0001&\scriptsize$\pm$0.0001\\
        \cline{3-10}\rule{0pt}{2.3ex}

        & \multirow{2}{*}{LeakyReLU}
        &2.1660 & 3.2254 & 0.1492 & 0.0688 & 0.9544 & 0.7962 & 0.5670 & 0.9258\\
        &&\scriptsize $\pm$0.5849&\scriptsize$\pm$1.0211&\scriptsize$\pm$0.0203&\scriptsize$\pm$0.0052&\scriptsize$\pm$0.0521&\scriptsize$\pm$0.0972&\scriptsize$\pm$0.2456&\scriptsize$\pm$0.1685\\
        
      \bottomrule

\end{longtable}
\clearpage

\subsection{Imposing symmetry regularization on the baseline model to learn force}
\label{sec:symmetry_for_baseline}


Table~\ref{tab:symmetry_regularization_result} reports the performance of the baseline model with symmetry regularization (see the discussion in  Sec.~\ref{sec:results_pignpi_vs_alternatives} and Sec.~\ref{sec:method_symmetry_regularization}). Results show that such symmetry regularization improves the performance of the baseline model with respect to $\textsf{MAE}_\textsf{symm}^{F}$, which was expected since the symmetry term was minimized. Furthermore, the symmetry regularization makes the baseline model perform better in terms of \textsf{MAE\textsubscript{acc}}, \textsf{MAE\textsubscript{ef}} and \textsf{MAE\textsubscript{nf}} on several datasets. However, \pignpi still significantly outperforms the extended baseline in terms of \textsf{MAE\textsubscript{acc}}, \textsf{MAE\textsubscript{ef}} and \textsf{MAE\textsubscript{nf}}, which are the most relevant performance evaluation metrics for physics-consistent particle interactions. 



\begin{table}[h!]
  \centering
    \caption{Comparison of pairwise force prediction of the baseline model, extended baseline model with symmetry regularization with different weights and \pignpi. Results averaged across five experiments.}
    \label{tab:symmetry_regularization_result}
    \setlength\extrarowheight{-6pt}
  \scalebox{0.95}{
  \begin{tabularx}{\textwidth}{ccX X X X X X X X}
    \toprule
  & &\makecell[Xt]{Spring \\ dim=2} &\makecell[Xt]{Spring \\ dim=3} & \makecell[Xt]{Charge\\ dim=2} & \makecell[Xt]{Charge\\ dim=3} &\makecell[Xt]{Orbital \\ dim=2} &\makecell[Xt]{Orbital \\ dim=3} & \makecell[Xt]{{Discnt} \\ dim=2} &\makecell[Xt]{{Discnt} \\ dim=3}\\
    \midrule

        \multirow{12}{*}{\textsf{MAE\textsubscript{acc}}} 
        &\multirow{2}{*}{Baseline}
        &0.0565 & 0.1076 & 0.2521 & 0.3824 & 0.0437 & 0.0439 & 0.0592 & 0.1171\\ 
        &&\scriptsize $\pm$0.0023 & \scriptsize $\pm$0.0012 & \scriptsize $\pm$0.0173 & \scriptsize $\pm$0.0559 & \scriptsize $\pm$0.0026 & \scriptsize $\pm$0.0014 & \scriptsize \scriptsize $\pm$0.0015 & \scriptsize $\pm$0.0010\\ 
        \cline{3-10}\rule{0pt}{2.3ex}

        & \multirow{2}{*}{$\alpha=0.1$}
        & 0.0756 & 0.1390 & 0.2611 & 0.2864 & 0.0544 & 0.0567 & 0.0623 & 0.1256\\
        &&\scriptsize$\pm$0.0015 & \scriptsize $\pm$0.0019 & \scriptsize $\pm$0.0279 & \scriptsize $\pm$0.0117 & \scriptsize $\pm$0.0017 & \scriptsize $\pm$0.0007 & \scriptsize $\pm$0.0016 & \scriptsize $\pm$0.0020\\

        \cline{3-10}\rule{0pt}{2.3ex}
        & \multirow{2}{*}{$\alpha=1.0$}
        & 0.0743 & 0.1465 & 0.2431 & 0.3121 & 0.0799 & 0.0769 & 0.0571 & 0.1135\\
        &&\scriptsize$\pm$0.0026 & \scriptsize $\pm$0.0013 & \scriptsize $\pm$0.0092 & \scriptsize $\pm$0.0436 & \scriptsize $\pm$0.0022 & \scriptsize $\pm$0.0014 & \scriptsize $\pm$0.0027 & \scriptsize $\pm$0.0026\\

        \cline{3-10}\rule{0pt}{2.3ex}
        & \multirow{2}{*}{$\alpha=10$}
        & 0.0676 & 0.1214 & 5.4372 & 0.7785 & 0.0769 & 0.0740 & 0.0538 & 0.1017\\
        &&\scriptsize$\pm$0.0012 & \scriptsize $\pm$0.0018 & \scriptsize $\pm$0.1597 & \scriptsize $\pm$0.0027 & \scriptsize $\pm$0.0037 & \scriptsize $\pm$0.0012 & \scriptsize $\pm$0.0019 & \scriptsize $\pm$0.0039\\

        \cline{3-10}\rule{0pt}{2.3ex}
        & \multirow{2}{*}{$\alpha=100$}
        & 0.0770 & 0.1381 & 5.5282 & 0.7758 & 0.0902 & 0.1068 & 0.0573 & 0.1064\\
        &&\scriptsize$\pm$0.0027 & \scriptsize $\pm$0.0024 & \scriptsize $\pm$0.0249 & \scriptsize $\pm$0.0042 & \scriptsize $\pm$0.0013 & \scriptsize $\pm$0.0028 & \scriptsize $\pm$0.0022 & \scriptsize $\pm$0.0017\\
        \cline{3-10}\rule{0pt}{2.3ex}
        
        &\multirow{2}{*}{\pignpi}
        &\textbf{0.0206} & \textbf{0.0278} & \textbf{0.0425} & \textbf{0.1191} & \textbf{0.0202} & \textbf{0.0182} & \textbf{0.0227} & \textbf{0.0399}\\
        &&\scriptsize\textbf{$\pm$0.0009} & \scriptsize \textbf{$\pm$0.0021} & \scriptsize \textbf{$\pm$0.0053} & \scriptsize \textbf{$\pm$0.0027} & \scriptsize \textbf{$\pm$0.0003} & \scriptsize \textbf{$\pm$0.0003} & \scriptsize \textbf{$\pm$0.0019} & \scriptsize \textbf{$\pm$0.0011}\\
        \hline\rule{0pt}{2.3ex}
    
        \multirow{12}{*}{  \textsf{MAE\textsubscript{ef}}}

        &\multirow{2}{*}{Baseline}
        &2.3979 & 3.8952 & 1.1832 & 0.6447 & 4.1010 & 3.5379 & 1.6536 & 2.5803\\
        &&\scriptsize$\pm$0.2095 & \scriptsize $\pm$0.7178 & \scriptsize $\pm$0.0955 & \scriptsize $\pm$0.1118 & \scriptsize $\pm$0.1467 & \scriptsize $\pm$0.7571 & \scriptsize $\pm$0.0640 & \scriptsize $\pm$0.2886\\
        \cline{3-10}\rule{0pt}{2.3ex}
        
        & \multirow{2}{*}{$\alpha=0.1$}
        & 1.6465 & 2.6250 & 1.2490 & 0.3751 & 2.4493 & 1.7196 & 0.5302 & 1.2120\\
        &&\scriptsize$\pm$0.1523 & \scriptsize $\pm$0.1347 & \scriptsize $\pm$0.1001 & \scriptsize $\pm$0.0167 & \scriptsize $\pm$0.1253 & \scriptsize $\pm$0.1193 & \scriptsize $\pm$0.0191 & \scriptsize $\pm$0.0456\\
        \cline{3-10}\rule{0pt}{2.3ex}
        
        & \multirow{2}{*}{$\alpha=1.0$}
        & 1.5304 & 2.4136 & 1.2811 & 0.3754 & 2.3558 & 1.7502 & 0.5250 & 0.9572\\
        &&\scriptsize$\pm$0.1035 & \scriptsize $\pm$0.0925 & \scriptsize $\pm$0.0343 & \scriptsize $\pm$0.0043 & \scriptsize $\pm$0.1536 & \scriptsize $\pm$0.0739 & \scriptsize $\pm$0.0861 & \scriptsize $\pm$0.0394\\

        \cline{3-10}\rule{0pt}{2.3ex}
        & \multirow{2}{*}{$\alpha=10$}
        & 1.6178 & 2.3723 & 1.2531 & 0.3790 & 2.3495 & 1.7543 & 0.4819 & 0.9746\\
        &&\scriptsize$\pm$0.0424 & \scriptsize $\pm$0.0327 & \scriptsize $\pm$0.0004 & \scriptsize $\pm$1E-11 & \scriptsize $\pm$0.0267 & \scriptsize $\pm$0.0317 & \scriptsize $\pm$0.0125 & \scriptsize $\pm$0.0191\\

        \cline{3-10}\rule{0pt}{2.3ex}
        & \multirow{2}{*}{$\alpha=100$}
        & 1.5694 & 2.3943 & 1.2528 & 0.3790 & 2.3632 & 1.7505 & 0.4893 & 0.9796\\        
        &&\scriptsize$\pm$0.0167 & \scriptsize $\pm$0.0078 & \scriptsize $\pm$2E-11 & \scriptsize $\pm$9E-12 & \scriptsize $\pm$0.0077 & \scriptsize $\pm$0.0066 & \scriptsize $\pm$0.0053 & \scriptsize $\pm$0.0056\\
        
        \cline{3-10}\rule{0pt}{2.3ex}
        
        &\multirow{2}{*}{\pignpi}
        &\textbf{0.0063} & \textbf{0.0101} & \textbf{0.0136} & \textbf{0.0363} & \textbf{0.0093} & \textbf{0.0095} & \textbf{0.0040} & \textbf{0.0079}\\
        &&\scriptsize\textbf{$\pm$0.0002} & \scriptsize \textbf{$\pm$0.0007} & \scriptsize \textbf{$\pm$0.0023} & \scriptsize \textbf{$\pm$0.0015} & \scriptsize \textbf{$\pm$0.0002} & \scriptsize \textbf{$\pm$0.0001} & \scriptsize \textbf{$\pm$0.0004} & \scriptsize \textbf{$\pm$0.0002}\\
        \hline\rule{0pt}{2.3ex}

        \multirow{12}{*}{ \textsf{MAE\textsubscript{nf}}}

        &\multirow{2}{*}{Baseline}
        &11.652 & 20.967 & 6.831 & 3.804 & 18.194 & 16.677 & 10.786 & 15.651\\
        &&\scriptsize$\pm$0.9890 & \scriptsize $\pm$3.8552 & \scriptsize $\pm$0.5548 & \scriptsize $\pm$0.7523 & \scriptsize $\pm$0.6884 & \scriptsize $\pm$3.5212 & \scriptsize $\pm$0.3764 & \scriptsize $\pm$1.7983\\
        \cline{3-10}\rule{0pt}{2.3ex}

        & \multirow{2}{*}{$\alpha=0.1$}
        & 7.9466 & 14.104 & 7.2125 & 1.9825 & 10.675 & 7.9656 & 3.4844 & 7.4980\\
        &&\scriptsize$\pm$0.7318 & \scriptsize $\pm$0.7268 & \scriptsize $\pm$0.5854 & \scriptsize $\pm$0.0927 & \scriptsize $\pm$0.5594 & \scriptsize $\pm$0.5602 & \scriptsize $\pm$0.1278 & \scriptsize $\pm$0.2871\\

        \cline{3-10}\rule{0pt}{2.3ex}
        
        & \multirow{2}{*}{$\alpha=1.0$}
        & 7.3867 & 12.954 & 7.3978 & 1.9827 & 10.257 & 8.1068 & 3.4632 & 5.9257\\
        &&\scriptsize$\pm$0.5031 & \scriptsize $\pm$0.4993 & \scriptsize $\pm$0.2000 & \scriptsize $\pm$0.0248 & \scriptsize $\pm$0.6695 & \scriptsize $\pm$0.3377 & \scriptsize $\pm$0.5716 & \scriptsize $\pm$0.2496\\

        \cline{3-10}\rule{0pt}{2.3ex}
        & \multirow{2}{*}{$\alpha=10$}
        & 7.8101 & 12.734 & 7.2335 & 2.0040 & 10.228 & 8.1252 & 3.1834 & 6.0471\\
        &&\scriptsize$\pm$0.2030 & \scriptsize $\pm$0.1777 & \scriptsize $\pm$0.0002 & \tiny $\pm$4.4E-8 & \scriptsize $\pm$0.1156 & \scriptsize $\pm$0.1463 & \scriptsize $\pm$0.0838 & \scriptsize $\pm$0.1201\\

        \cline{3-10}\rule{0pt}{2.3ex}
        & \multirow{2}{*}{$\alpha=100$}
        & 7.5755 & 12.851 & 7.2335 & 2.0040 & 10.289 & 8.1077 & 3.2333 & 6.0803\\
        &&\scriptsize$\pm$0.0808 & \scriptsize $\pm$0.0429 & \tiny $\pm$1.3E-7 & \tiny $\pm$2.5E-8 & \scriptsize $\pm$0.0334 & \scriptsize $\pm$0.0307 & \scriptsize $\pm$0.0349 & \scriptsize $\pm$0.0346\\

        \cline{3-10}\rule{0pt}{2.3ex}
        &\multirow{2}{*}{\pignpi}
        &\textbf{0.0219} & \textbf{0.0292} & \textbf{0.0488} & \textbf{0.1317} & \textbf{0.0260} & \textbf{0.0233} & \textbf{0.0239} & \textbf{0.0419}\\
        &&\scriptsize\textbf{$\pm$0.0010} & \scriptsize \textbf{$\pm$0.0022} & \scriptsize \textbf{$\pm$0.0059} & \scriptsize \textbf{$\pm$0.0033} & \scriptsize \textbf{$\pm$0.0005} & \scriptsize \textbf{$\pm$0.0004} & \scriptsize \textbf{$\pm$0.0020} & \scriptsize \textbf{$\pm$0.0011}\\
        \hline\rule{0pt}{2.3ex}

        \multirow{12}{*}{$\textsf{MAE}_\textsf{symm}^{F}$}

        &\multirow{2}{*}{Baseline}
        &1.1099 & 1.7452 & 0.1248 & 0.6938 & 2.2074 & 1.7684 & 0.9399 & 1.4118\\
        &&\scriptsize$\pm$0.0785 & \scriptsize $\pm$0.0467 & \scriptsize $\pm$0.0137 & \scriptsize $\pm$0.2670 & \scriptsize $\pm$0.1852 & \scriptsize $\pm$0.0941 & \scriptsize $\pm$0.0257 & \scriptsize $\pm$0.0722\\
        \cline{3-10}\rule{0pt}{2.3ex}

        & \multirow{2}{*}{$\alpha=0.1$}
        & 0.1116 & 0.2262 & 0.0718 & 0.0243 & 0.1979 & 0.1863 & 0.0418 & 0.0988\\
        &&\scriptsize$\pm$0.0057 & \scriptsize $\pm$0.0107 & \scriptsize $\pm$0.0057 & \scriptsize $\pm$0.0005 & \scriptsize $\pm$0.0024 & \scriptsize $\pm$0.0009 & \scriptsize $\pm$0.0046 & \scriptsize $\pm$0.0069\\

        \cline{3-10}\rule{0pt}{2.3ex}
        & \multirow{2}{*}{$\alpha=1.0$}
        & 0.0057 & 0.0129 & 0.0160 & 0.0066 & 0.0084 & 0.0083 & 0.0039 & 0.0076\\
        &&\scriptsize$\pm$0.0003 & \scriptsize $\pm$0.0004 & \scriptsize $\pm$0.0005 & \scriptsize $\pm$0.0014 & \scriptsize $\pm$0.0003 & \scriptsize $\pm$0.0002 & \scriptsize $\pm$0.0004 & \scriptsize $\pm$0.0002\\

        \cline{3-10}\rule{0pt}{2.3ex}
        & \multirow{2}{*}{$\alpha=10$}
        & 0.0012 & 0.0023 & 0.0010 & \small $\text{1.2E-6}$ & 0.0013 & 0.0018 & 0.0008 & 0.0016\\
        && \tiny $\pm$3.0E-5 & \scriptsize $\pm$0.0001 & \scriptsize $\pm$0.0014 & \tiny $\pm$1.0E-6 & \scriptsize $\pm$0.0001 & \scriptsize $\pm$0.0001 & \tiny $\pm$2.9E-5 & \scriptsize $\pm$0.0001\\
        \cline{3-10}\rule{0pt}{2.3ex}

        & \multirow{2}{*}{$\alpha=100$}
        & 0.0002 & 0.0004 & \small $\text{2.4E-6}$& \small $\text{6.7E-7}$ & 0.0003 & 0.0004 & 0.0002 & 0.0003\\
        && \tiny $\pm$8.3E-6 & \tiny $\pm$1.9E-5 & \tiny $\pm$2.1E-6 & \tiny $\pm$6.9E-7 & \tiny $\pm$7.6E-6 & \tiny $\pm$1.4E-5 & \tiny $\pm$1.6E-5 & \tiny $\pm$1.3E-5\\

        \cline{3-10}\rule{0pt}{2.3ex}
        &\multirow{2}{*}{\pignpi}
        &\textbf{0.0075} & \textbf{0.0133} & \textbf{0.0185} & \textbf{0.0345} & \textbf{0.0136} & \textbf{0.0134} & \textbf{0.0026} & \textbf{0.0066}\\
        &&\scriptsize\textbf{$\pm$0.0003} & \scriptsize \textbf{$\pm$0.0008} & \scriptsize \textbf{$\pm$0.0036} & \scriptsize \textbf{$\pm$0.0017} & \scriptsize \textbf{$\pm$0.0004} & \scriptsize \textbf{$\pm$0.0004} & \scriptsize \textbf{$\pm$0.0004} & \scriptsize \textbf{$\pm$0.0001}\\
      \bottomrule
\end{tabularx}
}
\end{table}

\clearpage

\subsection{Robustness to noise}
\label{sec:noisy_input_experiments}

In this subsection, we evaluate the performance of the ML models under the assumption that the position measurements are impacted by noise. To simulate measurement noise, we impose white noise on the particle positions at each time step. Then, we compute particle velocities and accelerations from the noisy positions. Here, we consider the following equation to impose noise on the measured positions:
\begin{equation}
\label{eq:absolute_position_noisy}
    {\tilde{r}}_{i,k}^t \leftarrow {r}_{i, k}^t + {\beta \times X_{i, k}^t}
\end{equation}
where ${\tilde{r}}_{i, k}^t$ is the $k$-th dimension of the noisy position of particle $i$ at time $t$, $X_{i, k}^t\sim \mathcal{N}(0,\,1)$ is the random number sampled independently from the standard normal distribution and $\beta$ is a constant controlling the level of noise. The second term in Eq.~(\ref{eq:absolute_position_noisy}) represents the noise that is relevant to how we measure the position and how we discretize the space.

Different values for $\beta$ will result in different noise levels for both inputs (position and velocity) and the learning target (acceleration). Here, we define the \textsf{noise level} as the \textbf{average relative change of the target}:
\begin{equation}
\label{eq:noise-level-definition}
\textsf{noise level}=
\frac{1}{T}\frac{1}{\abs{V}}\frac{1}{d} \sum_{t=1}^T\sum_{i \in V}\sum_{k \in d} \frac{\abs{{\tilde{a}}_{i,k}^t - {{a}}_{i,k}^t}}{\abs{{{a}}_{i, k}^t}}
\end{equation}
Here, ${\tilde{a}}_{i,k}^t$ is the $k$-th dimension of the noisy acceleration of particle $i$ at time $t$\footnote{When computing the \textsf{noise level}, we only consider those $\abs{{{a}}_{i, k}^t}$ that are strictly larger than zero because we want to avoid dividing zero.}. 
We test 1e-7, 5e-7, 1e-6, 5e-6 and 1e-5 as the values for $\beta$. 
The corresponding \textsf{noise levels} of each dataset are summarized in Table~\ref{tab:noise_level}.

\begin{table}[h!]
  \centering
\caption{The \textsf{noise level} (Eq.~(\ref{eq:noise-level-definition})) of each dataset with different values for $\beta$.}
\label{tab:noise_level}
  \scalebox{1}{
  \begin{tabularx}{\textwidth}{cX X X X X X X X}
    \toprule
  &\makecell[Xt]{Spring \\ dim=2} &\makecell[Xt]{Spring \\ dim=3} & \makecell[Xt]{Charge\\ dim=2} & \makecell[Xt]{Charge\\ dim=3} &\makecell[Xt]{Orbital \\ dim=2} &\makecell[Xt]{Orbital \\ dim=3} & \makecell[Xt]{{Discnt} \\ dim=2} &\makecell[Xt]{{Discnt} \\ dim=3}\\
    \midrule
    $\beta$=1e-7 & 0.0117 & 0.0075 & 0.3057 & 1.6369 & 0.0036 & 0.0092 & 0.0102 & 0.0187\\
    $\beta$=5e-7 & 0.0405 & 0.0347 & 1.6401 & 6.1344 & 0.0182 & 0.0399 & 0.0515 & 0.0696\\
    $\beta$=1e-6 & 0.1509 & 0.0790 & 3.0269 & 18.979 & 0.0369 & 0.0786 & 0.0979 & 0.1284\\
    $\beta$=5e-6 & 0.5137 & 0.3936 & 14.767 & 43.030 & 0.1696 & 0.3980 & 0.4634 & 0.9941\\
    $\beta$=1e-5 & 0.8897 & 0.7108 & 29.881 & 119.34 & 0.3664 & 0.8667 & 0.9809 & 1.9767\\
	
	  \bottomrule
\end{tabularx}
}
\end{table}


Table~\ref{tab:noisy_input_baseline_force_result} and Table~\ref{tab:noisy_input_PIGNPI_force_result} report the performances of baseline and \pignpi to learn pairwise force with the noisy input.

The results show the performance of \pignpi decreases with increasing noise level. This makes sense because adding noise makes the training target less similar to the uncorrupted target that is associated with the pairwise force (note that we do not corrupt the ground-truth pairwise forces during evaluation). However, \pignpi can still preform reasonably well with small scale noise.

The performance of baseline model fluctuates significantly with different noise levels. This also makes sense because the baseline model does not learn the particle interactions.


Developing \pignpi further to make it even more robust to noisy input is left for future work.

\begin{table}[h!]
  \centering
    \caption{Quality of pairwise force prediction of the baseline model with noisy data. The imposed noise corresponds to Eq.~(\ref{eq:absolute_position_noisy}). ``Uncorrupted'' refers to the data without noise. Results averaged across five experiments.}
    \label{tab:noisy_input_baseline_force_result}
    \setlength\extrarowheight{-5pt}
  \scalebox{0.9}{
  \begin{tabularx}{\textwidth}{ccX X X X X X X X}
    \toprule
  & &\makecell[Xt]{Spring \\ dim=2} &\makecell[Xt]{Spring \\ dim=3} & \makecell[Xt]{Charge\\ dim=2} & \makecell[Xt]{Charge\\ dim=3} &\makecell[Xt]{Orbital \\ dim=2} &\makecell[Xt]{Orbital \\ dim=3} & \makecell[Xt]{{Discnt} \\ dim=2} &\makecell[Xt]{{Discnt} \\ dim=3}\\
    \midrule
        
        \multirow{12}{*}{\textsf{MAE\textsubscript{acc}}} 

        &\multirow{2}{*}{Uncorrupted}
        &0.0565 & 0.1076 & 0.2521 & 0.3824 & 0.0437 & 0.0439 & 0.0592 & 0.1171\\ 
        &&\scriptsize $\pm$0.0023 & \scriptsize $\pm$0.0012 & \scriptsize $\pm$0.0173 & \scriptsize $\pm$0.0559 & \scriptsize $\pm$0.0026 & \scriptsize $\pm$0.0014 & \scriptsize \scriptsize $\pm$0.0015 & \scriptsize $\pm$0.0010\\ 
        \cline{3-10}\rule{0pt}{2.3ex}

        & \multirow{2}{*}{$\beta$=1e-7}
        & 0.0586 & 0.1085 & 0.2753 & 0.3544 & 0.0435 & 0.0446 & 0.0586 & 0.1158\\
        &&\scriptsize$\pm$0.0016 & \scriptsize $\pm$0.0023 & \scriptsize $\pm$0.0194 & \scriptsize $\pm$0.0220 & \scriptsize $\pm$0.0006 & \scriptsize $\pm$0.0009 & \scriptsize $\pm$0.0013 & \scriptsize $\pm$0.0034\\

        \cline{3-10}\rule{0pt}{2.3ex}

        & \multirow{2}{*}{$\beta$=5e-7}
        & 0.0639 & 0.1165 & 0.2735 & 0.4071 & 0.0497 & 0.0571 & 0.0719 & 0.1317\\
        &&\scriptsize$\pm$0.0037 & \scriptsize $\pm$0.0027 & \scriptsize $\pm$0.0246 & \scriptsize $\pm$0.0286 & \scriptsize $\pm$0.0014 & \scriptsize $\pm$0.0010 & \scriptsize $\pm$0.0028 & \scriptsize $\pm$0.0046\\
        \cline{3-10}\rule{0pt}{2.3ex}

        & \multirow{2}{*}{$\beta$=1e-6}
        & 0.0767 & 0.1340 & 0.2933 & 0.3973 & 0.0654 & 0.0828 & 0.0890 & 0.1556\\
        &&\scriptsize$\pm$0.0012 & \scriptsize $\pm$0.0013 & \scriptsize $\pm$0.0159 & \scriptsize $\pm$0.0391 & \scriptsize $\pm$0.0006 & \scriptsize $\pm$0.0008 & \scriptsize $\pm$0.0013 & \scriptsize $\pm$0.0031\\

        \cline{3-10}\rule{0pt}{2.3ex}

        & \multirow{2}{*}{$\beta$=5e-6}
        & 0.2430 & 0.3758 & 0.4138 & 0.6577 & 0.2228 & 0.3331 & 0.2553 & 0.4174\\
        &&\scriptsize$\pm$0.0011 & \scriptsize $\pm$0.0016 & \scriptsize $\pm$0.0198 & \scriptsize $\pm$0.0276 & \scriptsize $\pm$0.0014 & \scriptsize $\pm$0.0013 & \scriptsize $\pm$0.0032 & \scriptsize $\pm$0.0016\\

        \cline{3-10}\rule{0pt}{2.3ex}

        & \multirow{2}{*}{$\beta$=1e-5}
        & 0.4694 & 0.7196 & 0.6222 & 1.0038 & 0.4370 & 0.6683 & 0.4809 & 0.7552\\
        &&\scriptsize$\pm$0.0039 & \scriptsize $\pm$0.0021 & \scriptsize $\pm$0.0159 & \scriptsize $\pm$0.0572 & \scriptsize $\pm$0.0011 & \scriptsize $\pm$0.0020 & \scriptsize $\pm$0.0020 & \scriptsize $\pm$0.0040\\

        \hline\rule{0pt}{2.3ex}
    
        \multirow{12}{*}{  \textsf{MAE\textsubscript{ef}}}

        &\multirow{2}{*}{Uncorrupted}
        &2.3979 & 3.8952 & 1.1832 & 0.6447 & 4.1010 & 3.5379 & 1.6536 & 2.5803\\
        &&\scriptsize$\pm$0.2095 & \scriptsize $\pm$0.7178 & \scriptsize $\pm$0.0955 & \scriptsize $\pm$0.1118 & \scriptsize $\pm$0.1467 & \scriptsize $\pm$0.7571 & \scriptsize $\pm$0.0640 & \scriptsize $\pm$0.2886\\
        \cline{3-10}\rule{0pt}{2.3ex}

        & \multirow{2}{*}{$\beta$=1e-7}
        & 1.9321 & 4.2515 & 1.3544 & 0.5249 & 3.3631 & 3.8075 & 1.6651 & 2.6424\\
        &&\scriptsize$\pm$0.6308 & \scriptsize $\pm$0.4175 & \scriptsize $\pm$0.0752 & \scriptsize $\pm$0.0468 & \scriptsize $\pm$0.7493 & \scriptsize $\pm$0.1573 & \scriptsize $\pm$0.2816 & \scriptsize $\pm$0.2322\\
        \cline{3-10}\rule{0pt}{2.3ex}

        & \multirow{2}{*}{$\beta$=5e-7}
        & 2.5900 & 4.0157 & 1.2892 & 0.6376 & 2.5351 & 3.5454 & 1.6866 & 1.9837\\
        &&\scriptsize$\pm$0.4020 & \scriptsize $\pm$0.3880 & \scriptsize $\pm$0.0139 & \scriptsize $\pm$0.0887 & \scriptsize $\pm$0.8932 & \scriptsize $\pm$0.4215 & \scriptsize $\pm$0.1694 & \scriptsize $\pm$0.2123\\

        \cline{3-10}\rule{0pt}{2.3ex}

        & \multirow{2}{*}{$\beta$=1e-6}
        & 2.6234 & 3.8559 & 1.3248 & 0.5286 & 4.0880 & 3.7811 & 1.3008 & 2.3136\\
        &&\scriptsize$\pm$0.9036 & \scriptsize $\pm$0.4356 & \scriptsize $\pm$0.0811 & \scriptsize $\pm$0.0782 & \scriptsize $\pm$1.3048 & \scriptsize $\pm$0.3673 & \scriptsize $\pm$0.2183 & \scriptsize $\pm$0.4424\\

        \cline{3-10}\rule{0pt}{2.3ex}

        & \multirow{2}{*}{$\beta$=5e-6}
        & 1.6493 & 3.9151 & 1.3440 & 0.6125 & 3.9054 & 3.2323 & 1.3307 & 2.0825\\
        &&\scriptsize$\pm$0.8436 & \scriptsize $\pm$0.5906 & \scriptsize $\pm$0.1431 & \scriptsize $\pm$0.0889 & \scriptsize $\pm$1.4067 & \scriptsize $\pm$0.6269 & \scriptsize $\pm$0.1672 & \scriptsize $\pm$0.3995\\

        \cline{3-10}\rule{0pt}{2.3ex}

        & \multirow{2}{*}{$\beta$=1e-5}
        & 2.2223 & 3.3053 & 1.2780 & 0.6939 & 4.4303 & 2.9186 & 1.2169 & 1.9050\\
        &&\scriptsize$\pm$0.4609 & \scriptsize $\pm$0.4277 & \scriptsize $\pm$0.1208 & \scriptsize $\pm$0.1686 & \scriptsize $\pm$0.9807 & \scriptsize $\pm$0.5783 & \scriptsize $\pm$0.3209 & \scriptsize $\pm$0.3288\\
        \hline\rule{0pt}{2.3ex}

        \multirow{12}{*}{ \textsf{MAE\textsubscript{nf}}}

        &\multirow{2}{*}{Uncorrupted}
        &11.652 & 20.967 & 6.8310 & 3.8038 & 18.194 & 16.677 & 10.786 & 15.651\\
        &&\scriptsize$\pm$0.9890 & \scriptsize $\pm$3.8552 & \scriptsize $\pm$0.5548 & \scriptsize $\pm$0.7523 & \scriptsize $\pm$0.6884 & \scriptsize $\pm$3.5212 & \scriptsize $\pm$0.3764 & \scriptsize $\pm$1.7983\\
        \cline{3-10}\rule{0pt}{2.3ex}

        & \multirow{2}{*}{$\beta$=1e-7}
        & 9.3805 & 22.903 & 7.8383 & 3.0275 & 14.940 & 17.918 & 10.838 & 15.971\\
        &&\scriptsize$\pm$3.0398 & \scriptsize $\pm$2.3024 & \scriptsize $\pm$0.4344 & \scriptsize $\pm$0.2899 & \scriptsize $\pm$3.3092 & \scriptsize $\pm$0.7365 & \scriptsize $\pm$1.8261 & \scriptsize $\pm$1.4671\\

        \cline{3-10}\rule{0pt}{2.3ex}

        & \multirow{2}{*}{$\beta$=5e-7}
        & 12.588 & 21.635 & 7.455 & 3.751 & 11.284 & 16.671 & 10.957 & 11.936\\
        &&\scriptsize$\pm$1.9510 & \scriptsize $\pm$2.0799 & \scriptsize $\pm$0.0881 & \scriptsize $\pm$0.5713 & \scriptsize $\pm$3.9460 & \scriptsize $\pm$1.9563 & \scriptsize $\pm$1.0616 & \scriptsize $\pm$1.3494\\

        \cline{3-10}\rule{0pt}{2.3ex}

        & \multirow{2}{*}{$\beta$=1e-6}
        & 12.746 & 20.718 & 7.661 & 2.987 & 18.104 & 17.759 & 8.3936 & 14.016\\
        &&\scriptsize$\pm$4.3777 & \scriptsize $\pm$2.3664 & \scriptsize $\pm$0.4736 & \scriptsize $\pm$0.5201 & \scriptsize $\pm$5.7265 & \scriptsize $\pm$1.7564 & \scriptsize $\pm$1.4703 & \scriptsize $\pm$2.7942\\

        \cline{3-10}\rule{0pt}{2.3ex}

        & \multirow{2}{*}{$\beta$=5e-6}
        & 7.9912 & 21.037 & 7.8270 & 3.6208 & 17.303 & 15.218 & 8.4946 & 12.392\\
        &&\scriptsize$\pm$4.0598 & \scriptsize $\pm$3.1630 & \scriptsize $\pm$0.8340 & \scriptsize $\pm$0.5847 & \scriptsize $\pm$6.1886 & \scriptsize $\pm$2.9087 & \scriptsize $\pm$1.0542 & \scriptsize $\pm$2.4643\\

        \cline{3-10}\rule{0pt}{2.3ex}

        & \multirow{2}{*}{$\beta$=1e-5}
        & 10.798 & 17.752 & 7.5451 & 4.2474 & 19.595 & 13.720 & 7.6793 & 11.189\\
        &&\scriptsize$\pm$2.2379 & \scriptsize $\pm$2.3123 & \scriptsize $\pm$0.7052 & \scriptsize $\pm$1.0325 & \scriptsize $\pm$4.3285 & \scriptsize $\pm$2.6955 & \scriptsize $\pm$2.0393 & \scriptsize $\pm$2.0076\\
        \hline \rule{0pt}{2.3ex}

        \multirow{12}{*}{$\textsf{MAE}_\textsf{symm}^{F}$}
        
        &\multirow{2}{*}{Uncorrupted}
        &1.1099 & 1.7452 & 0.1248 & 0.6938 & 2.2074 & 1.7684 & 0.9399 & 1.4118\\
        &&\scriptsize$\pm$0.0785 & \scriptsize $\pm$0.0467 & \scriptsize $\pm$0.0137 & \scriptsize $\pm$0.2670 & \scriptsize $\pm$0.1852 & \scriptsize $\pm$0.0941 & \scriptsize $\pm$0.0257 & \scriptsize $\pm$0.0722\\
        \cline{3-10}\rule{0pt}{2.3ex}

        & \multirow{2}{*}{$\beta$=1e-7}
        & 1.0733 & 1.8067 & 0.1232 & 0.4565 & 1.9961 & 1.9239 & 1.0063 & 1.4453\\
        &&\scriptsize$\pm$0.1043 & \scriptsize $\pm$0.0760 & \scriptsize $\pm$0.0352 & \scriptsize $\pm$0.0538 & \scriptsize $\pm$0.1710 & \scriptsize $\pm$0.1455 & \scriptsize $\pm$0.1074 & \scriptsize $\pm$0.0999\\
        \cline{3-10}\rule{0pt}{2.3ex}

        & \multirow{2}{*}{$\beta$=5e-7}
        & 0.9284 & 1.8013 & 0.1666 & 0.6596 & 2.0250 & 1.7911 & 0.9921 & 1.3530\\
        &&\scriptsize$\pm$0.0798 & \scriptsize $\pm$0.0401 & \scriptsize $\pm$0.0190 & \scriptsize $\pm$0.1836 & \scriptsize $\pm$0.1278 & \scriptsize $\pm$0.0679 & \scriptsize $\pm$0.1084 & \scriptsize $\pm$0.0336\\
        \cline{3-10}\rule{0pt}{2.3ex}

        & \multirow{2}{*}{$\beta$=1e-6}
        & 1.0539 & 1.6816 & 0.1592 & 0.3622 & 2.1475 & 1.7512 & 0.9143 & 1.3089\\
        &&\scriptsize$\pm$0.0690 & \scriptsize $\pm$0.0809 & \scriptsize $\pm$0.0224 & \scriptsize $\pm$0.2052 & \scriptsize $\pm$0.1492 & \scriptsize $\pm$0.1160 & \scriptsize $\pm$0.1050 & \scriptsize $\pm$0.1026\\
        \cline{3-10}\rule{0pt}{2.3ex}

        & \multirow{2}{*}{$\beta$=5e-6}
        & 0.8486 & 1.6024 & 0.1551 & 0.5979 & 1.9721 & 1.7956 & 0.8552 & 1.1735\\
        &&\scriptsize$\pm$0.0614 & \scriptsize $\pm$0.0432 & \scriptsize $\pm$0.0126 & \scriptsize $\pm$0.2170 & \scriptsize $\pm$0.1302 & \scriptsize $\pm$0.0726 & \scriptsize $\pm$0.1154 & \scriptsize $\pm$0.1105\\
        \cline{3-10}\rule{0pt}{2.3ex}

        & \multirow{2}{*}{$\beta$=1e-5}
        & 0.7940 & 1.4555 & 0.1581 & 0.7711 & 2.0862 & 1.5985 & 0.7857 & 1.1565\\
        &&\scriptsize$\pm$0.1133 & \scriptsize $\pm$0.0619 & \scriptsize $\pm$0.0241 & \scriptsize $\pm$0.3974 & \scriptsize $\pm$0.1608 & \scriptsize $\pm$0.0812 & \scriptsize $\pm$0.1024 & \scriptsize $\pm$0.0908\\
      \bottomrule
\end{tabularx}
}
\end{table}

\begin{table}[h!]
  \centering
    \caption{Quality of pairwise force prediction of the \pignpi with noisy data. The imposed noise corresponds to Eq.~(\ref{eq:absolute_position_noisy}). ``Uncorrupted'' refers to the data without noise. Results averaged across five experiments.}
    \label{tab:noisy_input_PIGNPI_force_result}
    \setlength\extrarowheight{-6pt}
  \scalebox{0.95}{
  \begin{tabularx}{\textwidth}{ccX X X X X X X X}
    \toprule
  & &\makecell[Xt]{Spring \\ dim=2} &\makecell[Xt]{Spring \\ dim=3} & \makecell[Xt]{Charge\\ dim=2} & \makecell[Xt]{Charge\\ dim=3} &\makecell[Xt]{Orbital \\ dim=2} &\makecell[Xt]{Orbital \\ dim=3} & \makecell[Xt]{{Discnt} \\ dim=2} &\makecell[Xt]{{Discnt} \\ dim=3}\\
    \midrule

        \multirow{12}{*}{\textsf{MAE\textsubscript{acc}}} 

        &\multirow{2}{*}{Uncorrupted}
        &{0.0206} & {0.0278} & {0.0425} & {0.1191} & {0.0202} & {0.0182} & {0.0227} & {0.0399}\\
        &&\scriptsize$\pm$0.0009 & \scriptsize $\pm$0.0021 & \scriptsize $\pm$0.0053 & \scriptsize $\pm$0.0027 & \scriptsize $\pm$0.0003 & \scriptsize $\pm$0.0003 & \scriptsize $\pm$0.0019 & \scriptsize $\pm$0.0011\\
        \cline{3-10}\rule{0pt}{2.3ex}

        & \multirow{2}{*}{$\beta$=1e-7}
        & 0.0213 & 0.0305 & 0.0421 & 0.1208 & 0.0208 & 0.0203 & 0.0274 & 0.0429\\
        &&\scriptsize$\pm$0.0009 & \scriptsize $\pm$0.0020 & \scriptsize $\pm$0.0031 & \scriptsize $\pm$0.0030 & \scriptsize $\pm$0.0005 & \scriptsize $\pm$0.0005 & \scriptsize $\pm$0.0045 & \scriptsize $\pm$0.0009\\
        \cline{3-10}\rule{0pt}{2.3ex}

        & \multirow{2}{*}{$\beta$=5e-7}
        & 0.0315 & 0.0449 & 0.0510 & 0.1393 & 0.0302 & 0.0374 & 0.0377 & 0.0614\\
        &&\scriptsize$\pm$0.0007 & \scriptsize $\pm$0.0012 & \scriptsize $\pm$0.0020 & \scriptsize $\pm$0.0055 & \scriptsize $\pm$0.0004 & \scriptsize $\pm$0.0003 & \scriptsize $\pm$0.0008 & \scriptsize $\pm$0.0006\\
        \cline{3-10}\rule{0pt}{2.3ex}

        & \multirow{2}{*}{$\beta$=1e-6}
        & 0.0499 & 0.0720 & 0.0697 & 0.1704 & 0.0473 & 0.0658 & 0.0585 & 0.0902\\
        &&\scriptsize$\pm$0.0004 & \scriptsize $\pm$0.0007 & \scriptsize $\pm$0.0021 & \scriptsize $\pm$0.0038 & \scriptsize $\pm$0.0005 & \scriptsize $\pm$0.0001 & \scriptsize $\pm$0.0038 & \scriptsize $\pm$0.0012\\
        \cline{3-10}\rule{0pt}{2.3ex}

        & \multirow{2}{*}{$\beta$=5e-6}
        & 0.2050 & 0.3062 & 0.2209 & 0.4163 & 0.2033 & 0.3088 & 0.2124 & 0.3257\\
        &&\scriptsize$\pm$0.0004 & \scriptsize $\pm$0.0004 & \scriptsize $\pm$0.0032 & \scriptsize $\pm$0.0063 & \scriptsize $\pm$0.0003 & \scriptsize $\pm$0.0010 & \scriptsize $\pm$0.0027 & \scriptsize $\pm$0.0006\\
        \cline{3-10}\rule{0pt}{2.3ex}

        & \multirow{2}{*}{$\beta$=1e-5}
        & 0.4060 & 0.6136 & 0.4215 & 0.7661 & 0.4102 & 0.6348 & 0.4146 & 0.6231\\
        &&\scriptsize$\pm$0.0009 & \scriptsize $\pm$0.0008 & \scriptsize $\pm$0.0016 & \scriptsize $\pm$0.0097 & \scriptsize $\pm$0.0009 & \scriptsize $\pm$0.0026 & \scriptsize $\pm$0.0017 & \scriptsize $\pm$0.0007\\

        \hline\rule{0pt}{2.3ex}

        \multirow{12}{*}{\textsf{MAE\textsubscript{ef}}}

        &\multirow{2}{*}{Uncorrupted}
        &{0.0063} & {0.0101} & {0.0136} & {0.0363} & {0.0093} & {0.0095} & {0.0040} & {0.0079}\\
        &&\scriptsize$\pm$0.0002 & \scriptsize $\pm$0.0007 & \scriptsize $\pm$0.0023 & \scriptsize $\pm$0.0015 & \scriptsize $\pm$0.0002 & \scriptsize $\pm$0.0001 & \scriptsize $\pm$0.0004 & \scriptsize $\pm$0.0002\\
        \cline{3-10}\rule{0pt}{2.3ex}

        & \multirow{2}{*}{$\beta$=1e-7}
        & 0.0064 & 0.0107 & 0.0135 & 0.0359 & 0.0092 & 0.0101 & 0.0047 & 0.0082\\
        &&\scriptsize$\pm$0.0002 & \scriptsize $\pm$0.0007 & \scriptsize $\pm$0.0016 & \scriptsize $\pm$0.0009 & \scriptsize $\pm$0.0003 & \scriptsize $\pm$0.0003 & \scriptsize $\pm$0.0010 & \scriptsize $\pm$0.0001\\
        \cline{3-10}\rule{0pt}{2.3ex}

        & \multirow{2}{*}{$\beta$=5e-7}
        & 0.0068 & 0.0112 & 0.0143 & 0.0381 & 0.0097 & 0.0111 & 0.0043 & 0.0089\\
        &&\scriptsize$\pm$0.0002 & \scriptsize $\pm$0.0005 & \scriptsize $\pm$0.0011 & \scriptsize $\pm$0.0016 & \scriptsize $\pm$0.0001 & \scriptsize $\pm$0.0001 & \scriptsize $\pm$0.0002 & \scriptsize $\pm$0.0001\\
        \cline{3-10}\rule{0pt}{2.3ex}

        & \multirow{2}{*}{$\beta$=1e-6}
        & 0.0078 & 0.0131 & 0.0163 & 0.0413 & 0.0108 & 0.0136 & 0.0055 & 0.0106\\
        &&\scriptsize$\pm$0.0001 & \scriptsize $\pm$0.0004 & \scriptsize $\pm$0.0013 & \scriptsize $\pm$0.0016 & \scriptsize $\pm$0.0002 & \scriptsize $\pm$0.0002 & \scriptsize $\pm$0.0013 & \scriptsize $\pm$0.0003\\
        \cline{3-10}\rule{0pt}{2.3ex}

        & \multirow{2}{*}{$\beta$=5e-6}
        & 0.0162 & 0.0275 & 0.0265 & 0.0572 & 0.0223 & 0.0373 & 0.0114 & 0.0211\\
        &&\scriptsize$\pm$0.0002 & \scriptsize $\pm$0.0004 & \scriptsize $\pm$0.0024 & \scriptsize $\pm$0.0024 & \scriptsize $\pm$0.0005 & \scriptsize $\pm$0.0010 & \scriptsize $\pm$0.0027 & \scriptsize $\pm$0.0006\\
        \cline{3-10}\rule{0pt}{2.3ex}

        & \multirow{2}{*}{$\beta$=1e-5}
        & 0.0321 & 0.0545 & 0.0411 & 0.0984 & 0.0421 & 0.0826 & 0.0247 & 0.0410\\
        &&\scriptsize$\pm$0.0007 & \scriptsize $\pm$0.0006 & \scriptsize $\pm$0.0005 & \scriptsize $\pm$0.0065 & \scriptsize $\pm$0.0004 & \scriptsize $\pm$0.0029 & \scriptsize $\pm$0.0031 & \scriptsize $\pm$0.0007\\

        \hline\rule{0pt}{2.3ex}

        \multirow{12}{*}{ \textsf{MAE\textsubscript{nf}}}
        \rule{0pt}{2.3ex}
        &\multirow{2}{*}{Uncorrupted}
        &{0.0219} & {0.0292} & {0.0488} & {0.1317} & {0.0260} & {0.0233} & {0.0239} & {0.0419}\\
        &&\scriptsize$\pm$0.0010 & \scriptsize $\pm$0.0022 & \scriptsize $\pm$0.0059 & \scriptsize $\pm$0.0033 & \scriptsize $\pm$0.0005 & \scriptsize $\pm$0.0004 & \scriptsize $\pm$0.0020 & \scriptsize $\pm$0.0011\\
        \cline{3-10}\rule{0pt}{2.3ex}

        & \multirow{2}{*}{$\beta$=1e-7}
        & 0.0230 & 0.0324 & 0.0486 & 0.1334 & 0.0266 & 0.0260 & 0.0294 & 0.0456\\
        &&\scriptsize$\pm$0.0010 & \scriptsize $\pm$0.0021 & \scriptsize $\pm$0.0035 & \scriptsize $\pm$0.0034 & \scriptsize $\pm$0.0007 & \scriptsize $\pm$0.0006 & \scriptsize $\pm$0.0046 & \scriptsize $\pm$0.0008\\
        \cline{3-10}\rule{0pt}{2.3ex}

        & \multirow{2}{*}{$\beta$=5e-7}
        & 0.0370 & 0.0528 & 0.0607 & 0.1581 & 0.0386 & 0.0481 & 0.0439 & 0.0704\\
        &&\scriptsize$\pm$0.0007 & \scriptsize $\pm$0.0011 & \scriptsize $\pm$0.0024 & \scriptsize $\pm$0.0058 & \scriptsize $\pm$0.0005 & \scriptsize $\pm$0.0004 & \scriptsize $\pm$0.0008 & \scriptsize $\pm$0.0007\\
        \cline{3-10}\rule{0pt}{2.3ex}

        & \multirow{2}{*}{$\beta$=1e-6}
        & 0.0610 & 0.0880 & 0.0846 & 0.1974 & 0.0606 & 0.0846 & 0.0701 & 0.1070\\
        &&\scriptsize$\pm$0.0004 & \scriptsize $\pm$0.0007 & \scriptsize $\pm$0.0023 & \scriptsize $\pm$0.0046 & \scriptsize $\pm$0.0007 & \scriptsize $\pm$0.0002 & \scriptsize $\pm$0.0038 & \scriptsize $\pm$0.0011\\
        \cline{3-10}\rule{0pt}{2.3ex}

        & \multirow{2}{*}{$\beta$=5e-6}
        & 0.2605 & 0.3876 & 0.2782 & 0.5092 & 0.2602 & 0.3968 & 0.2673 & 0.4078\\
        &&\scriptsize$\pm$0.0003 & \scriptsize $\pm$0.0004 & \scriptsize $\pm$0.0036 & \scriptsize $\pm$0.0082 & \scriptsize $\pm$0.0006 & \scriptsize $\pm$0.0011 & \scriptsize $\pm$0.0027 & \scriptsize $\pm$0.0005\\
        \cline{3-10}\rule{0pt}{2.3ex}

        & \multirow{2}{*}{$\beta$=1e-5}
        & 0.5143 & 0.7800 & 0.5353 & 0.9458 & 0.5273 & 0.8101 & 0.5250 & 0.7843\\
        &&\scriptsize$\pm$0.0008 & \scriptsize $\pm$0.0008 & \scriptsize $\pm$0.0019 & \scriptsize $\pm$0.0102 & \scriptsize $\pm$0.0009 & \scriptsize $\pm$0.0029 & \scriptsize $\pm$0.0019 & \scriptsize $\pm$0.0006\\

        \hline\rule{0pt}{2.3ex}

        \multirow{12}{*}{$\textsf{MAE}_\textsf{symm}^{F}$}
        \rule{0pt}{2.3ex}
        &\multirow{2}{*}{Uncorrupted}
        &{0.0075} & {0.0133} & {0.0185} & {0.0345} & {0.0136} & {0.0134} & {0.0026} & {0.0066}\\
        &&\scriptsize$\pm$0.0003 & \scriptsize $\pm$0.0008 & \scriptsize $\pm$0.0036 & \scriptsize $\pm$0.0017 & \scriptsize $\pm$0.0004 & \scriptsize $\pm$0.0004 & \scriptsize $\pm$0.0004 & \scriptsize $\pm$0.0001\\
        \cline{3-10}\rule{0pt}{2.3ex}

        & \multirow{2}{*}{$\beta$=1e-7}
        & 0.0076 & 0.0139 & 0.0180 & 0.0339 & 0.0132 & 0.0141 & 0.0031 & 0.0069\\
        &&\scriptsize$\pm$0.0002 & \scriptsize $\pm$0.0008 & \scriptsize $\pm$0.0030 & \scriptsize $\pm$0.0010 & \scriptsize $\pm$0.0005 & \scriptsize $\pm$0.0005 & \scriptsize $\pm$0.0006 & \scriptsize $\pm$0.0000\\
        \cline{3-10}\rule{0pt}{2.3ex}

        & \multirow{2}{*}{$\beta$=5e-7}
        & 0.0083 & 0.0149 & 0.0201 & 0.0369 & 0.0139 & 0.0155 & 0.0032 & 0.0079\\
        &&\scriptsize$\pm$0.0003 & \scriptsize $\pm$0.0007 & \scriptsize $\pm$0.0022 & \scriptsize $\pm$0.0016 & \scriptsize $\pm$0.0002 & \scriptsize $\pm$0.0002 & \scriptsize $\pm$0.0002 & \scriptsize $\pm$0.0002\\
        \cline{3-10}\rule{0pt}{2.3ex}

        & \multirow{2}{*}{$\beta$=1e-6}
        & 0.0098 & 0.0173 & 0.0223 & 0.0407 & 0.0154 & 0.0194 & 0.0046 & 0.0096\\
        &&\scriptsize$\pm$0.0001 & \scriptsize $\pm$0.0005 & \scriptsize $\pm$0.0017 & \scriptsize $\pm$0.0012 & \scriptsize $\pm$0.0002 & \scriptsize $\pm$0.0005 & \scriptsize $\pm$0.0012 & \scriptsize $\pm$0.0001\\
        \cline{3-10}\rule{0pt}{2.3ex}

        & \multirow{2}{*}{$\beta$=5e-6}
        & 0.0222 & 0.0382 & 0.0354 & 0.0653 & 0.0314 & 0.0529 & 0.0124 & 0.0238\\
        &&\scriptsize$\pm$0.0003 & \scriptsize $\pm$0.0004 & \scriptsize $\pm$0.0023 & \scriptsize $\pm$0.0023 & \scriptsize $\pm$0.0008 & \scriptsize $\pm$0.0014 & \scriptsize $\pm$0.0037 & \scriptsize $\pm$0.0011\\

        \cline{3-10}\rule{0pt}{2.3ex}

        & \multirow{2}{*}{$\beta$=1e-5}
        & 0.0450 & 0.0768 & 0.0485 & 0.1184 & 0.0601 & 0.1174 & 0.0314 & 0.0518\\
        &&\scriptsize$\pm$0.0010 & \scriptsize $\pm$0.0011 & \scriptsize $\pm$0.0016 & \scriptsize $\pm$0.0085 & \scriptsize $\pm$0.0008 & \scriptsize $\pm$0.0049 & \scriptsize $\pm$0.0048 & \scriptsize $\pm$0.0011\\

      \bottomrule
\end{tabularx}
}
\end{table}
\clearpage

\subsection{Visualization of force and potential functions used in simulation}
\label{sec:force-potential-functions-visualization}
Fig.~\ref{fig:force-potential-functions-plot} shows the inter-particle potential energy $\potEnergy$ and the inter-particle pairwise force $\force$ used for generating the simulations. $\potEnergy$ and $\force$ are the functions of relative distance.

\begin{figure}[h!]
    \centering
    \includegraphics[width=1.0\linewidth]{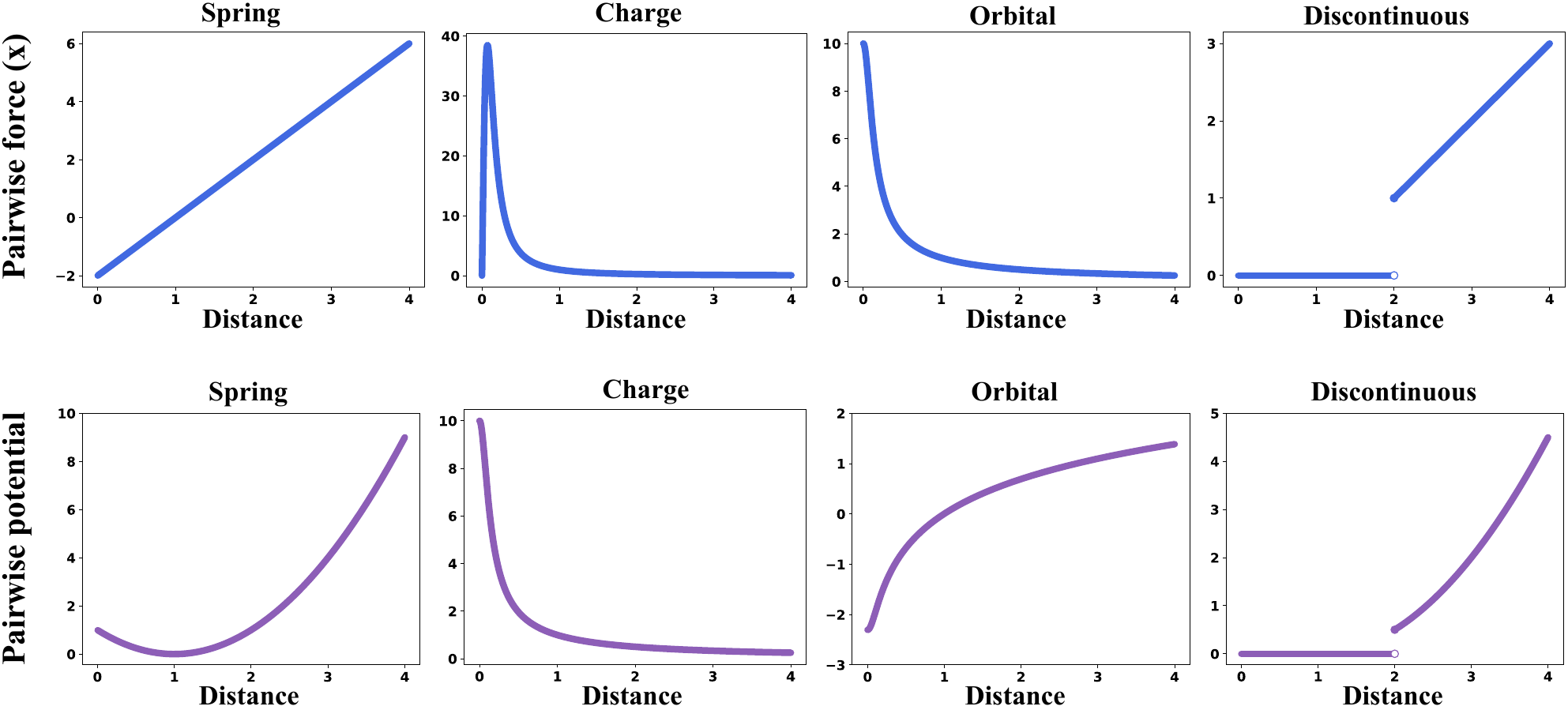}
    \caption{Visualization of pairwise force and potential with different distances. Blue color shows the pairwise force in x dimension as the function of the relative distance between particles. Purple color in second row shows the pairwise potential with different distance. In this visualization, we set the electric charge and particle masses to one. }
    \label{fig:force-potential-functions-plot}
\end{figure}

\clearpage

\end{document}